\pdfoutput=1
\documentclass[10pt, logo, twocolumn, copyright]{nv}
\usepackage{graphicx}
\definecolor{nvidiagreen}{HTML}{76B900}

\usepackage{mdframed}
\usepackage{color}
\usepackage{xcolor}
\usepackage[utf8]{inputenc} 
\usepackage[T1]{fontenc}    

\usepackage{amsfonts}       
\usepackage{nicefrac}       
\usepackage{microtype}      
\usepackage{multirow}
\usepackage{multicol}
\usepackage{tabto}
\usepackage{xspace}
\usepackage{amsmath}
\usepackage{adjustbox}
\usepackage{enumitem}
\usepackage{wrapfig}
\usepackage{dblfloatfix}
\usepackage{times}
\usepackage{verbatim}
\usepackage{amssymb}
\usepackage{mathtools}
\usepackage{caption}
\usepackage{subcaption}
\usepackage{array}
\usepackage{colortbl}
\usepackage{booktabs}
\usepackage{bbm}
\usepackage{makecell}
\usepackage{float}
\usepackage{siunitx}
\usepackage{pifont}
\usepackage{marvosym}
\usepackage{listings}
\usepackage{pdflscape}
\usepackage{footmisc}
\definecolor{codebg}{RGB}{245, 245, 245} 
\definecolor{keywordcolor}{RGB}{0, 0, 153} 
\definecolor{commentcolor}{RGB}{34, 139, 34} 
\definecolor{stringcolor}{RGB}{163, 21, 21}
\definecolor{numbercolor}{RGB}{128, 128, 128}
\usepackage{url}
\usepackage{tabularx}
\usepackage{arydshln}
\usepackage{hhline}
\usepackage{diagbox}
\usepackage{tcolorbox}
\usepackage[nameinlink]{cleveref}
\usepackage{hyperref}
\usepackage{fp} 
\crefname{equation}{Eq.}{Eqs.}
\crefname{figure}{Fig.}{Figs.}
\crefname{section}{Sec.}{Sec.}
\crefname{appendix}{App.}{App.}
\crefname{table}{Tab.}{Tabs.}
\crefname{algorithm}{Algo}{Algo}
\crefname{thm}{Thm}{Thm}
\Crefname{thm}{Thm}{Thm}
\crefname{prop}{Prop}{Prop}

\newcommand{\xmarker}{\ding{55}}
\newcommand{\cmarker}{\ding{51}}

\title{Eagle 2: Building Post-Training Data Strategies from Scratch for
Frontier Vision-Language Models}

\author{
Zhiqi Li\textsuperscript{1,$*$} ~~
Guo Chen\textsuperscript{1,$*$} ~~
Shilong Liu\textsuperscript{2,$*$} ~~
Shihao Wang\textsuperscript{3,$*$} ~~
Vibashan VS\textsuperscript{4,$*$} ~~
Yishen Ji\textsuperscript{1} ~~
Shiyi Lan ~~
Hao Zhang ~~
Yilin Zhao\textsuperscript{5,$*$} ~~
Subhashree Radhakrishnan ~~
Nadine Chang ~~
Karan Sapra ~~
Amala Deshmukh ~~
Tuomas Rintamaki ~~
Matthieu Le ~~
De-An Huang ~~
Ilia Karmanov ~~
Lukas Voegtle ~~
Philipp Fischer ~~
Timo Roman ~~
Tong Lu\textsuperscript{1} ~~~~~~~~~~~~~~
Jose M. Alvarez ~~
Bryan Catanzaro ~~
Jan Kautz ~~
Andrew Tao ~~
Guilin Liu\textsuperscript{$\dag$} ~~
Zhiding Yu\textsuperscript{$\dag$}
}
\correspondingauthor{X}

\begin{abstract}
\textbf{Abstract:} Recently, promising progress has been made by open-source vision-language models (VLMs) in bringing their capabilities closer to those of proprietary frontier models. However, most open-source models only publish their final model weights, leaving the critical details of data strategies and implementation largely opaque. In this work, we address VLM post-training from a data-centric perspective, showing the key role of data strategy in developing frontier VLMs. By studying and building our post-training data strategy from scratch, we share detailed insights into the development processes, aiming to benefit the development of competitive models for the open-source community. Our introduced data strategy, together with training recipes and model design, leads to a family of performant VLMs named \textit{Eagle 2}. Specifically, Eagle2-9B achieves state-of-the-art results across various multimodal benchmarks, matching certain competitive models with up to 70B parameters.
\vspace{2mm}
\newline
\textbf{Links:} \hspace{2pt}
{\hypersetup{urlcolor=nvidiagreen}
\href{https://github.com/NVlabs/EAGLE?tab=readme-ov-file}{Github Code} | \href{https://huggingface.co/collections/nvidia/Eagle2-6764ba887fa1ef387f7df067}{HF Models} | \href{http://eagle.viphk1.nnhk.cc/}{Demo}}
\end{abstract}

\begin{document}

\maketitle

\section{Introduction}
\label{sec:intro}
Built upon large language models (LLMs), vision-language models (VLMs)~\cite{alayrac2022flamingo,li2022blip,chen2022pali,liu2023llava} aim to enable LLMs to see.
With the ability to visually perceive the world, VLMs are able to take in multimodal information, and as a result, handle a broader range of intelligent applications. There is thus a growing interest to use VLMs as the backbone for reasoning and decision making in various applications, such as intelligent agents~\cite{hong2023cogagent}, autonomous driving~\cite{tian2024drivevlm,marcu2023lingoqa}, and embodied AI~\cite{brohan2023rt,driess2023palme,kim2024openvla}.

The community has delved deeply into the architecture and training methodologies of VLMs with significant advances. A predominant strategy to align the vision and language modalities is through post-training on pre-trained LLMs, with the LLaVA family ~\cite{liu2023llava} being the representative examples. Based on the level of transparency, current VLM models can also be broadly categorized into three types: 1) commercially closed-source models (e.g., GPT-4v/o~\cite{gpt4v} and Claude~\cite{claude3series2024}), 2) frontier models with publicly available weights (e.g., Qwen2-VL~\cite{yang2024qwen2}, InternVL2~\cite{chen2024internvl2} and Llama 3.1~\cite{dubey2024llama3}), and 3) fully open-source models (e.g., Cambrian-1~\cite{tong2024cambrian} and the LLaVA family~\cite{liu2023llava, li2024llavaonevision}).
\begin{figure}[t]
    \vspace{8mm}
    \centering
    \includegraphics[width=\linewidth]{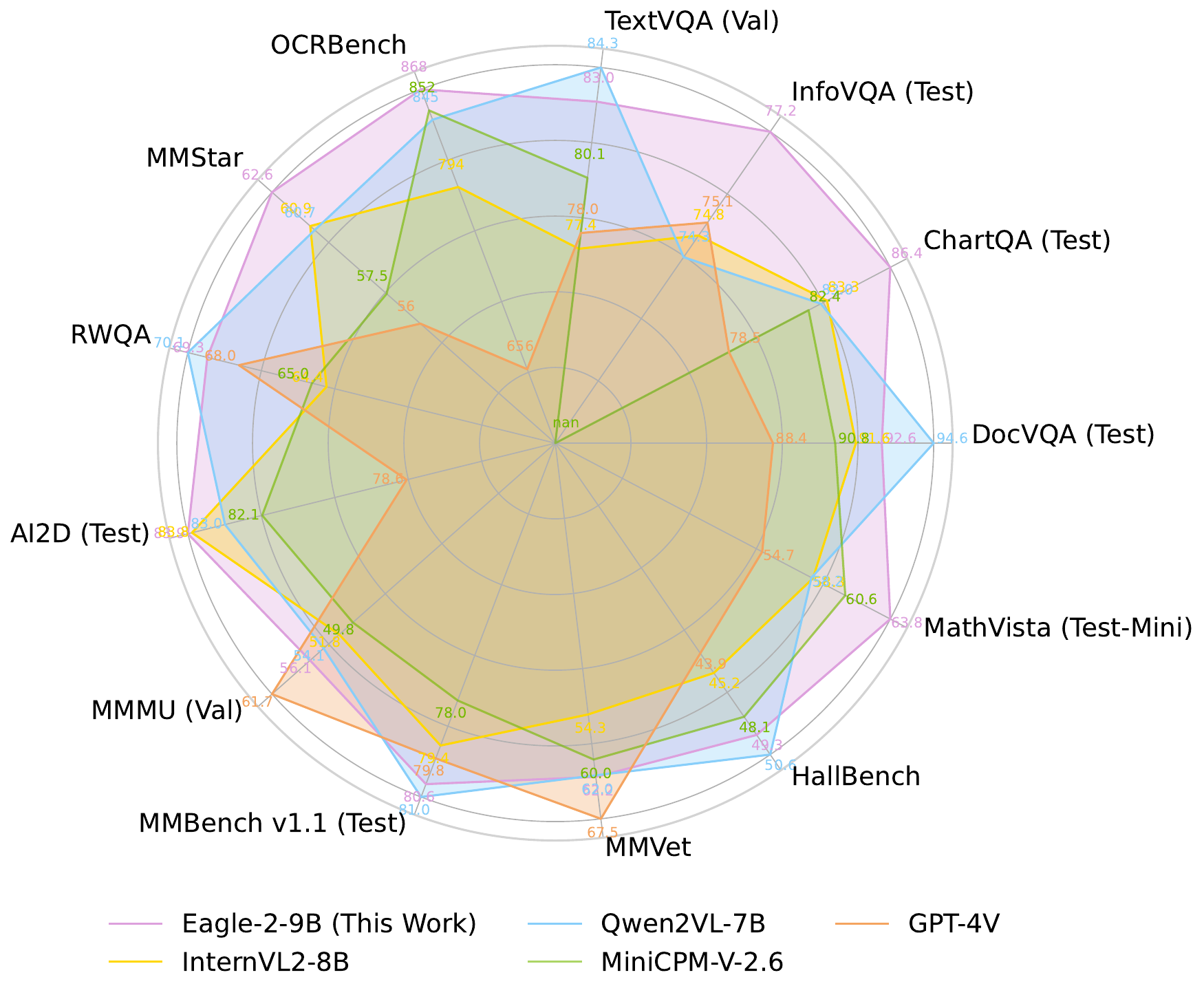}
    \caption{Overview of Eagle2-9B’s result across different multimodal benchmarks, in comparison to state-of-the-art open-source and commercial frontier models.}
    \label{fig:radar_compare}
\end{figure}

Recently, some frontier models with publicly available weights have been shown to match closed-source commercial models on key benchmarks while offering better customization for downstream applications. However, the technical details provided by these models are often insufficient for reproduction. On the other hand, fully open-source models tend to disclose extensive technical details, including both the dataset strategies and training recipes. These details unveils the secret sources in building customized VLM models, which enables easier reproduction and helps the community to develop technologies faster. However, most of the open source models still lag behind their frontier counterparts. For instance, on the OpenCompass~\cite{opencompass2023} benchmark, LLaVA-OneVision-72B~\cite{li2024llavaonevision} still ranks slightly behind InternVL2-40B~\cite{chen2024internvl2} despite having a stronger LLM backbone. We thus ask the following question: \textit{{What could help the community to develop more competitive open-source frontier VLMs?}}

\subsection{Data Strategy}
Our answer to the above question is the data-strategy. Assuming the same pre-trained LLM backbone, we posit that data is the most decisive factor to obtain high-quality models. We thus adopt a centralized strategy to build our post-training data. For fully open-source models, various constraints such as computing resources may limit the study on more dataset sources, despite their intention to make the data recipe publicly available. This limitation often affects their capability compared to models that can access and utilize a wider range of data sources.

\textbf{``Diversity first, then quality''.} 
We follow this principle throughout our development and push it to the extreme. Our optimization of the data results in consistent improvements in model. Our contributions involve: 1) a data collection strategy leading to a large-scale highly diverse data pool with \textbf{180+} sources, 2) a data filtering strategy to remove low-quality samples, 3) a data selection strategy to construct high-quality subsets, and 4) a series of data augmentation techniques to enrich the existing data. This series of strategies are shown to improve the model significantly.

\begin{figure}[t]
    \centering
    \includegraphics[width=\linewidth]{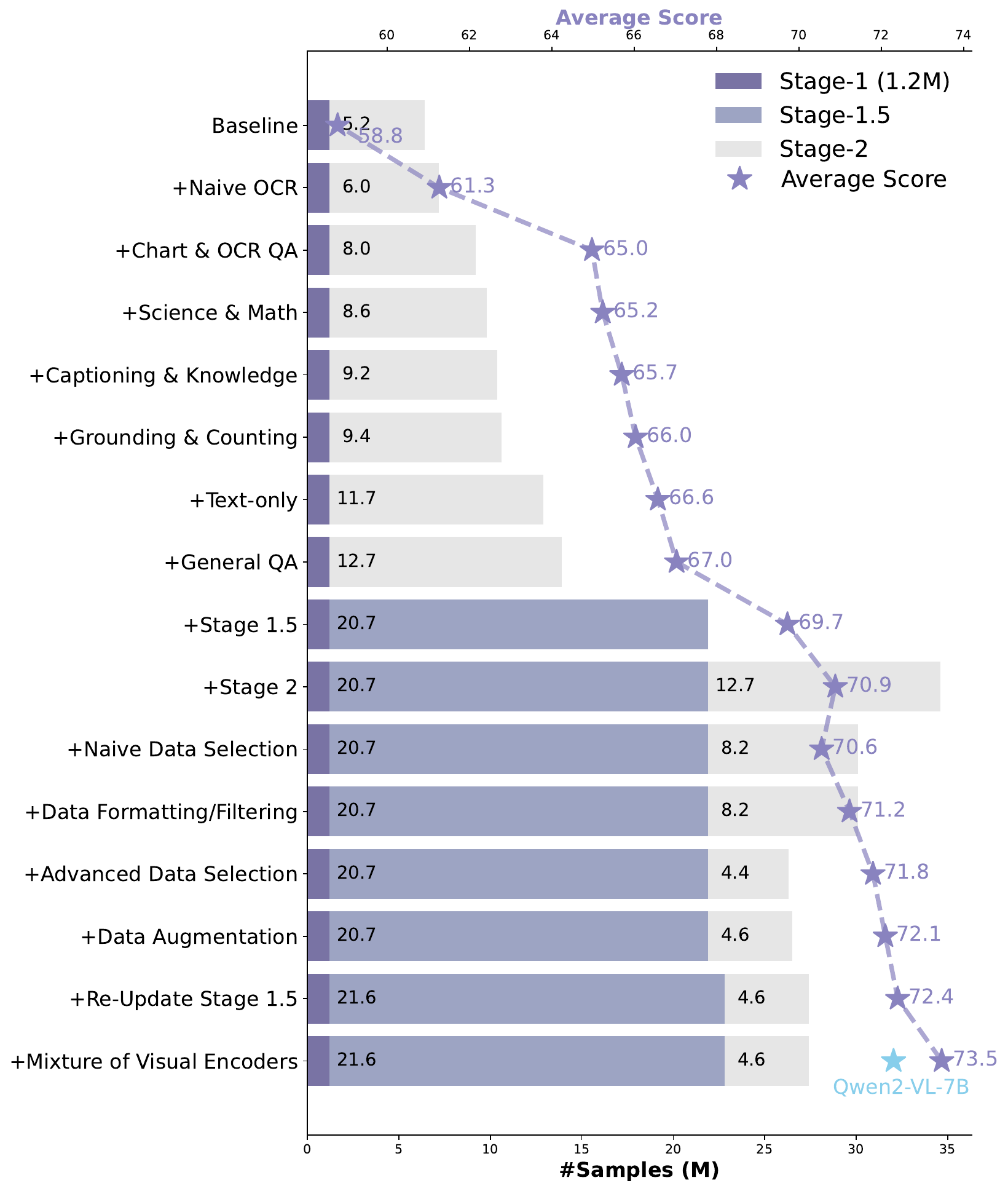}
    \caption{Step-by-Step ablation of Eagle 2. We show the average scores of 13 benchmarks, detailed numbers are in Table~\ref{tab:abl_baseline} and~\ref{tab:stage_1_5}.}
    \label{fig:line_abl}
\end{figure}

\subsection{Model Architecture}
Besides data strategy, another important factor considered in this work is model architecture. Since different architecture designs have been well-studied in open-source models, their properties are relatively transparent to the community. Despite various designs such as Q-Former~\cite{zhang2024vision_qformer} and Hybrid-Attention~\cite{bai2023qwenvl}, simple MLP connector is still the most popular choice to connect the vision encoder and LLM. With the rapid advances in architecture designs in contemporary VLMs, model structure is no longer a primary factor driving performance differences among models. However, this does not imply that there is no room for further improvement in architecture.

\textbf{Tiled mixture of vision encoders.} Inspired by the works of InternVL~\cite{chen2024far,chen2024internvl2}, Eagle~\cite{eagle} and Cambrian-1~\cite{tong2024cambrian}, we follow a vision-centric design where we adopt both dynamic tiling and mixture of vision encoders (MoVE) in one unified design. Specifically, each image tile is encoded by channel-concatenated MoVE, therefore allowing high-resolution input from tiling while maintaining the robust perception from MoVE. Similar to~\cite{eagle}, we follow a ``SigLIP~\cite{zhai2023siglip} + X (ConvNeXt~\cite{liu2022convnet})'' configuration. Compared to SigLIP alone, tiled MoVE yields significant improvements despite having tiling, particularly in tasks like OCR and Chart/Document VQA.

\subsection{Training Recipe}
Which training recipe to be used? In this context, the training recipe primarily refers to various configurations for training a VLM. With the same dataset, different recipes can still have a significant impact on the final performance. Although the training recipes for the state-of-the-art VLMs~\cite{wang2024qwen2vl, chen2024internvl2, openai2023gpt4} are somewhat unclear, the details shared by existing work~\cite{tong2024cambrian, liu2023llava, li2024llavaonevision} can offer a solid baseline.  However, to further improve model performance, it is necessary to explore more effective training recipe.

\textbf{Three-stage training.} We adopt a three-stage training strategy to best leverage the training data. In particular, the first stage (Stage-1) is used to align language and image modality via training the MLP connector. The next stage (Stage-1.5) trains the full model with a large-scale diverse data. The final stage (Stage-2) continues training the full model with a carefully crafted, high-quality visual instruction tuning dataset. In Stage-1.5, we incorporate all available visual instruction data, rather than limiting it to captioning or knowledge data alone. Our results show that this approach yields substantial improvements over the commonly used two-stage training strategy~\cite{liu2023llava}. We also identify limitations in existing open-source frameworks concerning data packing and introduce a balanced data packing approach to address the issue.

\subsection{Summary}
Our extensive exploration on data strategy, model architecture and training recipe is shown in Fig.~\ref{fig:line_abl}, resulting in a family of VLMs named \textit{Eagle 2}. Through sharing the journey of exploration, we aim to ``teach readers to fish than to give them a fish'', by sharing our data strategies, model design and training recipe, detailing the iterative development process than merely displaying the final benchmark results and demonstrations. The Eagle 2 model family spans a range of various scales, including 1B, 2B and 9B parameters. As shown in Fig.~\ref{fig:radar_compare}, Eagle2-9B can match or outperform frontier open-source and commercially closed-source VLMs on a set of common multimodal benchmarks.

\section{Method}

\begin{table}[t]
\centering
\setlength{\tabcolsep}{14pt}
\resizebox{0.45\textwidth}{!}{%
\renewcommand{\arraystretch}{0.9}
\begin{tabular}{l|c|c}
\multicolumn{2}{c|}{Name}                  & Eagle2-Baseline \\
\hline
\multirow{3}{*}{Model} & Vision Encoder & Siglip-400M~\cite{zhai2023siglip}\\
& Connector & MLP\\
& LLM & Qwen2.5-7B-Instruct~\cite{qwen2.5}\\
\hline
\multicolumn{2}{c|}{Resolution} & 448$\times$\{(1,1), (1,2), (2,2) $\cdots$, (1,12)\}\\
\hline
\multirow{2}{*}{Stage-1} & Training Data & ALLaVA(1.2M)~\cite{liu2023llava} \\
& Trainable Module & Connector\\
\hline
\multirow{2}{*}{Stage-2} & Training Data & Cambrian-1 (5.2M)~\cite{tong2024cambrian}\\
& Trainable Module & Full Model\\
\hline
\end{tabular}%
}
\caption{Baseline Settings.}
\label{baseline}
\end{table}
\subsection{Baseline Setting}
As shown in Tab.~\ref{baseline}, our initial baseline starts with the Cambrian dataset~\cite{tong2024cambrian} using LLaVA's~\cite{liu2023llava} two-stage training recipe. We remove some low-quality data from Cambrian-7M, such as ShareGPT-4V, GPT-77K and Data-Engine-161K, ultimately resulting in a subset of 5.2M samples. The model incorporates an MLP connector to bridge the vision encoder with the LLM and employs image tiling for dynamic resolution. 
Starting from this baseline, we enhance Eagle 2 in three key aspects: (1) data strategy, (2) training recipe, and (3) model architecture. These optimizations enable the model to achieve state-of-the-art performance.

\begin{table*}[t]
	\renewcommand{\arraystretch}{1.5}
	\begin{subtable}{\textwidth}
	    \resizebox{\textwidth}{!}{
		    \begin{tabular}{c|p{17cm}}
			\textbf{Category} & \textbf{Dataset}\\
			\midrule
			\multirow{2}{*}{\raisebox{+3pt}{Captioning \& Knowledge}} & ShareGPT4o~\cite{opengvlab_sharegpt4o_dataset}, KVQA~\cite{shah2019kvqa}, Movie-Posters~\cite{skvarre_movie_posters_100k}, Google-Landmark~\cite{weyand2020googlelandmark}, WikiArt~\cite{wikiart_dataset}, Weather-QA~\cite{ma2024weatherqa}, Coco-Colors~\cite{mscoco-controlnet-canny-less-colors}, music-sheet~\cite{sheet_music_clean}, SPARK~\cite{yu2024spark}, Image-Textualization~\cite{pi2024image_textualization}, SAM-Caption~\cite{pixart_alpha_sam_llava_captions10m}, Tmdb-Celeb-10k~\cite{ashraq_tmdb_celeb_10k}\\
			\rowcolor{gray!15}
			\multirow{2}{*}{\raisebox{+3pt}{Mathematics}} & GeoQA+~\cite{cao2022geoqa_plus}, MathQA~\cite{yu2023mathqa}, CLEVR-Math/Super~\cite{lindstrom2022clevrmath, li2023superclevr}, Geometry3K~\cite{lu2021geometry3k}, MAVIS-math-rule-geo~\cite{zhang2024mavis}, MAVIS-math-metagen~\cite{zhang2024mavis}, InterGPS~\cite{lu2021intergps}, Raven~\cite{zhang2019raven}, GEOS~\cite{seo2015geos}, UniGeo~\cite{chen2022unigeo}\\
			\multirow{2}{*}{\raisebox{+3pt}{Science}} & AI2D~\cite{kembhavi2016ai2d}, ScienceQA~\cite{lu2022scienceqa}, TQA~\cite{kembhavi2017tqa}, PathVQA~\cite{he2020pathvqa}, SciQA~\cite{auer2023sciqa}, \textcolor{magenta}{Textbooks-QA}, VQA-RAD~\cite{lau2018vqarad}, VisualWebInstruct~\cite{tiger_lab_visualwebinstruct}\\
			\rowcolor{gray!15}
			\multirow{2}{*}{\raisebox{+3pt}{Chart \& Table}} & ChartQA~\cite{masry2022chartqa}, MMC-Inst~\cite{liu2023mmcinst}, DVQA~\cite{kafle2018dvqa}, PlotQA~\cite{methani2020plotqa}, LRV-Instruction~\cite{liu2023lrv-instruction}, TabMWP~\cite{lu2022tablemwp}, UniChart~\cite{masry2023unichart}, Vistext~\cite{tang2023vistext}, TAT-DQA~\cite{zhu2022tatdqa}, VQAonBD~\cite{VQAonDB}, FigureQA~\cite{kahou2017figureqa}, Chart2Text~\cite{kantharaj2022chart2text}, RobuT-\{Wikisql, SQA, WTQ\}~\cite{zhao2023robut}, MultiHiertt~\cite{zhao2022multihiertt}\\
			\multirow{5}{*}{\raisebox{+2.75\height}{Naive OCR}} & SynthDoG~\cite{kim2022synthdog}, MTWI~\cite{he2018icpr2018_MTWI}, LVST~\cite{sun2019lsvt}, SROIE~\cite{huang2019icdar_sroie}, FUNSD~\cite{jaume2019funsd}, Latex-Formula~\cite{oleehyo_latex_formulas}, IAM~\cite{marti2002iam}, Handwriting-Latex~\cite{aida}, ArT~\cite{chng2019art}, CTW~\cite{yuan2019ctw}, ReCTs~\cite{zhang2019rects}, COCO-Text~\cite{veit2016cocotext}, SVRD~\cite{yu2023icdar_svrd}, Hiertext~\cite{long2023icdar_hiertext}, RoadText~\cite{tom2023icdar_roadtext}, MapText~\cite{li2024icdar_maptext}, CAPTCHA~\cite{captcha}, Est-VQA~\cite{wang2020estvqa}, HME-100K~\cite{tal}, TAL-OCR-ENG~\cite{tal}, TAL-HW-MATH~\cite{tal}, IMGUR5K~\cite{krishnan2023textstylebrush_Imgur5K}, ORAND-CAR~\cite{diem2014icfhr_RAND_CAR}, Invoices-and-Receipts-OCR~\cite{mychen76_invoices_receipts_ocr_v1},  Chrome-Writting~\cite{mouchere2016icfhr2016_chrome_writing}, IIIT5k~\cite{mishra2012scene_iiit5k}, K12-Printing~\cite{tal}, Memotion~\cite{ramamoorthy2022memotion}, \textcolor{magenta}{Arxiv2Markdown}, Handwritten-Mathematical-Expression~\cite{Azu}, WordArt~\cite{xie2022toward_wordart}, 
			RenderedText~\cite{wendlerc_renderedtext}, Handwriting-Forms~\cite{ift_handwriting_forms}\\
			\rowcolor{gray!15}
			\multirow{4}{*}{\raisebox{+2\height}{OCR QA}} &  DocVQA~\cite{clark2017docqa}, InfoVQA~\cite{mathew2022infographicvqa}, TextVQA~\cite{singh2019textvqa}, ArxivQA~\cite{li2024multimodal_arxivQA},
			ScreencQA~\cite{hsiao2022screenqa}, DocReason~\cite{mplug_docreason25k}, Ureader~\cite{ye2023ureader}, FinanceQA~\cite{Sujet-Finance-QA-Vision-100k}, DocMatrix~\cite{laurenccon2024building_docmatrix}, A-OKVQA~\cite{schwenk2022aokvqa}, Diagram-Image-To-Text~\cite{kamizuru00_diagram_image_to_text}, MapQA~\cite{chang2022mapqa}, OCRVQA~\cite{mishra2019ocrvqa}, ST-VQA~\cite{biten2019stvqa}, SlideVQA~\cite{tanaka2023slidevqa}, PDF-VQA~\cite{ding2023PDFvqa}, SQuAD-VQA, VQA-CD~\cite{mahamoud2024chic_vqa_cd}, Block-Diagram~\cite{shreyanshu09_block_diagram}, MTVQA~\cite{tang2024mtvqa}, ColPali~\cite{faysse2024colpali}, BenthamQA~\cite{mathew2021asking_benthamqa}\\
			Grounding \& Counting & TallyQA~\cite{acharya2019tallyqa}, OODVQA~\cite{tu2023many}, RefCOCO/+/g (en)~\cite{yu2016refcoco,mao2016refcocog}, GroundUI~\cite{zheng2024agentstudio_groundui}\\
			\rowcolor{gray!15}
			\multirow{5}{*}{\raisebox{+2.75\height}{General VQA}} & LLaVA-150K~\cite{liu2023llava}, LVIS-Instruct4V~\cite{wang2023lvisinstruct4v}, ALLaVA~\cite{chen2024allava},  Laion-GPT4V~\cite{laion_gpt4v_dataset}, LLAVAR~\cite{zhang2023llavar}, SketchyVQA~\cite{tu2023many}, VizWiz~\cite{gurari2018vizwiz}, IDK~\cite{cha2024visually}, AlfworldGPT, LNQA~\cite{pont2020connecting_lnqa}, Face-Emotion~\cite{fastjob_visual_emotional_analysis}, SpatialSense~\cite{yang2019spatialsense}, Indoor-QA~\cite{keremberke_indoor_scene_classification}, Places365~\cite{zhou2017places365}, MMinstruct~\cite{liu2024mminstruct}, DriveLM~\cite{sima2023drivelm}, YesBut~\cite{nandy2024yesbut}, WildVision~\cite{lu2024wildvision}, LLaVA-Critic-113k~\cite{xiong2024llava_critic}, RLAIF-V~\cite{yu2024rlaif_v}, VQAv2~\cite{goyal2017vqav2}, MMRA~\cite{wu2024mmra}, KONIQ~\cite{hosu2020koniq}, MMDU~\cite{liu2024mmdu}, Spot-The-Diff~\cite{jhamtani2018learning_spotthediff}, Hateful-Memes~\cite{kiela2020hateful_memes}, COCO-QA~\cite{ren2015exploring_cocoqa}, NLVR~\cite{suhr2017corpus_nlvr2}, Mimic-CGD~\cite{laurenccon2024matters_mimic_cgd}, Datikz~\cite{belouadi2023automatikz_datikz},
			Chinese-Meme~\cite{emo_visual_data_chinese_meme}, IconQA~\cite{lu2021iconqa}, Websight~\cite{laurenccon2024unlocking_websight}\\
			\multirow{3}{*}{\raisebox{+1.3\height}{Text-only}} & Orca~\cite{lian2023openorca}, Orca-Math~\cite{mitra2024orca}, OpenCodeInterpreter~\cite{zheng2024opencodeinterpreter}
			MathInstruct~\cite{yue2023mammoth_mathinstruct}, WizardLM~\cite{xu2023wizardlm}, TheoremQA~\cite{chen2023theoremqa}, OpenHermes2.5~\cite{OpenHermes2_5}, NuminaMath-CoT~\cite{numina_math_datasets}, Python-Code-25k~\cite{flytech_python_codes_25k}, Infinity-Instruct~\cite{baai_infinity_instruct},
			Python-Code-Instructions-18k-Alpaca~\cite{iamtarun_python_code_instructions_18k_alpaca}, Ruozhiba~\cite{looksjuicy_ruozhiba}, InfinityMATH~\cite{zhang2024infinitymath}, StepDPO~\cite{lai2024stepDPO}, TableLLM~\cite{zhang2024tablellm}, UltraInteract-sft~\cite{yuan2024advancing_ultrainteract}\\
			\end{tabular}
		}
		\vspace{-1mm}
		\caption{Summary of the collected Eagle 2 SFT datasets}
		\label{tab:data_sft}
		\vspace{-1mm}
	\end{subtable}
	
	\begin{subtable}{\textwidth}
		\resizebox{\textwidth}{!}{
		\begin{tabular}{c|p{17cm}}
			\textbf{Category} & \textbf{Dataset}\\
			\midrule
			Captioning \& Knowledge &  CC3M~\cite{sharma2018cc3m}, TextCaps~\cite{sidorov2020textcaps}, ShareGPT-4V~\cite{chen2023sharegpt4v}, DenseFusion-1M~\cite{li2024densefusion}\\
			\rowcolor{gray!15}
			Grounding \& Counting & Object 365~\cite{shao2019objects365}\\
			Text-only & OpenMathInstruct~\cite{toshniwal2024openmathinstruct}\\
		\end{tabular}}
		\vspace{-1mm}
		\caption{Summary of the additional Stage 1.5 datasets}
		\label{tab:dataset_1_5}
		\vspace{-1mm}
	\end{subtable}
	\caption{Dataset used in Eagle 2. \textcolor{magenta}{Dataset in Magenta} is internal data.}
	\label{tab:dataset}
\end{table*}
\begin{figure}[t!]
    \centering
    \includegraphics[width=\linewidth]{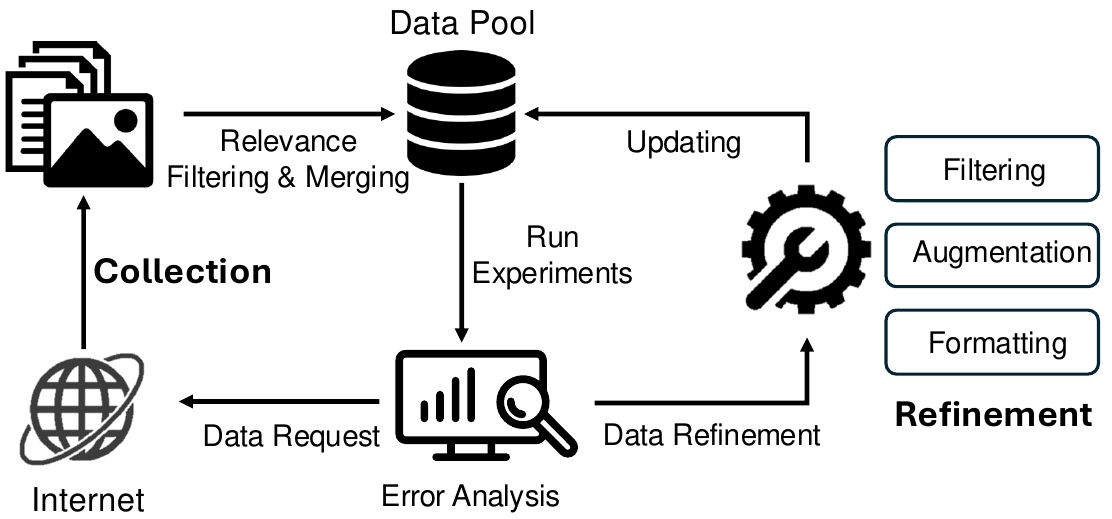}
    \caption{An overview of our data strategy. The upper part shows the date collection pipeline and the lower part shows the data refinement pipeline.}
    \label{fig:data_strategy}
\end{figure}

\subsection{Data Strategy}
Training data is essential for defining a VLM’s capabilities. However, most commercial VLMs and leading VLMs with publicly available weights keep their data strategies confidential. In this work, we conducted an in-depth exploration to create a diverse and high-quality dataset with a series of data strategies to iteratively refine and optimize our data pool. The resulting dataset significantly boosts model performance,  far surpassing the initial baseline.
Fig.~\ref{fig:data_strategy} illustrates our overall data strategy consisting of two core components: data collection and optimizing existing data.  More technical details have been provided in the appendix.

\noindent\textbf{Data collection - diversity is the key.}
A model's capability is strongly correlated with the diversity of data. As such, collecting data as diverse as possible is a key principle of this work, leading to two main strategies:
\begin{itemize}[leftmargin=3mm]
    \item \textit{Passive gathering:} Monitoring the latest related datasets from arXiv manuscripts and \href{https://huggingface.co/docs/datasets/en/index}{HuggingFace Datasets} and adding them into our candidate list.
    \item \textit{Proactive searching:} Addressing the bucket effect. As shown in Fig.~\ref{fig:data_strategy}, for each update of the data pool, we generate error analysis to identify model weaknesses and perform targeted searches for new data.
\end{itemize}

\noindent Our diverse data sources are summarized in Tab.~\ref{tab:data_sft} and generally publicly available. We utilize some pre-organized dataset collections~\cite{tong2024cambrian, li2024llavaonevision, laurenccon2024matters_mimic_cgd} to speed up preparation but also conducted careful inspection to prevent issues like test data leakage\footnote{The test split of AI2D is used in Cambrian-1 training data.}.
We also collect a large amount of public non-QA data, such as Google Landmark~\cite{weyand2020googlelandmark}, and convert them into VQA data using specific rules or auto-labeling tools.

To reduce training costs, we avoid performing ablation for each dataset individually. Instead, datasets with similar domains are added in batches to the data pool when meeting the following criteria:
\begin{itemize}[leftmargin=3mm]
\item \textit{Maintaining overall accuracy without noticeable regression for every considered benchmark.}
\item \textit{Introducing meaningful diversity to the current domains.}
\end{itemize}
To help quantify the diversity, we define a metric called \textit{Similarity Score} to measure the relevance between a new data source and the current data pool as follows:
\vspace{-1mm}
\begin{equation}
S_k = \frac{1}{N} \sum_{i=1}^{N} \max_{1 \leq j \leq M_k} \left( \text{Sim}(I_i, I_j) \times \text{Sim}(T_i, T_j) \right),
\end{equation}

\noindent where $i$ is the index of a new data source with $N$ samples, and $j$ is the index of the existing pool with $M$ samples, with $k$ denoting the data category. We compute similarity scores only within the same category, as inter-category similarity is generally low. Image embeddings $I_i$ and $I_j$ are generated from SSCD~\cite{pizzi2022sscd}, and text embeddings $T_i$ and $T_j$ from all-mpnet-base-v2~\cite{all_mpnet_base_v2}. The similarity score between samples is the product of image and text similarity. This metric shows most sources have low similarity, with a few high-similarity samples removed as duplicates.

\begin{figure}[t]
    \centering
    \includegraphics[width=\linewidth]{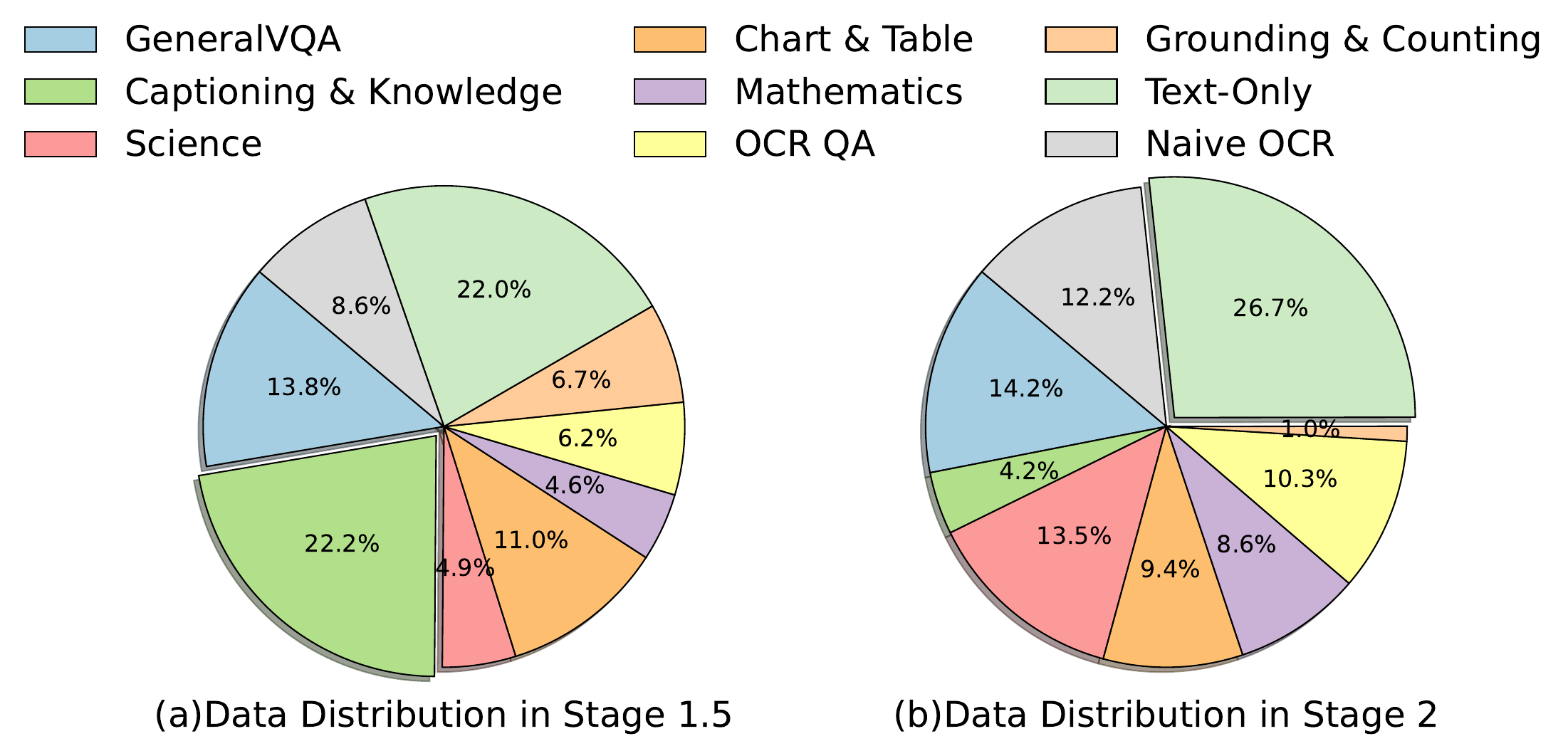}
    \caption{\footnotesize Data Distribution for Stage-1.5 and Stage-2.}
    \vspace{-4mm}
    \label{fig:data_dist}
\end{figure}

Following our data collection protocol and the refinement steps stated below, our final model uses 21.6 M samples in Stage-1.5 and 4.6 M samples in Stage-2, with the distribution illustrated in  Fig.~\ref{fig:data_dist}. We make sure text-only data occupy over 20\%. Captioning data account for the largest proportion in Stage-1.5; however, in Stage-2, we reduce its share primarily due to concerns about the overly monotonous instructions.

\begin{figure}[t!]
    \centering
    \includegraphics[width=\linewidth]{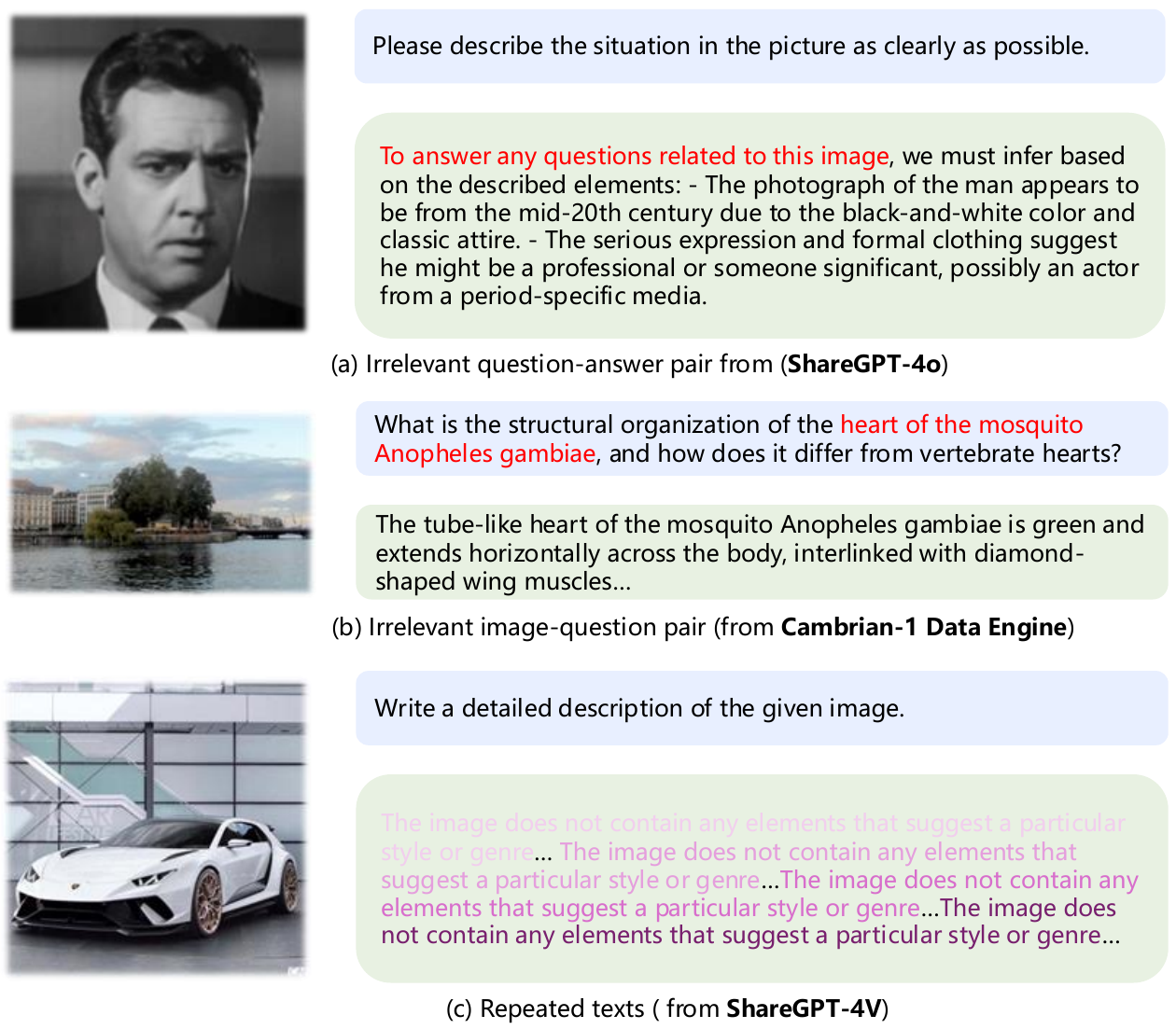}
    \caption{Typical low-quality samples.}
    \label{fig:low_quality_data}
\end{figure}

\begin{figure}[t!]
    \centering
    \includegraphics[width=\linewidth]{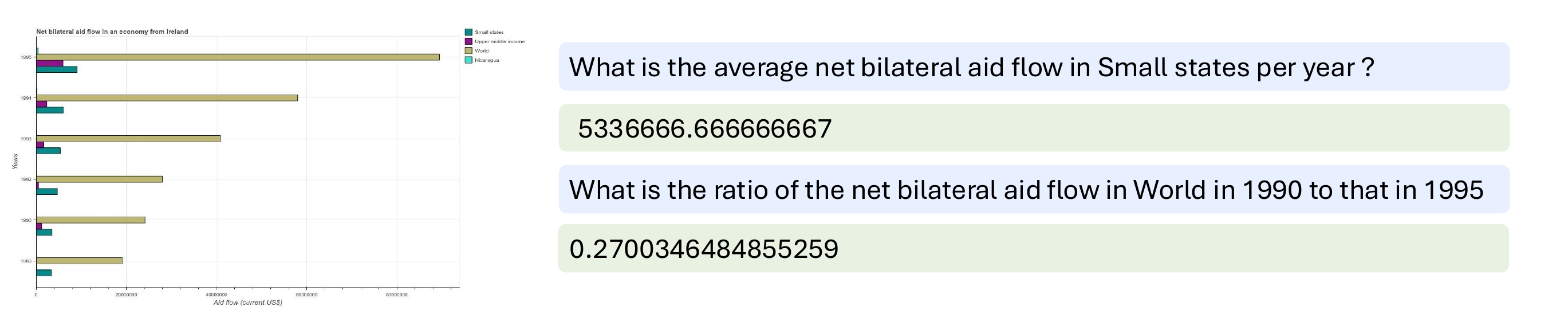}
    \caption{Public datasets~\cite{methani2020plotqa, liu2023mmcinst} often do not rigorously handle numerical precision, resulting in high decimal precision impossible to directly extract from the image.}
    \label{fig:long_digit_number}
\end{figure}

\noindent\textbf{Data filtering - ``A rotten apple spoils the barrel.''}
Public datasets often contain many low-quality samples.
We find that most low-quality cases belong to the following categories, which we use as our filtering criteria:
\begin{itemize}[leftmargin=3mm]
\item \textit{Mismatching question-answer pair.} E.g., Fig.~\ref{fig:low_quality_data} (a) from ShareGPT4o~\cite{opengvlab_sharegpt4o_dataset}.
\item \textit{Irrelevant image-question pair with unrelated image and question.} E.g., Fig.~\ref{fig:low_quality_data} (b) from Cambrian-1~\cite{tong2024cambrian}.
\item  \textit{Repeated texts.} E.g., Fig.~\ref{fig:low_quality_data} (c) from ShareGPT-4V~\cite{chen2023sharegpt4v}.
\item \textit{Numeric formatting issue.} Excessive decimal precision or overly precise numerical answers lacking corresponding information in the image. E.g., Fig.~\ref{fig:long_digit_number}.
\end{itemize}

\noindent Since most low-quality data are generated from synthesis, they often present characteristics making them distinguishable for removal through rule-based filtering.

\noindent\textbf{Subset selection - ``every data comes with a price''.}\label{sec:subset_selection}
Selecting optimal subsets is key to high-quality training. Recent work~\cite{tong2024cambrian} suggests limiting the number of samples from each source to be typically no more than $K$ (e.g. 350K). Our data selection adopts on two main principles:

\begin{itemize}[leftmargin=3mm]
\item\textit{Subset quantity determination}. Data source diversity and distribution determine the sample quantity. Auto-labeled sources are featured by larger sizes, but often contain errors and lack diversity. Instead, manually labeled datasets are often smaller. Thus, datasets with larger original sizes are generally applied with smaller sampling ratios. In our Stage-2 data, the average size per source is around 20K, with the largest subset VisualWebInstruct~\cite{tiger_lab_visualwebinstruct} having 263K samples.

\item\textit{K-means clustering selection}. Once the subset size is determined, the next step is to select the samples. Current methods often use random selection, which is suboptimal. For example, in chart data, histogram samples are more frequent than other types like line charts or pie charts, and random sampling wouldn't ensure balance across these types. To address this, we applied unsupervised K-means clustering on SSCD~\cite{pizzi2022sscd} image embeddings, where samples with similar chart types are clustered closer, allowing for target data selection, such as including all the line and pie chart samples as needed. While K-means using SSCD image embeddings performs poorly on natural scene images, it excels with mathematical, medical, and document-based data.
\end{itemize}

\begin{figure}[t!]
    \centering
    \includegraphics[width=\linewidth]{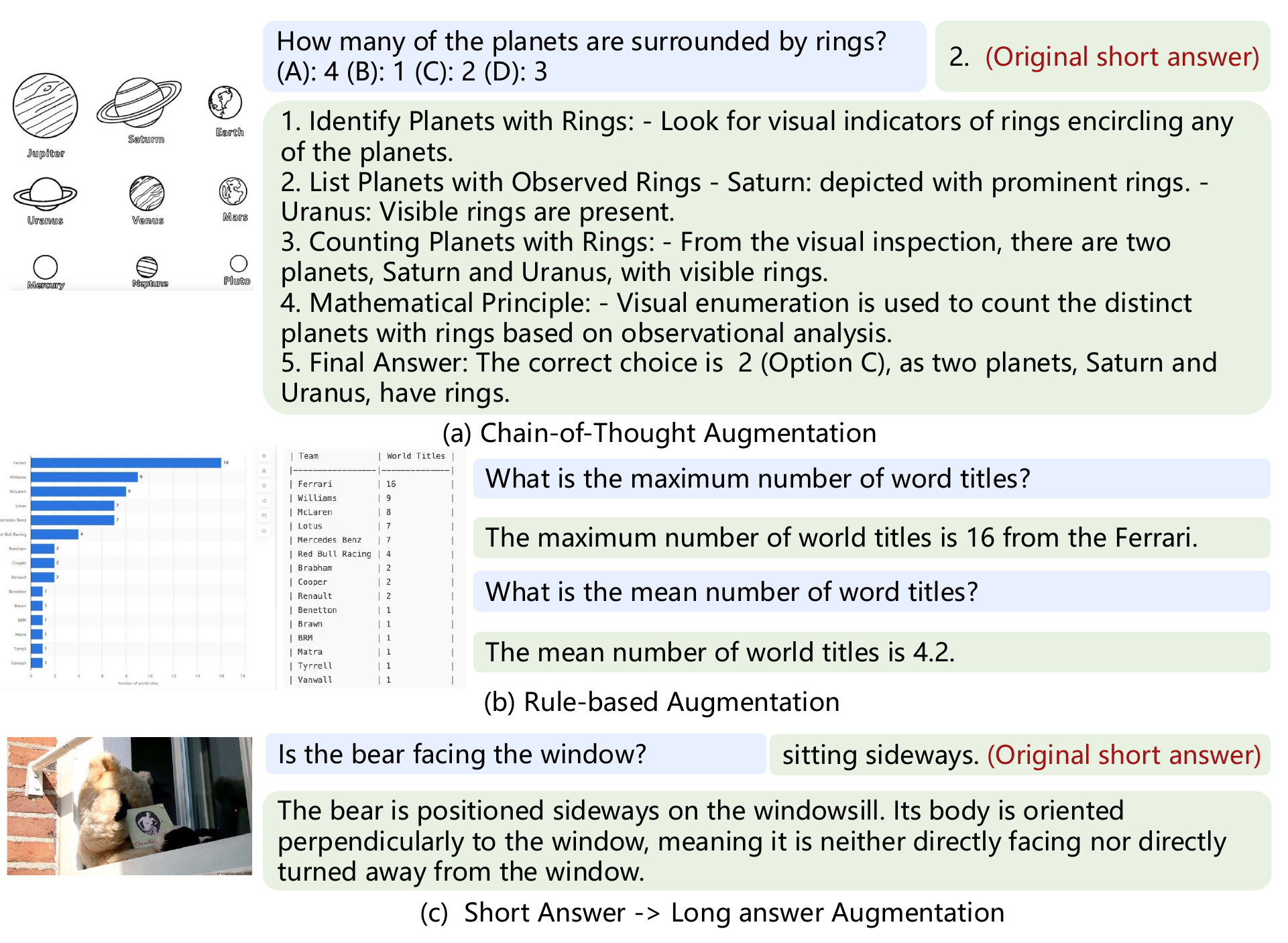}
    \caption{Our three most commonly used data augmentation methods.These methods rely on rule-based approaches or utilize VLM models for automatic labeling.}
    \label{fig:data_aug}
\end{figure}

\noindent\textbf{Data augmentation.}
Data augmentation aims to mine the rich information from input images that is not fully present in the existing QA annotations. In order to mine the potentially useful information from image space, a common approach is to use third-party VLMs to generate fine-grained descriptions of the images. We adopt this approach, as illustrated in Fig.~\ref{fig:data_aug}.
\begin{itemize}[leftmargin=3mm]
\item \textit{Adding CoT (Chain-of-Thought) explanations.}
\item \textit{Rule-based QA generation.}
\item \textit{Expanding short answers into longer responses.}
\end{itemize}
For details of the above generation process, kindly refer to the supplementary.

\begin{table}[t]
\centering
\resizebox{\linewidth}{!}{
\begin{tabular}{m{0.2\textwidth}|m{\linewidth}}
\textbf{Formular Image} & \textbf{\LaTeX\, Annotation} \\
\hline
\includegraphics[width=0.2\textwidth, height=1cm]{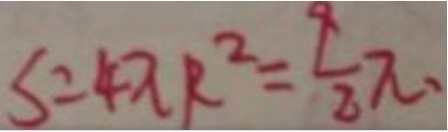}& \texttt{S=4\textbackslash piR\^ \,2=\textbackslash\,frac\{9\}\{2\}\textbackslash\,pi}  \\
\hline
\includegraphics[width=0.1\textwidth]{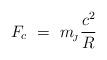} & \textcolor{red}{\textbackslash begin\{align*\}}F\_c=m\_\{J\}\textbackslash frac\{c\^\,2\}\{R\}\textcolor{red}{\textbackslash end\{align*\}}
\end{tabular}} 
\caption{Two samples with same "Extract \LaTeX\, from image" task but with different format.}
\label{tab:format}
\end{table}

\noindent\textbf{Data formatting.}
Transforming data into the correct format is also a crucial step in data preparation. One of our basic principles is: \textit{``same task, similar format; different tasks, clearly distinct formats.''} Our data formatting includes but is not limited to:
\begin{itemize}[leftmargin=3mm]
    \item \textit{Removing unnecessary decorations.} We present a seemingly inconspicuous example that, nonetheless, has a profound impact on the final results in Tab.~\ref{tab:format}. We show two samples from different sources for the task of extracting LaTeX formulas from images. The second sample includes an unnecessary fixed equation environment. Even with limited use of such data, the model consistently outputs this fixed template (in red texts). After removing the fixed equation environment, the model returns to normal behavior, leading to a significant improvement on OCRBench~\cite{liu2023ocrbench}.
    \item \textit{Appending more specific instructions.} Adding detailed instructions to the original instruction based on the response is a common approach. For example, appending ``Provide a short answer'' to brief responses helps prevent a model from becoming an ``answering machine'' that is used to giving short answers. However, overusing this can also hinder generalization. For instance, adding ``Please answer yes or no'' to every yes-or-no question may impair the model's ability to answer correctly without such prompt during inference.
\end{itemize}

\subsection{Training Recipe}
Our data strategy enables us to build a high-quality and diverse dataset, but applying different training recipes to the same data pool still has a decisive impact on the final results. Our recipe is built upon the following core points.

	\begin{table}[t]
    \centering
    \setlength{\tabcolsep}{2pt}
    \renewcommand{\arraystretch}{1.2}
    \resizebox{\linewidth}{!}{%
    \begin{tabular}{@{}ll|ccc@{}}
    \toprule
    & & \textbf{Stage-1} & \textbf{Stage-1.5} & \textbf{Stage-2} \\ 
    \midrule 
    \multirow{2}{*}{Vision}
    & \textbf{Resolution} & \multicolumn{3}{c}{$\{\text{448}_{\text{SigLIP}}, 512_{\text{ConvNeXt*}}\}\times \{(i, j) \mid i, j \in \mathbb{Z}^+, \, i \times j \leq 12 \}$}\\
    & Tokens & \multicolumn{3}{c}{$ (i\times j+1) \times 256$}\\
    \midrule 
    \multirow{2}{*}{Data}
    & \textbf{Dataset} & ALLaVA &  Rich Diverse Data  &  High-Quality Data
    \\
    & \#Samples & 1.2M & 21.6M & 4.6M \\
    \midrule
   \multirow{4}{*}{Model}
   & \textbf{Trainable} & MLP Connector & Full Model & Full Model \\
    & Qwen2.5-0.5B & 4.9M & 0.9B & 0.9B  \\
    & Qwen2.5-1.5B & 9.4M & 2.0B & 2.0B  \\
    & Qwen2.5-7B & 40.0M & 8.9B & 8.9B \\ 
     \midrule 
    \multirow{3}{*}{Training}
    & \textbf{Batch Size} & 1024 & 1024 & 256 \\
    & \textbf{Learning Rate} & 2$\times 10^{-4}$ & \{2, 4\} $\times 10^{-5}$ & \{2, 4\} $\times 10^{-5}$ \\ 
     & \textbf{Max Length} & 4096& 8192 & 16384\\ 
    \bottomrule
    \end{tabular}}
    \caption{We present our three-stage training settings, where Eagle2-9B/2B/1B builds upon Qwen2.5-32B/7B/1.5B/0.5B~\cite{qwen2.5}, respectively. *: For small scale model with 0.5/1.5B LLM, we only use SigLIP as visual encoder and learning rate of $4\!\times\!10^{-5}$ in Stage-1.5 \& 2. }
    \label{tab:training_strategy}
    \end{table}
	\noindent\textbf{Post-pretraining stage is necessary}. We initially begin with LLaVA~\cite{liu2023llava}'s two-stage training strategy, where we train an MLP connector followed by full model training with SFT data. While efficient, this approach proved unsuitable for quick SFT data updates, as the expanding SFT data makes it harder to track the impact of new data and reduces the experimental efficiency. For instance, we observe improvements from expanding the Cambrian-1~\cite{tong2024cambrian} SFT data. However, the gap remains between the model and state-of-the-art ones. Considering that the main limitation of the two-stage strategy is the lack of robust pre-training, we add an additional pre-training stage (Stage-1.5). Stage-1.5 pre-trains the model on a larger dataset to reduce dependency on SFT data in subsequent training.
    
    \begin{figure}[t]
    \centering
    \includegraphics[width=0.96\linewidth]{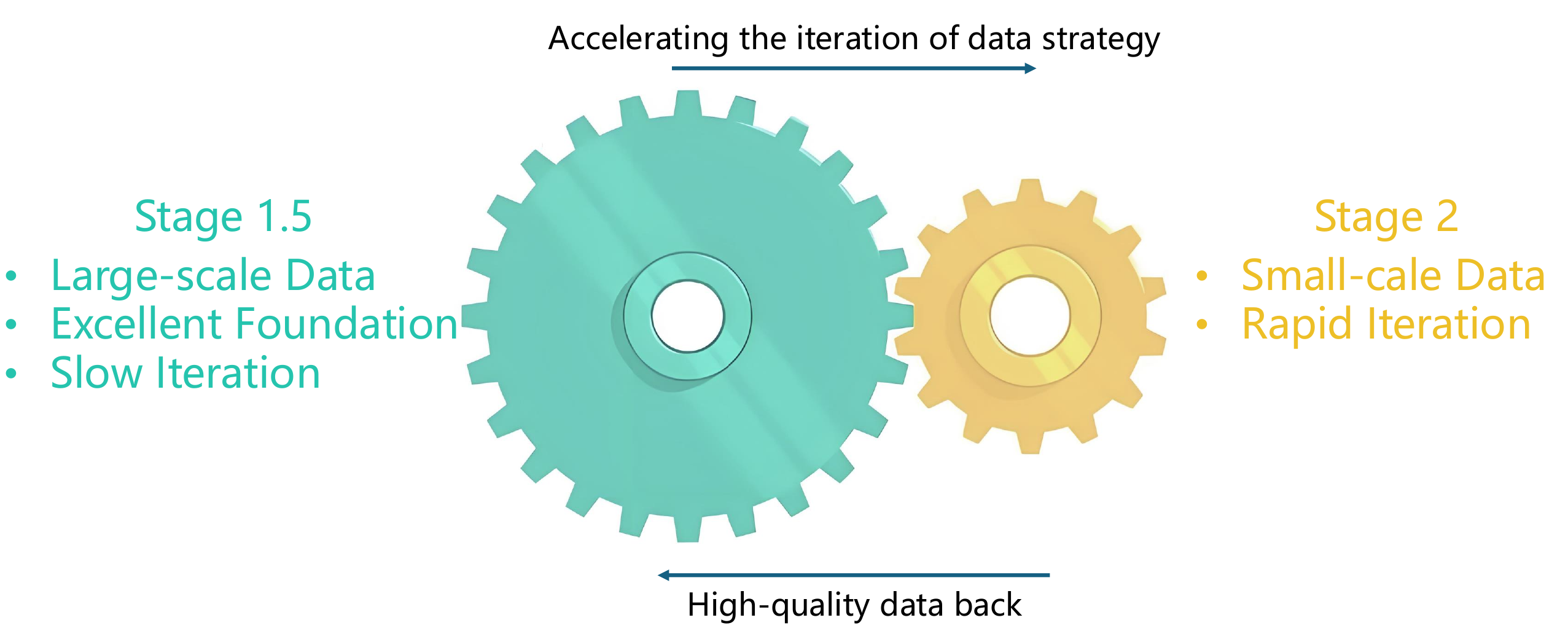} 
    \caption{Stage-1.5 accelerates Stage-2 and Stage-2 gives Stage-1.5 feedback.}
    \label{flywheel}
\end{figure}

    \noindent\textbf{``Large wheel drives small wheel.''} Three-stage pre-training is, in fact, widely used in existing works, such as LLaVA-OneVision~\cite{li2024llavaonevision}. However, we have a distinctly different view to the data that using in Stage-1.5. Other works tend to use more knowledge-related data, such as captioning data, at this stage. In this work, we add all data sources intended for visual instruction to Stage-1.5, simultaneously introducing several other datasets as shown in Tab.~\ref{tab:dataset_1_5}. As shown in Fig.~\ref{flywheel}, training Stage-2 based on Stage-1.5 enables rapid iteration on a high-performance foundation. The derived conclusions are more robust than those obtained from ungeneralizable ablation experiments on toy-scale data. In addition, the effective conclusions obtained from Stage-2 can be used to update Stage-1.5, further driving improvements in model performance. Detailed settings are shown in Tab.~\ref{tab:training_strategy}.
    
    \begin{figure}[tbh]
    \centering
    \begin{subfigure}{0.49\linewidth}
        \centering
        \includegraphics[width=\linewidth]{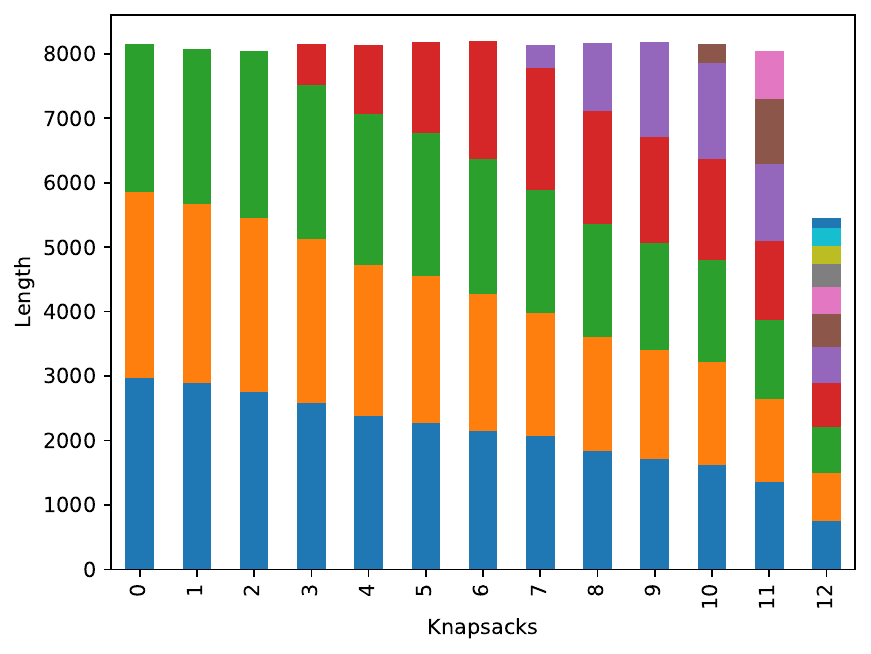} %
        \caption{Knapsacks of naive greedy knapsack method.}
    \end{subfigure}~
    \begin{subfigure}{0.49\linewidth}
        \centering
        \includegraphics[width=\linewidth]{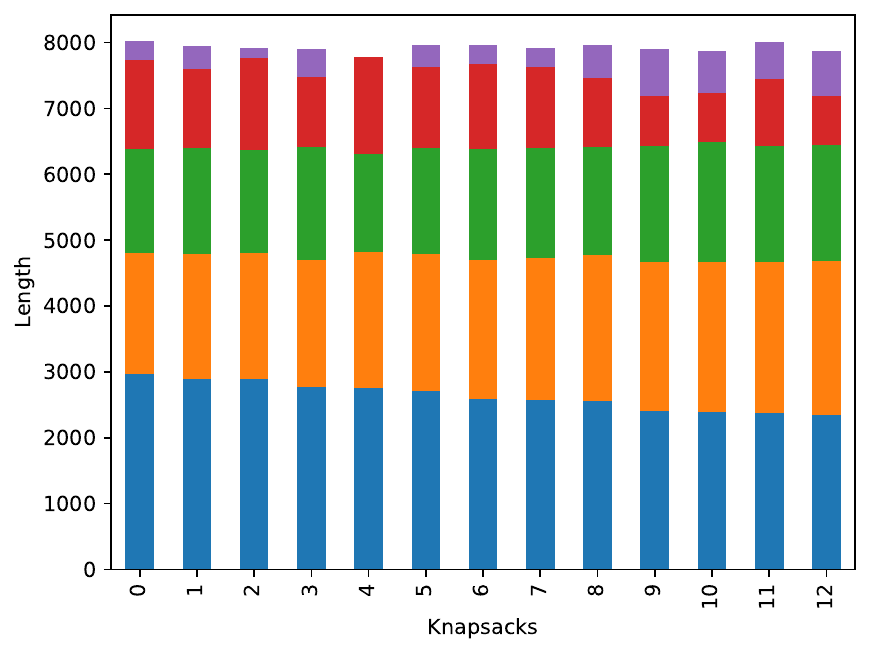} 
        \caption{Knapsacks of balanced knapsack method.}
    \end{subfigure}
 
    \caption{We pack 64 samples of varying lengths into combined samples, each with a length of less than 8192. The naive greedy knapsack approach in LLaMa-Factory~\cite{zheng2024llamafactory} leads to uneven length distributions (left), while the proposed method gives more balanced length distributions within every knapsack (right).}
    \label{pack}
 
\end{figure}     
    \noindent \textbf{Balance-aware data packing matters.} Data packing speeds up training by concatenating shorter samples, reducing padding use. In our experiments, using packing accelerate the training by 2-3 times.
    A key step in packing is arranging $N$ short samples of varying lengths into $M$ long samples without exceeding max length. Existing frameworks such as LLaMa-Factory use a naive greedy knapsack algorithm\footnote{ \scriptsize\url{https://github.com/hiyouga/LLaMA-Factory/blob/main/src/llamafactory/data/processors/processor_utils.py}}, but this often produces packs with uneven length distributions. As shown in Fig.~\ref{pack}, the naive greedy knapsack method groups long and short samples separately, which is not desirable to model training.

Therefore, we design a balance-aware greedy knapsack algorithm that creates packs with a more uniform length distribution, as shown in Fig.~\ref{fig:greedy_knapsack}, ensuring that each pack contains both long and short samples. Unlike SPFHP (Shortest-Pack-First Histogram Packing)~\cite{krell2021efficient}, our method prioritizes balanced length distribution over packing efficiency, helping balance loss weights between long and short samples. Further details are in the appendix.

\begin{figure}[t]
\footnotesize 
\begin{lstlisting}
# Our proposed greedy knapsack method
def balanced_greedy_knapsack(samples, L):
    # Step 1: Sort the samples
    samples.sort(reverse=True)
    total_length = sum(samples)
    min_knapsacks = (total_length + L - 1) // L 
    # Step 2: Initialize knapsacks
    knapsacks=[[] for _ in range(min_knapsacks)]
    knapsack_lengths = [0] * min_knapsacks
    # Step 3: Distribute samples across knapsacks
    ks_index = 0
    sample_index = 0
    while sample_index < len(samples):
        length = samples[sample_index]
        if knapsack_lengths[ks_index]+length<=L:
            knapsacks[ks_index].append(length)
            knapsack_lengths[ks_index] += length
            sample_index += 1
        else:
            knapsacks.append([])
            knapsack_lengths.append(0)
        ks_index = argmin(knapsack_lengths)

    return knapsacks
\end{lstlisting}
\caption{Python code for the proposed balance-aware greedy knapsack method. $L$ is the max length and ``samples" is a list of token lengths.}
\label{fig:greedy_knapsack}
\vspace{-1mm}
\end{figure}

\subsection{Tiled Mixture of Vision Encoders}

Following Eagle~\cite{eagle}, we use SigLIP~\cite{zhai2023siglip} and ConvNeXt-XXLarge~\cite{liu2022convnet, cherti2023reproducible} as vision encoders. Additionally, to handle arbitrarily high-resolution images, we employ image tiling following InternVL-1.5~\cite{chen2024far}. The input resolution of every image tile of SigLIP is $448\!\times\!448$, while the input size of ConvNeXt is $512\!\times\!512$. To make sure they output same number of image tokens, we use PixelShuffle to conduct a 2$\times$ downsampling on the image features from SigLIP, resulting a feature shape of $16\!\times\!16$, matching the output size of ConvNeXt (32$\!\times\!$ downsampling of input). We then concatenate these features along the channel dimension and align with LLM via an MLP layer.

\section{Experiments}

\begin{table*}[t!]
\centering
\setlength\tabcolsep{1.5pt}
\renewcommand{\arraystretch}{1.2}
\resizebox{\textwidth}{!}{
\begin{tabular}{l|cc|ccccccccccccc|c}
\hline
\multirow{2}{*}{Model} & \multirow{2}{*}{Stage-1} & \multirow{2}{*}{Stage-2} & DocVQA & ChartQA & InfoVQA & TextVQA & OCRBench & MMstar & RWQA & AI2D & MMMU & MMB$_{1.1}$ & MMVet & HallB & MathVista & Average \\
                       &                          &                          & Test   & Test    & Test    & Val     & Test    & Test   & Test & Test & Val  & EN-Val       & Test   & Test  & Test-Mini & score   \\
\hline
Cambrian-7B~\cite{tong2024cambrian} & 1.2M & 7M  & 77.8 & 73.3 & -    & 71.7 & 624 & 50.7 & 64.2 & 74.6 & 42.7 & -     & 48.0 & 30.6 & 49.0 & -    \\
\hline
Eagle2-Baseline            & 1.2M & 5.2M & 77   & 65.9 & 50   & 69.9 & 589 & 47.3 & 60.3 & 74.7 & 49.7 & 71.2  & 45.9 & 44.8 & 49.2 & 58.8 \\
+ Naive OCR                        & 1.2M & +0.8M & 78   & 67.0 & 57   & 73.7 & 719 & 49.3 & 59.9 & 74.4 & 50.7 & 72.1  & 45.5 & 46.3 & 50.7 & 61.3 \\
+ Chart, Table \& OCR QA           & 1.2M & +2.0M & 88   & 78.3 & 65   & 77.6 & 783 & 51.7 & 62.7 & 76.2 & 50.1 & 72.7  & 50.1 & 39.9 & 54.1 & 65.0 \\
+ Science \& Math                  & 1.2M & +0.7M & 89   & 78.8 & 64   & 77.7 & 777 & 52.5 & 63.4 & 77.1 & 50.7 & 75.6  & 51.5 & 38.9 & 51.2 & 65.2 \\
+ Caption                          & 1.2M & +0.5M & 88   & 79.0 & 63   & 78.0 & 784 & 53.7 & 61.0 & 77.2 & 52.9 & 77.9  & 55.8 & 39.5 & 49.7 & 65.7 \\
+ Grounding \& Counting            & 1.2M & +0.2M & 88   & 79.4 & 64   & 77.9 & 792 & 54.3 & 61.7 & 77.1 & 51.8 & 77.6  & 54.2 & 39.6 & 53.7 & 66.0 \\
+ Text-Only                        & 1.2M & +2.3M & 88   & 78.5 & 63   & 77.9 & 792 & 55.5 & 65.4 & 76.5 & 51.4 & 76.5  & 58.8 & 37.3 & 57.5 & 66.6 \\
+ General                          & 1.2M & +1.0M & 88   & 80.0 & 63   & 77.8 & 797 & 55.9 & 65.9 & 76.9 & 52.2 & 78.9  & 56.8 & 40.8 & 55.5 & 67.0 \\
\hline
\end{tabular}}

\caption{\textbf{Data ablation under 2-Stage training.} The Stage-2 baseline uses Cambrian-1 data. In subsequent experiments, we gradually increase the SFT data, adding only specific categories each time.}
\label{tab:abl_baseline}

\end{table*}

\begin{table*}[t!]
\centering
\setlength\tabcolsep{1.5pt}
\renewcommand{\arraystretch}{1.2}
\resizebox{\textwidth}{!}{
\begin{tabular}{l|ccc|ccccccccccccc|c}
\hline
\multirow{2}{*}{Model} & \multirow{2}{*}{Stage-1} & \multirow{2}{*}{Stage-1.5} & \multirow{2}{*}{Stage-2} & DocVQA & ChartQA & InfoVQA & TextVQA & OCRBench & MMstar & RWQA & AI2D & MMMU & MMB$_{1.1}$ & MMVet & HallB & MathVista & Average \\
                       &                           &                           &                           & Test   & Test    & Test    & Val     & Test    & Test   & Test & Test & Val  & EN-Val       & Test  & Test  & Test-Mini & Score   \\
\hline
Cambrian-7B~\cite{tong2024cambrian} & 1.2M & -    & 7M     & 77.8  & 73.3  & -     & 71.7 & 624 & 50.7 & 64.2 & 74.6 & 42.7 & 68.2 & 48.0 & 30.6 & 49.0 & -     \\
\hline
Introducing Stage 1.5                         & 1.2M & 21M  & -      & 89.9  & 82.7  & 71.1  & 79.5 & 812 & 58.5 & 69.4 & 78.6 & 50.9 & 81.7 & 55.7 & 47.1 & 60.1 & 69.7  \\
+ Stage 2                           & 1.2M & 21M  & 12.7M  & 91.0  & 84.0  & 72.5  & 81.0 & 825 & 61.4 & 69.0 & 81.0 & 52.0 & 83.0 & 56.3 & 46.4 & 61.4 & 70.9  \\
+ Naive Subset Selection            & 1.2M & 21M  & 8.2M   & 90.4  & 83.7  & 72.0  & 79.7 & 798 & 62.8 & 67.8 & 83.6 & 51.7 & 81.9 & 56.1 & 46.7 & 61.0 & 70.6  \\
+ Data Formatting/Filtering         & 1.2M & 21M  & 8.2M   & 91.1  & 84.5  & 72.3  & 81.2 & 843 & 62.1 & 67.7 & 83.0 & 53.2 & 82.2 & 55.8 & 46.6 & 62.0 & 71.2  \\
+ Advanced Subset Selection          & 1.2M & 21M  & 4.6M   & 90.8  & 84.1  & 73.3  & 81.4 & 843 & 62.7 & 68.9 & 84.1 & 52.5 & 82.5 & 59.3 & 49.2 & 60.5 & 71.8  \\
+ Data Augmentation                 & 1.2M & 21M  & 4.6M   & 91.8  & 85.1  & 73.6  & 81.1 & 839 & 61.4 & 68.4 & 83.9 & 53.6 & 82.4 & 60.1 & 48.4 & 63.5 & 72.1  \\
+ Re-Update Stage 1.5               & 1.2M & 22M  & 4.6M   & 91.3  & 85.9  & 73.3  & 81.9 & 842 & 61.7 & 68.2 & 83.5 & 53.6 & 82.4 & 61.3 & 49.0 & 65.2 & 72.4  \\
+ Mixture of Vision Encoders        & 1.2M & 22M  & 4.6M   & 92.6  & 86.4  & 77.2  & 83.0 & 868 & 62.6 & 69.3 & 83.9 & 56.1 & 81.9 & 62.2 & 49.3 & 63.8 & 73.5  \\
\hline
\end{tabular}}
\caption{Employing three-stage training strategy allows us to reduce the amount of training data required in the Stage-2, thereby expediting the data iteration process. The resultant efficient data strategies can then be leveraged to refresh and optimize the data in Stage-1.5.}
\label{tab:stage_1_5}
\end{table*}

\begin{table*}[t!]
\centering
\setlength\tabcolsep{1.5pt}
\renewcommand{\arraystretch}{1.2}
\resizebox{\textwidth}{!}{
\begin{tabular}{l|ccccccccccccc|c}
\hline
\multirow{2}{*}{Model} & DocVQA & ChartQA & InfoVQA & TextVQA & OCRBench & MMstar & RWQA & AI2D & MMMU & MMB$_{1.1}$ & MMVet & HallB & MathVista & \textbf{Open-} \\
                       & Test   & Test    & Test    & Val     & Test    & Test   & Test & Test & Val  & Test        & Test  & Test  & Test-Mini & \textbf{Compass} \\
\midrule
\multicolumn{3}{l}{\textbf{\textit{Closed-Source Models}}} \\
\midrule
GPT-4o-0513~\cite{openai2023gpt4o} & 92.8 & 85.7 & -     & -    & 736 & 63.9 & 75.4 & 84.6 & 69.2 & 82.2 & 69.1 & 55.0 & 61.3 & 69.9 \\
GPT-4V~\cite{gpt4v} & 88.4 & 78.5 & 75.1 & 78.0 & 656 & 56.0 & 68.0 & 78.6 & 61.7 & 79.8 & 67.5 & 43.9 & 54.7 & 63.5 \\
Gemini-1.5-Pro~\cite{reid2024gemini1_5} & 93.1 & 87.2 & 81.0 & 78.7 & 754 & -    & 70.4 & -    & 62.2 & -    & -    & -    & 63.9 & 64.4 \\
\midrule
\multicolumn{3}{l}{\textbf{\textit{Publicly Available Models}}} \\
\midrule
LLaVa-OneVision-0.5B~\cite{li2024llavaonevision}  & 70.0 & 61.4 & 41.8 &- &565& 37.7 & 55.6 & 57.1 & 31.4 & 50.3 & 32.2 & 31.7& 33.8& 41.3 \\
InternVL2-1B~\cite{chen2024internvl2} & 81.7 & 72.9 & 50.9 & 70.0 & 754 & 45.7 & 50.3 & 64.1 & 36.7 & 59.7 & 32.7 & 34.0 & 37.7 & 48.3 \\
\rowcolor[gray]{0.9}
Eagle2-1B & 81.8 & 77.0 & 54.8 & 76.6 & 767 & 48.5 & 55.4 & 70.9 & 38.8 & 63.0 & 40.9 & 35.3 & 45.3 & 52.4* \\
\midrule
InternVL2-2B~\cite{chen2024internvl2}& 86.9 & 76.2 & 58.9 & 73.4 & 784 & 50.1 & 57.3 & 74.1 & 36.3 & 69.6 & 39.5 & 37.9 & 46.3 & 54.0\\
Qwen2-VL-2B~\cite{wang2024qwen2vl}& 90.1 & 73.0 & 65.5 & 79.7 & 809 & 48.0 & 62.6 & 78.9 & 41.1 & 72.2 & 49.5 & 41.7 & 43.0& 57.2 \\
\rowcolor[gray]{0.9}
Eagle2-2B & 88.0 & 82.3 & 65.8 & 79.1 & 818 & 56.4 & 63.1 & 79.3 & 43.1 & 74.9 & 53.8 & 45.8 & 54.7 & 61.2* \\
\midrule
InternVL2-8B~\cite{chen2024internvl2} & 91.6 & 83.3 & 74.8 & 77.4 & 794 & 60.9 & 64.4 & 83.8 & 51.8 & 79.4 & 54.2 & 45.2 & 58.3 & 64.1 \\
Qwen2-VL-7B~\cite{qwen2.5} & 94.6 & 83.0 & 74.3 & 84.3 & 845 & 60.7 & 70.1 & 83.0 & 54.1 & 81.0 & 62.0 & 50.5 & 58.2 & 67.0 \\
MiniCPM-V2.6~\cite{hu2024minicpm} & 90.8 & 82.4 & -     & 80.1 & 852 & 57.5 & 65.0 & 82.1 & 49.8 & 78.0 & 60.0 & 48.1 & 60.6 & 65.2 \\
LLaVA-One-Vision-7B~\cite{li2024llavaonevision} & 87.5 & 80.0 & 68.8 & -    & 622 & 61.7 & 66.3 & 81.4 & 48.8 & 80.9 & 57.5 & 31.6 & 63.2 & 60.1 \\
        \textcolor{gray}{InternVL2-26B}~\cite{chen2024internvl2} & \textcolor{gray}{92.9} & \textcolor{gray}{84.9} & \textcolor{gray}{75.9} & \textcolor{gray}{82.3} & \textcolor{gray}{825} & \textcolor{gray}{61.0} & \textcolor{gray}{68.3} & \textcolor{gray}{84.5} & \textcolor{gray}{50.7} & \textcolor{gray}{81.2} & \textcolor{gray}{62.1} & \textcolor{gray}{50.7} & \textcolor{gray}{59.4} & \textcolor{gray}{66.4}  \\
        \textcolor{gray}{LLaVA-One-Vision-72B}~\cite{li2024llavaonevision} & \textcolor{gray}{91.7} & \textcolor{gray}{83.7} & \textcolor{gray}{74.9} & \textcolor{gray}{-} & \textcolor{gray}{741} & \textcolor{gray}{66.1} & \textcolor{gray}{71.9} & \textcolor{gray}{85.6} & \textcolor{gray}{56.6} & \textcolor{gray}{84.5} & \textcolor{gray}{60.6} & \textcolor{gray}{47.5} & \textcolor{gray}{68.4} & \textcolor{gray}{68.0} \\
        \textcolor{gray}{LLaMa-3.2-90B-Vision}~\cite{dubey2024llama3} & \textcolor{gray}{90.1} & \textcolor{gray}{85.5} & \textcolor{gray}{-} & \textcolor{gray}{-} & \textcolor{gray}{783} & \textcolor{gray}{55.3} & \textcolor{gray}{-} & \textcolor{gray}{-} & \textcolor{gray}{60.3} & \textcolor{gray}{77.3} & \textcolor{gray}{64.1} & \textcolor{gray}{44.1} & \textcolor{gray}{57.3} & \textcolor{gray}{63.4} \\

\rowcolor[gray]{0.9}
Eagle2-9B & 92.6 & 86.4 & 77.2 & 83.0 & 868 & 62.6 & 69.3 & 83.9 & 56.1 & 80.6 & 62.2 & 49.3 & 63.8 & 68.2* \\

\hline
\end{tabular}}

\caption{\textbf{Comparison with SoTA models on Various Benchmarks}. *: We obtain the OpenCompass~\cite{opencompass2023} score by averaging across Eagle benchmarks (OCRBench, MMStar, AI2D, MMMU, MMB$_{1.1}$, MMVet, HallusionBench, and MathVista).}
\label{tab:compare_with_sota}

\end{table*}

\subsection{Evolution of Eagle 2}

\noindent\textbf{Scaling Stage-2 training data.} We initially explore the impact of scaling Stage-2 data, as shown in Tab.~\ref{tab:abl_baseline}. Our findings reveal that model’s overall performance improved steadily with additional data, with the most notable gains arising from the inclusion of 2M (million) VQA samples focused on charts, tables, and OCR. While data scaling indicates potential for further gains beyond 10M samples, our experiments' costs have risen sharply, and the efficiency of data iteration has decreased. Moreover, we observe considerable performance fluctuations across specific benchmarks at this scale, especially in challenging benchmarks like MMMU, MathVista, and MMVet. Another obstacle is that, as illustrated by the data-performance growth trend in Fig.\ref{fig:line_abl}, reaching the performance of frontier VLMs like Qwen2-VL would be difficult. These challenges leads us to consider adopting a more effective training strategy.

\begin{figure}[t]
    \centering
    \includegraphics[width=\linewidth]{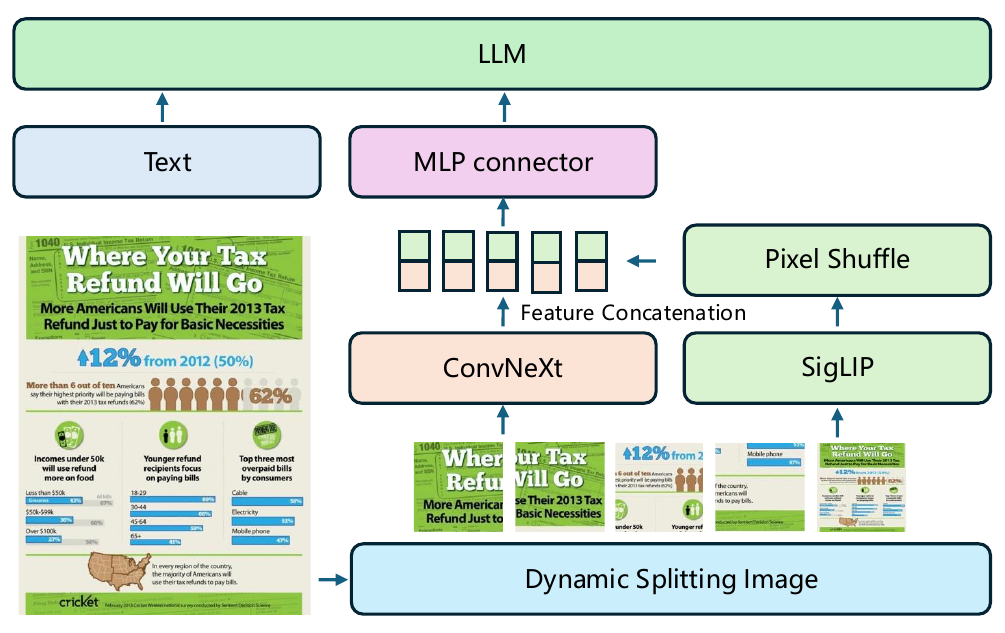}
    \caption{Tiled Mixture of Vision Encoders.}
    \label{fig:model_arch}
\end{figure}

\noindent\textbf{Introducing Stage-1.5.} To build a robust pre-trained model, we implement Stage-1.5 where we focus on maximizing the data utilization to strengthen the model's foundational capabilities. As shown in Tab.~\ref{tab:stage_1_5}, the Stage-1.5 checkpoint is competitive by itself, and subsequent Stage-2 training further improves the previous best model's performance by average $3.9\%$.

\noindent\textbf{Naive data selection.}
Using a naive data selection strategy with maximum thresholds and random sampling, we reduce the training data to 8.6M; unfortunately, this led to a decline in performance. We speculate it might be that the randomly selected data have inadvertently excluded some valuable samples, while also failing to adequately ensure a balanced data distribution.

\noindent\textbf{Data formatting \& filtering.}
After filtering low-quality data and formatting the training set, we see clear improvements on 8 out of 14 benchmarks, including a remarkable 45-point gain on OCRBench~\cite{liu2023ocrbench}. This implies the importance of carefully reviewing and utilizing existing data, as well as exercising caution with publicly available data.

\noindent\textbf{Advanced data selection.}
By employing the comprehensive data selection strategy introduced in Sec.~\ref{sec:subset_selection}, we further reduce the dataset to 4.6M samples. By selecting a more balanced data subset and ensuring data quality, we achieve a further improvement in average score with a reduced amount of data.

\noindent\textbf{Data augmentation.}
By employing our data augmentation strategy, we introduce a greater volume of automatically generated CoT training data, leading to notable performance improvements on MMMU and MathVista. The rule-based data augmentation on the chart data also brings 1 point improvement on ChartQA.

\noindent\textbf{Re-updating stage-1.5.} 
The effective data strategies we explored in Stage-2, such as data filtering, formatting, and augmentation, can be applied to update the Stage-1.5 data, thereby further enhancing the model's ultimate capability. By updating Stage-1.5 checkpoint, we obtain clear improvement on ChartQA, MMVet and MathVista.

\noindent\textbf{Mixture of vision encoders.} 
Introducing mixture of vision encoder has brought performance improvements on 12 out of 14 benchmarks, particularly in benchmarks related to documents, charts, and OCR. This clearly demonstrates that the mixture of vision encoders significantly enhances the model's understanding to visual spaces.

\subsection{Comparison with SOTA Models}

As shown in Tab.~\ref{tab:compare_with_sota}, we conducted comparisons across 14 diverse benchmarks with the representative state-of-the-art public avaiable and closed-source models.
Our Eagle2-9B, building on top of Qwen2.5-7B~\cite{qwen2.5}, outperforms InternVL2-8B~\cite{chen2024internvl2} and MiniCPM-v2.6~\cite{hu2024minicpm} across all 14 benchmarks and leads Qwen2-VL-7B~\cite{wang2024qwen2vl} in 9 out of the 14 benchmarks and beats it on OpenCompass. Eagle2-9B holds its ground against much larger VLMs such as InternVL2-26B, LLaVa-OneVision-72B~\cite{li2024llavaonevision} and LLaMa-3.2-90B-Vision~\cite{dubey2024llama3}. Apart from MMVet and MMMU, we comprehensively surpass GPT-4V. Eagle2-9B surpasses GPT-4o~\cite{openai2023gpt4o} on ChartQA, OCRBench, and MathVista, while achieving performance very close to GPT-4o on DocVQA, MMStar, AI2D and OpenCompass.

\section{Related Work}
\textbf{Vision-Language Models (VLMs)} LLMs~\cite{OpenAI_ChatGPT, ouyang2022training, chen2022pali} have transformed natural language processing (NLP) and reshaped the broader AI landscape. 
The advancement of LLMs has spurred significant progress in visual understanding by integrating visual features with LLMs, leading to the emergence of {Visual-Language Models(VLMs)}~\cite{li2023blip2,gpt4v, liu2023llava, zhu2023minigpt4}. 
The performance of VLMs with public available weights~\cite{liu2023llava, chen2022pali, wang2023cogvlm, li2023otter, liu2024llavanext, li2024llavaonevision, yao2024minicpm, chen2024internvl2, wang2024qwen2vl, beyer2024paligemma, dubey2024llama3, dai2024nvlm, lin2023vila, fang2024vila, liu2024nvila} continues to make breakthroughs, reaching or even surpassing the most advanced commercial models such as GPT-4V/4o~\cite{openai2023gpt4o, gpt4v} and Gemini-1.5~\cite{reid2024gemini1_5}. 
Fully open-source VLMs~\cite{li2024llavaonevision, tong2024cambrian, idefics2023} have released their training data and code base, further accelerating the VLM research.

\noindent\textbf{Vision-Centric VLMs.} Our work adopts a vision-centric VLM design that emphasizes strong vision foundation and HD input. This is aligned with the spirit of various related areas, including: 1) Vision foundation for VLMs~\cite{radford2021clip, openclip, sun2023evaclip} and improved designs~\cite{chen2023internvl, zhai2023siglip, beyer2024paligemma, xiao2024florence, ranzinger2024radio, heinrich2024radio}, 2) Mixture of vision encoder designs~\cite{lin2023sphinx, liu2024prismer, karamcheti2024prismatic, li2024miniGemini, luo2024llava_hr, tong2024cambrian, eagle}, and 3) Tiling and HD input designs~\cite{chen2023palix, chen2023pali3, li2023monkey, xu2024llava_uhd, chen2024far, liu2024llavanext, yao2024minicpm, dong2024xc24khd, chen2024internvl2, wang2024qwen2vl}. To our best knowledge, this work is the first to explore the tiled mixture of vision encoder (MoVE) design, which is shown to inherit the benefits from both worlds. The proposed tiled MoVE design also introduces additional flexibility to incorporate advanced vision foundation models.

\noindent\textbf{Data Efforts in VLMs.}
Data strategy is crucial in training VLMs, encompassing aspects of data set construction, balance and filtering, and training methodologies. Early endeavors such as LLaVA-150K~\cite{liu2023llava} used instructed tuning with GPT-4~\cite{gpt4v}, which was later enriched by successors~\cite{liu2023improved, instructblip, liu2024llavanext,chen2023internvl} incorporating academic training data from various tasks into the supervised fine-tuning stage. Studies also broadened data types to include video~\cite{li2023mimicit, dubey2024llama3}, multi-image inputs~\cite{li2024llavaonevision,chen2024internvl2}, image-text interleaved data~\cite{li2024omnicorpus, awadalla2024mint1t}, multilingual data~\cite{hu2024minicpm}, and synthetic datasets~\cite{dubey2024llama3}. However, simply expanding data sets can compromise model performance due to varying quality and size. Approaches like Instruct-BLIP~\cite{instructblip} and Cambrian-1~\cite{tong2024cambrian} addressed this by devising optimal data ratios and balancing techniques, while others like Llama3~\cite{dubey2024llama3} and Molmo~\cite{deitke2024molmo} focused on enhancing data quality by removing duplicates with SSCD~\cite{pizzi2022sscd} and incorporating human-annotated images, respectively.  In addition, Training strategies have also evolved, with LLaVA~\cite{liu2023llava} proposing a two-stage training process that has become a standard, and later models~\cite{li2024llavaonevision} introducing intermediate stages. VLM surveys~\cite{bai2024survey, yin2023survey, wu2023multimodal} also discuss various training recipes and data strategies for building VLMs, however, they lack qualitative analysis and do not provide a detailed enough path for training cutting-edge VLMs.
\section{Conclusion}
As publicly available frontier VLMs continue to approach or even surpass proprietary commercial models, the detailed data strategies of these leading VLMs remains unknown to the community. In this paper, we have unveiled many details on the post-training data strategy for training frontier VLMs. Our covered data strategy is effective and comprehensive. We hope this work offers a transparent practice to inspire the community.

\section{Demos}
This section provides some examples to demonstrate Eagle2 capabilities. To avoid cherry-picking, we directly select demo cases from other works (Qwen2-VL and InternVL2) as our test cases.

\begin{figure*}[htbp]
\centering
\begin{tcolorbox}[colback=black!5!white,colframe=gray!75!black,title=Document Parsing with Dense Formulas (Example borrowed from Qwen2-VL paper)]
\centering
\includegraphics[width=5cm]{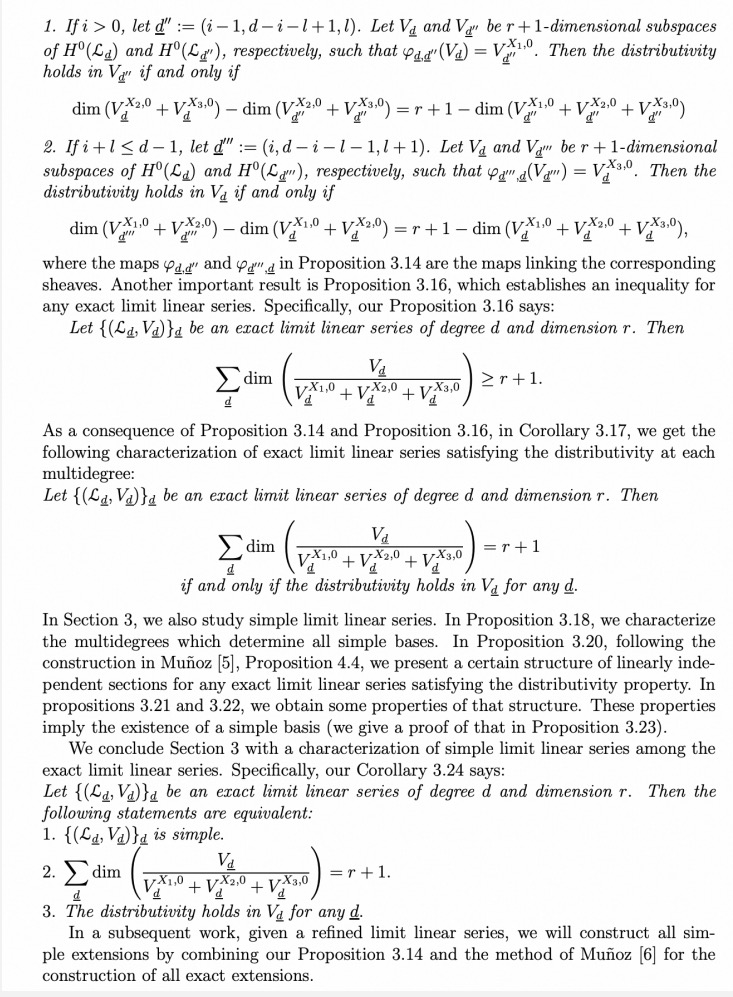}

\tcbsubtitle[colback=gray]{Input: Extract the Text content.}
\tcbsubtitle[colback=gray]{Model Response}
\includegraphics[width=6cm]{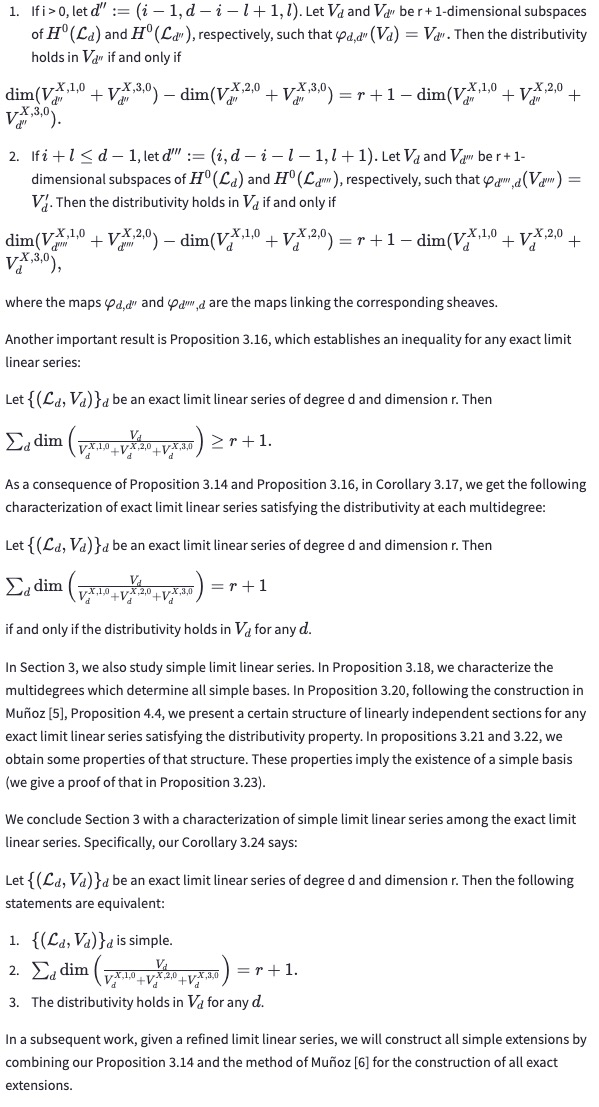}
\label{mt_format1}
\end{tcolorbox}
\caption{Eagle2-9B has strong OCR recognition capabilities.}
\end{figure*}

\begin{figure*}[htbp]
\centering
\begin{tcolorbox}[colback=black!5!white,colframe=gray!75!black,title=Multilingual Text Recognition (Example borrowed from Qwen2-VL paper)]
\centering
\includegraphics[width=9cm]{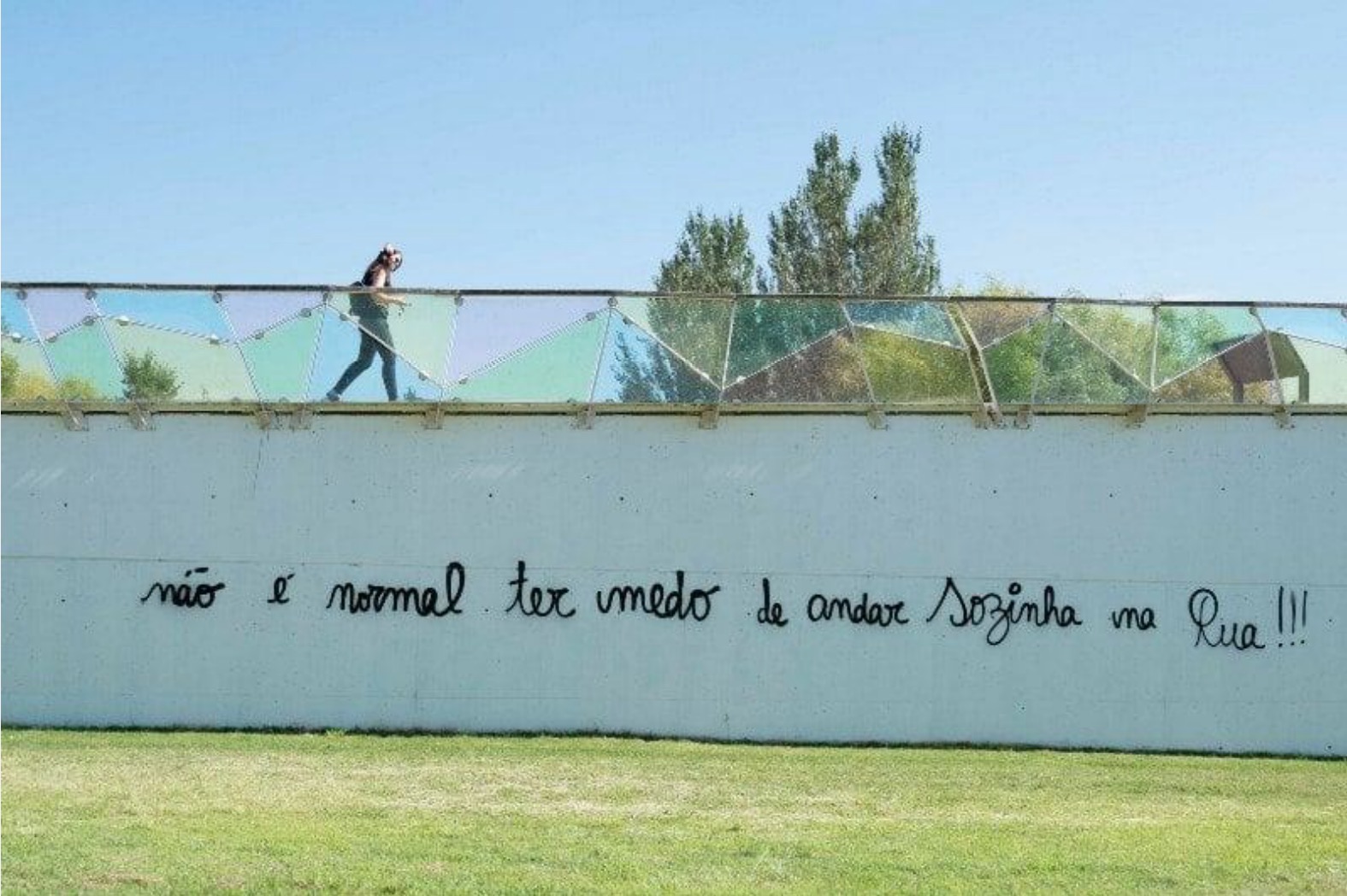}

\tcbsubtitle[colback=gray]{Input and Model Response}
\includegraphics[width=9cm]{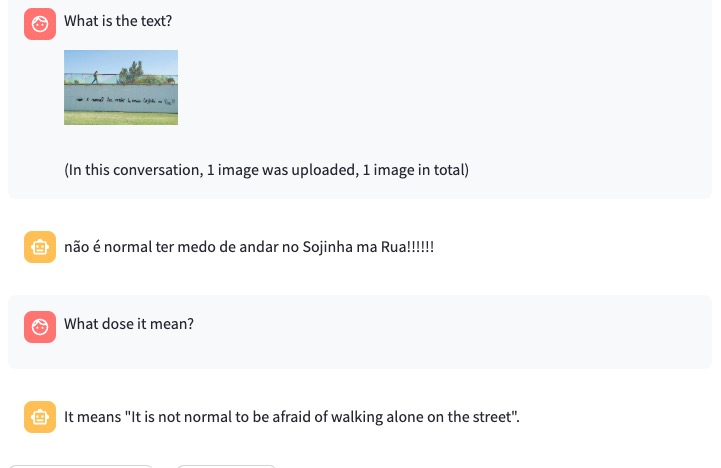}
\label{mt_format2}
\end{tcolorbox}
\caption{Eagle2-9B has Multilingual OCR Recognition Capability.}
\end{figure*}

\begin{figure*}[htbp]
\centering
\begin{tcolorbox}[colback=black!5!white,colframe=gray!75!black,title=Mathematical Problem Solving (Example borrowed from Qwen2-VL paper)]
\centering
\includegraphics[width=9cm]{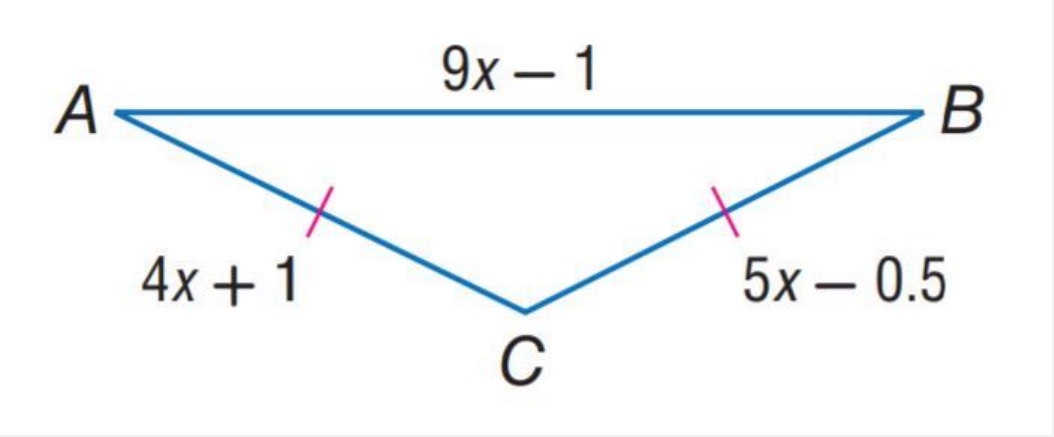}

\tcbsubtitle[colback=gray]{Input and Model Response}
\includegraphics[width=9cm]{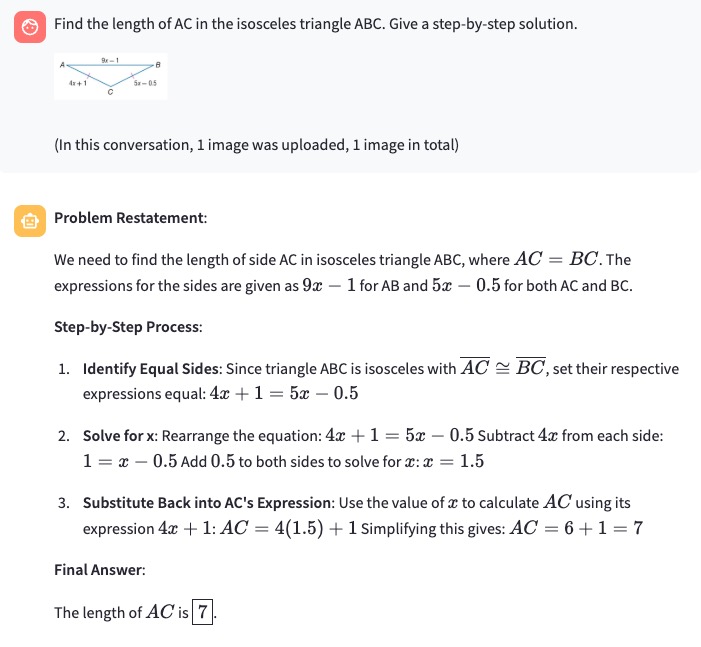}
\label{mt_format3}
\end{tcolorbox}
\caption{CoT Inference ability.}
\end{figure*}

\begin{figure*}[htbp]
\centering
\begin{tcolorbox}[colback=black!5!white,colframe=gray!75!black,title=Algorithmic Problem Solving (Example borrowed from Qwen2-VL paper)]
\centering
\includegraphics[width=9cm]{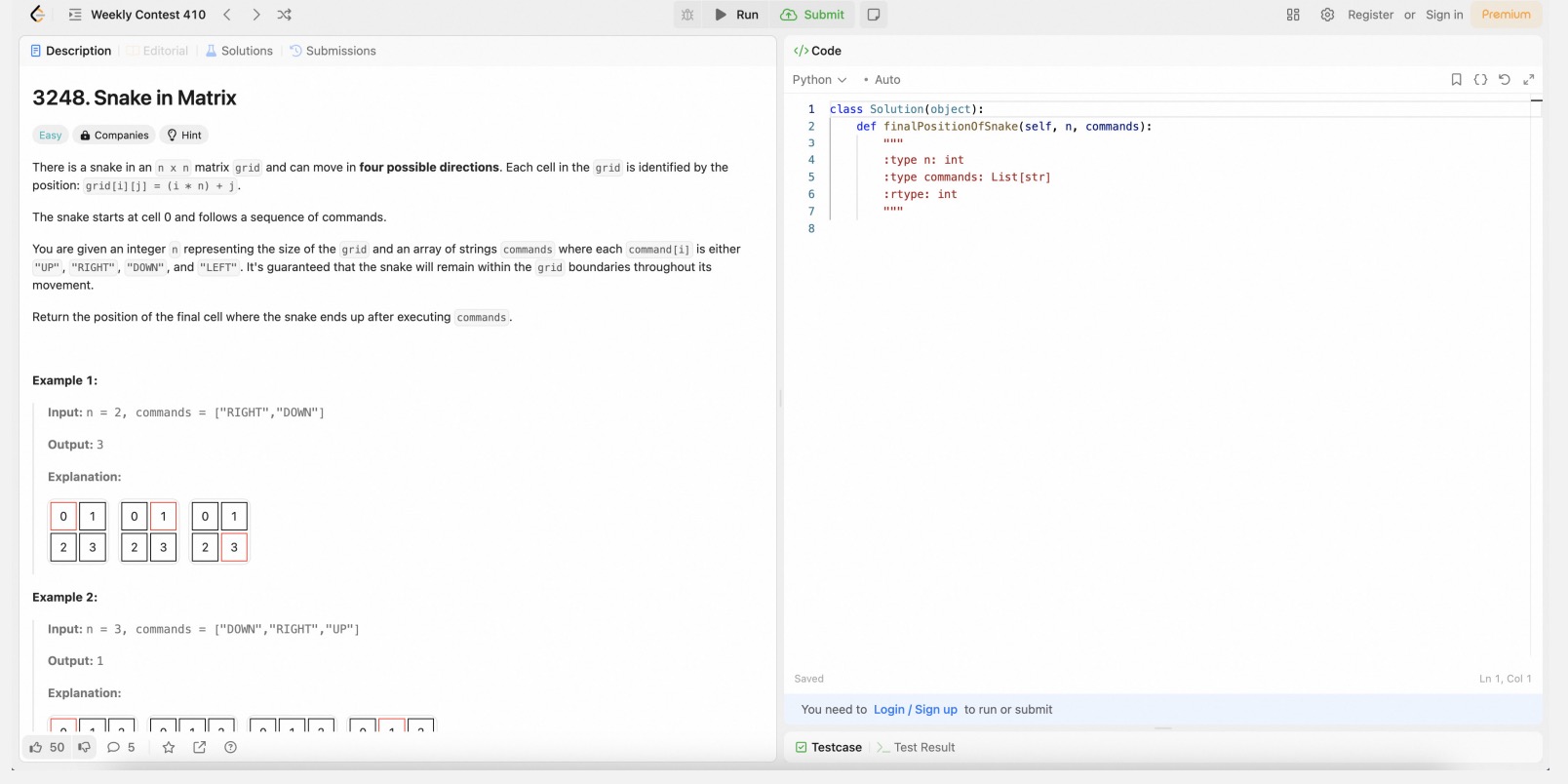}
\tcbsubtitle[colback=gray]{Input: Solve the algorithmic problem in the web page.}
\tcbsubtitle[colback=gray] {Model Response}
\includegraphics[width=9cm]{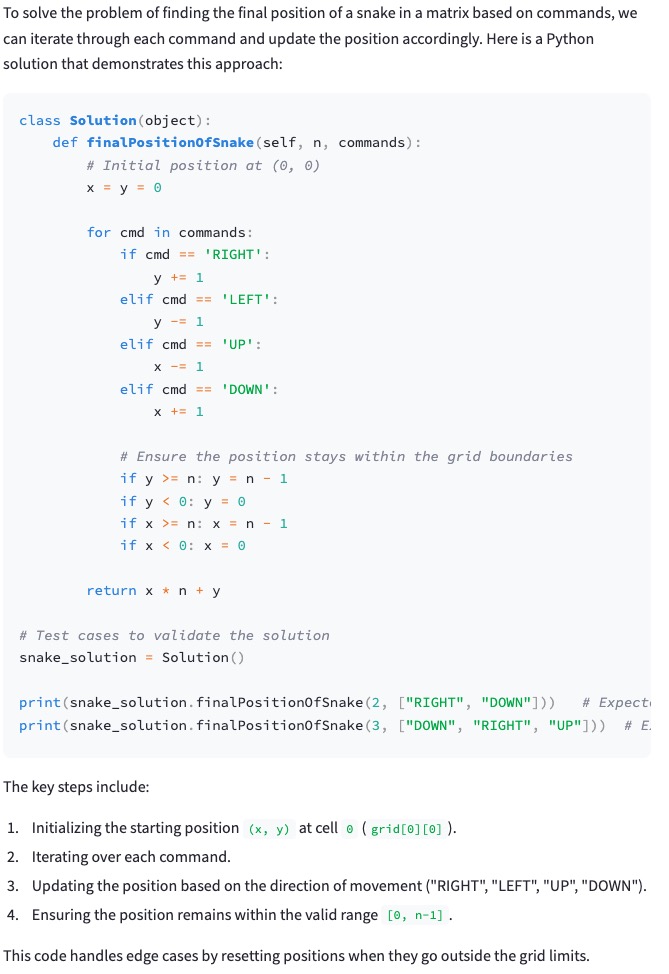}
\label{mt_format4}
\end{tcolorbox}
\caption{Solving ``Easy" algorithmic problem via Eagle2-9B.}
\end{figure*}

\begin{figure*}[htbp]
\centering
\begin{tcolorbox}[colback=black!5!white,colframe=gray!75!black,title= Image Analysis (Example borrowed from InternVL2 demos)]
\centering
\includegraphics[width=9cm]{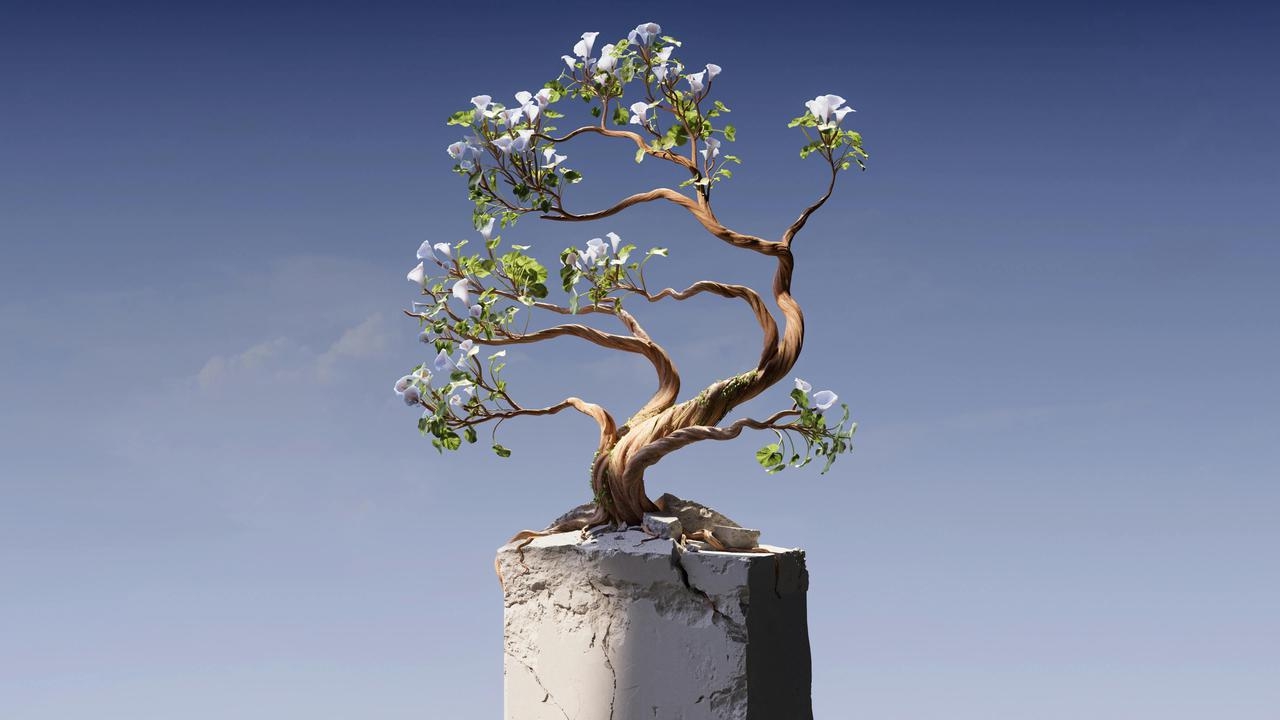}
\tcbsubtitle[colback=gray] {Input and Model Response}
\includegraphics[width=9cm]{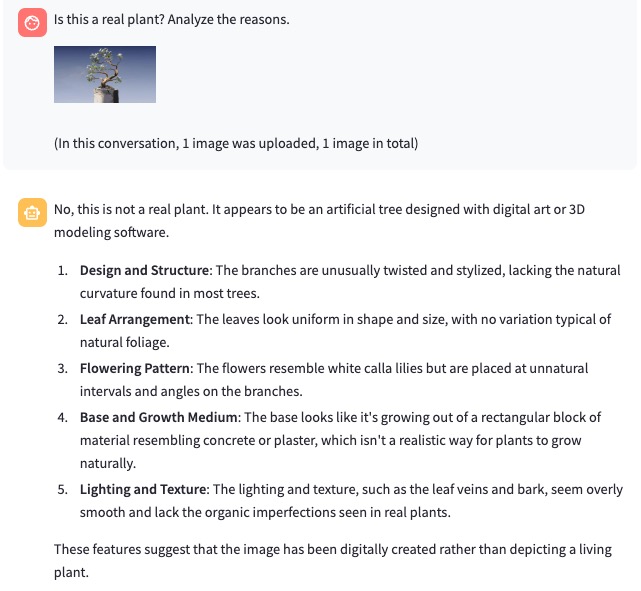}
\label{mt_format5}
\end{tcolorbox}
\caption{Eagle2-9B is able to analyze anomalies in images.}
\end{figure*}

\begin{figure*}[htbp]
\centering
\begin{tcolorbox}[colback=black!5!white,colframe=gray!75!black,title= Image Analysis (Example borrowed from InternVL2 demos)]
\centering
\includegraphics[width=9cm]{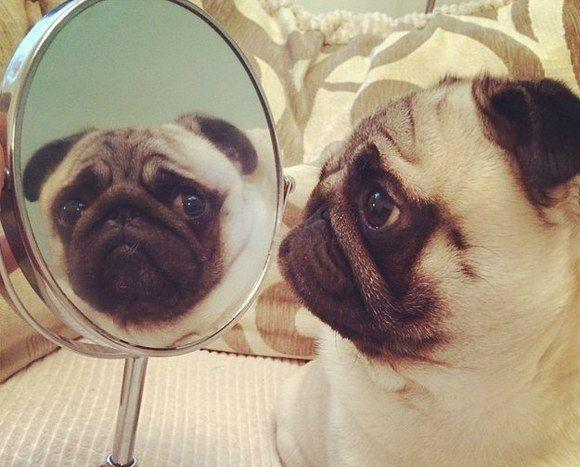}
\tcbsubtitle[colback=gray] {Input and Model Response}
\includegraphics[width=9cm]{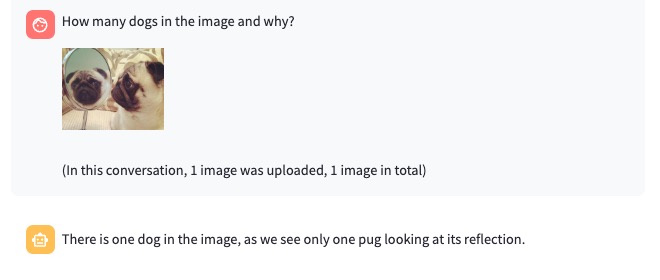}
\label{mt_format6}
\end{tcolorbox}
\caption{Eagle2-9B can correctly distinguish objects in the mirror.}
\end{figure*}

\begin{figure*}[htbp]
\centering
\begin{tcolorbox}[colback=black!5!white,colframe=gray!75!black,title=Handwritting OCR (Example borrowed from InternVL2 demos)]
\centering
\includegraphics[width=9cm]{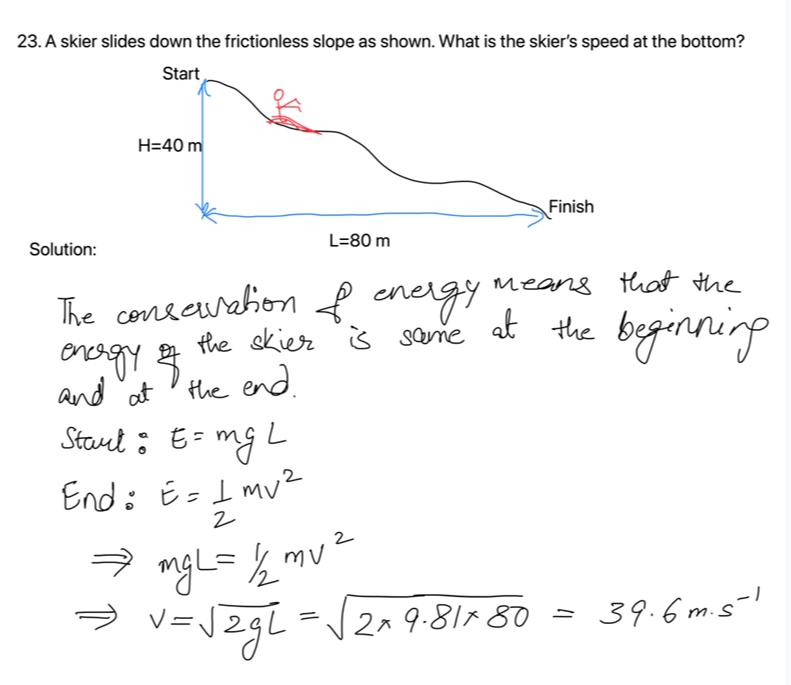}
\tcbsubtitle[colback=gray] {Input and Model Response}
\includegraphics[width=9cm]{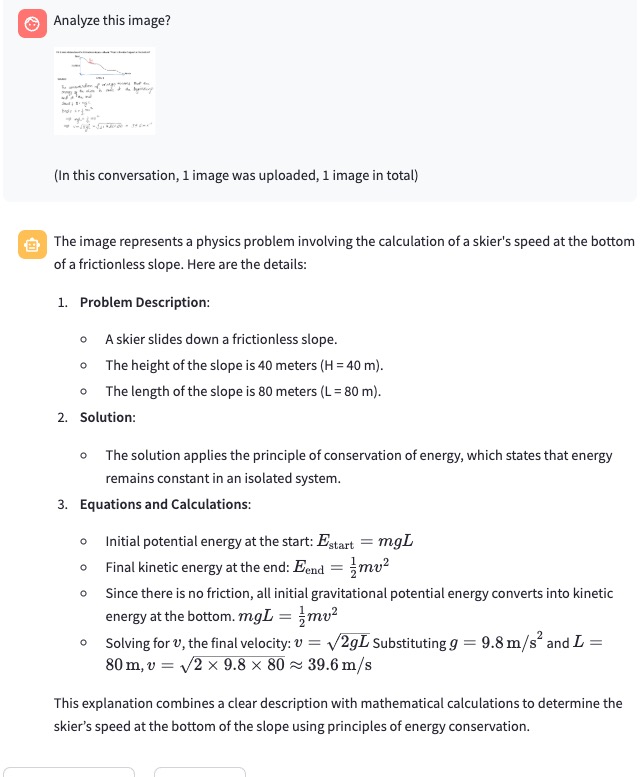}
\label{mt_format7}
\end{tcolorbox}
\caption{Eagle2-9B has excellent handwriting recognition capabilities.}
\end{figure*}

\section{Acknowledgments}
The team would like to thank the valuable discussions and input from Jihao Liu, Yunze Man, Yue Zhao, Min Shi, Fuxiao Liu, Jindong Jiang, Ming-Chang Chiu, Wonmin Beyon, Jason Lu, Wei Ping, Jon Barker, Zhuolin Yang, Wenliang Dai, Nayeon Lee, Boxin Wang, Mohammad Shoeybi, Tyler Poon, Jim Fan, Yuke Zhu, Danny Yin, Mike Ranzinger, Greg Heinrich, Yonggan Fu, Xin Dong, Shizhe Diao, Pavlo Molchanov, Yi Dong, Shiwei Sheng, Xiaolong Li, Vidya Murali, Parthasarathy Sriram, Song Han, Monika Jhuria, Shubham Agrawal, Wenyuan Zhang, Ke Chen, Yan Wang, Byung-Kwan Lee, Yueh-Hua Wu, Min-Hung Chen, Ryo Hachiuma, Chao-Han Huck Yang, Yao Xu, Osvald Nitski, Elena Lantz, Qing Miao, Ofri Masad, Ofer Baratz, Niket Agarwal, Sriharsha Niverty, and Padmavathy Subramanian.
{
  \small
  \bibliographystyle{unsrt}
  \bibliography{main}

\begin{thebibliography}{100}

\bibitem{alayrac2022flamingo}
Jean-Baptiste Alayrac, Jeff Donahue, Pauline Luc, Antoine Miech, Iain Barr, Yana Hasson, Karel Lenc, Arthur Mensch, Katherine Millican, Malcolm Reynolds, et~al.
\newblock Flamingo: a visual language model for few-shot learning.
\newblock {\em NeurIPS}, 2022.

\bibitem{li2022blip}
Junnan Li, Dongxu Li, Caiming Xiong, and Steven Hoi.
\newblock Blip: Bootstrapping language-image pre-training for unified vision-language understanding and generation.
\newblock In {\em ICML}, 2022.

\bibitem{chen2022pali}
Xi~Chen, Xiao Wang, Soravit Changpinyo, AJ~Piergiovanni, Piotr Padlewski, Daniel Salz, Sebastian Goodman, Adam Grycner, Basil Mustafa, Lucas Beyer, et~al.
\newblock Pali: A jointly-scaled multilingual language-image model.
\newblock In {\em ICLR}, 2022.

\bibitem{liu2023llava}
Haotian Liu, Chunyuan Li, Qingyang Wu, and Yong~Jae Lee.
\newblock Visual instruction tuning.
\newblock In {\em NeurIPS}, 2023.

\bibitem{hong2023cogagent}
Wenyi Hong, Weihan Wang, Qingsong Lv, Jiazheng Xu, Wenmeng Yu, Junhui Ji, Yan Wang, Zihan Wang, Yuxiao Dong, Ming Ding, et~al.
\newblock Cogagent: A visual language model for gui agents.
\newblock {\em arXiv:2312.08914}, 2023.

\bibitem{tian2024drivevlm}
Xiaoyu Tian, Junru Gu, Bailin Li, Yicheng Liu, Yang Wang, Zhiyong Zhao, Kun Zhan, Peng Jia, Xianpeng Lang, and Hang Zhao.
\newblock {DriveVLM}: The convergence of autonomous driving and large vision-language models.
\newblock In {\em CoRL}, 2024.

\bibitem{marcu2023lingoqa}
Ana-Maria Marcu, Long Chen, Jan H{\"u}nermann, Alice Karnsund, Benoit Hanotte, Prajwal Chidananda, Saurabh Nair, Vijay Badrinarayanan, Alex Kendall, Jamie Shotton, et~al.
\newblock {LingoQA}: Video question answering for autonomous driving.
\newblock In {\em ECCV}, 2023.

\bibitem{brohan2023rt}
Anthony Brohan, Noah Brown, Justice Carbajal, Yevgen Chebotar, Xi~Chen, Krzysztof Choromanski, Tianli Ding, Danny Driess, Avinava Dubey, Chelsea Finn, et~al.
\newblock {RT-2}: Vision-language-action models transfer web knowledge to robotic control.
\newblock {\em arXiv:2307.15818}, 2023.

\bibitem{driess2023palme}
Danny Driess, Fei Xia, Mehdi~SM Sajjadi, Corey Lynch, Aakanksha Chowdhery, Brian Ichter, Ayzaan Wahid, Jonathan Tompson, Quan Vuong, Tianhe Yu, et~al.
\newblock Palm-e: An embodied multimodal language model.
\newblock {\em arXiv:2303.03378}, 2023.

\bibitem{kim2024openvla}
Moo~Jin Kim, Karl Pertsch, Siddharth Karamcheti, Ted Xiao, Ashwin Balakrishna, Suraj Nair, Rafael Rafailov, Ethan Foster, Grace Lam, Pannag Sanketi, et~al.
\newblock {OpenVLA}: An open-source vision-language-action model.
\newblock {\em arXiv:2406.09246}, 2024.

\bibitem{gpt4v}
OpenAI.
\newblock {GPT-4V(ision)} system card.
\newblock \url{https://cdn.openai.com/papers/GPTV_System_Card.pdf}, 2023.

\bibitem{claude3series2024}
{Anthropic}.
\newblock The claude 3 model family: Opus, sonnet, haiku.
\newblock \url{https://www.anthropic.com}, 2024.

\bibitem{yang2024qwen2}
An~Yang, Baosong Yang, Binyuan Hui, Bo~Zheng, Bowen Yu, Chang Zhou, Chengpeng Li, Chengyuan Li, Dayiheng Liu, Fei Huang, et~al.
\newblock Qwen2 technical report.
\newblock {\em arXiv:2407.10671}, 2024.

\bibitem{chen2024internvl2}
OpenGVLab Team.
\newblock {InternVL2}: Better than the best—expanding performance boundaries of open-source multimodal models with the progressive scaling strategy.
\newblock \url{https://internvl.github.io/blog/2024-07-02-InternVL-2.0/}, 2024.

\bibitem{dubey2024llama3}
Abhimanyu Dubey, Abhinav Jauhri, Abhinav Pandey, Abhishek Kadian, Ahmad Al-Dahle, Aiesha Letman, Akhil Mathur, Alan Schelten, Amy Yang, Angela Fan, et~al.
\newblock The llama 3 herd of models.
\newblock {\em arXiv:2407.21783}, 2024.

\bibitem{tong2024cambrian}
Shengbang Tong, Ellis Brown, Penghao Wu, Sanghyun Woo, Manoj Middepogu, Sai~Charitha Akula, Jihan Yang, Shusheng Yang, Adithya Iyer, Xichen Pan, et~al.
\newblock Cambrian-1: A fully open, vision-centric exploration of multimodal llms.
\newblock {\em arXiv:2406.16860}, 2024.

\bibitem{li2024llavaonevision}
Bo~Li, Yuanhan Zhang, Dong Guo, Renrui Zhang, Feng Li, Hao Zhang, Kaichen Zhang, Yanwei Li, Ziwei Liu, and Chunyuan Li.
\newblock Llava-onevision: Easy visual task transfer.
\newblock {\em arXiv:2408.03326}, 2024.

\bibitem{opencompass2023}
OpenCompass Contributors.
\newblock Opencompass: A universal evaluation platform for foundation models.
\newblock \url{https://github.com/open-compass/opencompass}, 2023.

\bibitem{zhang2024vision_qformer}
Qiming Zhang, Jing Zhang, Yufei Xu, and Dacheng Tao.
\newblock Vision transformer with quadrangle attention.
\newblock {\em IEEE Transactions on Pattern Analysis and Machine Intelligence}, 2024.

\bibitem{bai2023qwenvl}
Jinze Bai, Shuai Bai, Shusheng Yang, Shijie Wang, Sinan Tan, Peng Wang, Junyang Lin, Chang Zhou, and Jingren Zhou.
\newblock Qwen-vl: A frontier large vision-language model with versatile abilities.
\newblock {\em arXiv:2308.12966}, 2023.

\bibitem{chen2024far}
Zhe Chen, Weiyun Wang, Hao Tian, Shenglong Ye, Zhangwei Gao, Erfei Cui, Wenwen Tong, Kongzhi Hu, Jiapeng Luo, Zheng Ma, et~al.
\newblock How far are we to gpt-4v? closing the gap to commercial multimodal models with open-source suites.
\newblock {\em arXiv:2404.16821}, 2024.

\bibitem{eagle}
Min Shi, Fuxiao Liu, Shihao Wang, Shijia Liao, Subhashree Radhakrishnan, De-An Huang, Hongxu Yin, Karan Sapra, Yaser Yacoob, Humphrey Shi, Bryan Catanzaro, Andrew Tao, Jan Kautz, Zhiding Yu, and Guilin Liu.
\newblock Eagle: Exploring the design space for multimodal llms with mixture of encoders.
\newblock {\em arXiv:2408.15998}, 2024.

\bibitem{zhai2023siglip}
Xiaohua Zhai, Basil Mustafa, Alexander Kolesnikov, and Lucas Beyer.
\newblock Sigmoid loss for language image pre-training.
\newblock In {\em ICCV}, 2023.

\bibitem{liu2022convnet}
Zhuang Liu, Hanzi Mao, Chao-Yuan Wu, Christoph Feichtenhofer, Trevor Darrell, and Saining Xie.
\newblock A convnet for the 2020s.
\newblock In {\em CVPR}, 2022.

\bibitem{wang2024qwen2vl}
Peng Wang, Shuai Bai, Sinan Tan, Shijie Wang, Zhihao Fan, Jinze Bai, Keqin Chen, Xuejing Liu, Jialin Wang, Wenbin Ge, et~al.
\newblock {Qwen2-VL}: Enhancing vision-language model's perception of the world at any resolution.
\newblock {\em arXiv:2409.12191}, 2024.

\bibitem{openai2023gpt4}
Josh Achiam, Steven Adler, Sandhini Agarwal, Lama Ahmad, Ilge Akkaya, Florencia~Leoni Aleman, Diogo Almeida, Janko Altenschmidt, Sam Altman, Shyamal Anadkat, et~al.
\newblock Gpt-4 technical report.
\newblock {\em arXiv:2303.08774}, 2023.

\bibitem{qwen2.5}
Qwen Team.
\newblock Qwen2.5: A party of foundation models, September 2024.

\bibitem{opengvlab_sharegpt4o_dataset}
OpenGVLab.
\newblock Sharegpt-4o dataset.
\newblock \url{https://huggingface.co/datasets/OpenGVLab/ShareGPT-4o}, 2024.

\bibitem{shah2019kvqa}
Sanket Shah, Anand Mishra, Naganand Yadati, and Partha~Pratim Talukdar.
\newblock Kvqa: Knowledge-aware visual question answering.
\newblock In {\em AAAI}, 2019.

\bibitem{skvarre_movie_posters_100k}
skvarre.
\newblock Movie posters-100k dataset.
\newblock \url{https://huggingface.co/datasets/skvarre/movie_posters-100k}, 2024.

\bibitem{weyand2020googlelandmark}
Tobias Weyand, André Araujo, Bingyi Cao, and Jack Sim.
\newblock {Google Landmarks Dataset v2 - A Large-Scale Benchmark for Instance-Level Recognition and Retrieval}.
\newblock In {\em CVPR}, 2020.

\bibitem{wikiart_dataset}
HugGAN.
\newblock Wikiart dataset.
\newblock \url{https://huggingface.co/datasets/huggan/wikiart}, 2024.

\bibitem{ma2024weatherqa}
Chengqian Ma, Zhanxiang Hua, Alexandra Anderson-Frey, Vikram Iyer, Xin Liu, and Lianhui Qin.
\newblock Weatherqa: Can multimodal language models reason about severe weather?
\newblock {\em arXiv:2406.11217}, 2024.

\bibitem{mscoco-controlnet-canny-less-colors}
hazal karakus.
\newblock mscoco-controlnet-canny-less-colors dataset.
\newblock \url{https://huggingface.co/datasets/hazal-karakus/mscoco-controlnet-canny-less-colors}, 2024.

\bibitem{sheet_music_clean}
Emile Esmaili.
\newblock sheet music clean ataset.
\newblock \url{https://huggingface.co/datasets/EmileEsmaili/sheet_music_clean}, 2024.

\bibitem{yu2024spark}
Youngjoon Yu, Sangyun Chung, Byung-Kwan Lee, and Yong~Man Ro.
\newblock Spark: Multi-vision sensor perception and reasoning benchmark for large-scale vision-language models.
\newblock {\em arXiv:2408.12114}, 2024.

\bibitem{pi2024image_textualization}
Renjie Pi, Jianshu Zhang, Jipeng Zhang, Rui Pan, Zhekai Chen, and Tong Zhang.
\newblock Image textualization: An automatic framework for creating accurate and detailed image descriptions.
\newblock {\em arXiv:2406.07502}, 2024.

\bibitem{pixart_alpha_sam_llava_captions10m}
PixArt-alpha.
\newblock Sam-llava-captions10m dataset.
\newblock \url{https://huggingface.co/datasets/PixArt-alpha/SAM-LLaVA-Captions10M}, 2024.

\bibitem{ashraq_tmdb_celeb_10k}
Ashraq.
\newblock Tmdb-celeb-10k dataset.
\newblock \url{https://huggingface.co/datasets/ashraq/tmdb-celeb-10k}, 2024.

\bibitem{cao2022geoqa_plus}
Jie Cao and Jing Xiao.
\newblock An augmented benchmark dataset for geometric question answering through dual parallel text encoding.
\newblock In {\em COLING}, 2022.

\bibitem{yu2023mathqa}
Longhui Yu, Weisen Jiang, Han Shi, Jincheng Yu, Zhengying Liu, Yu~Zhang, James~T Kwok, Zhenguo Li, Adrian Weller, and Weiyang Liu.
\newblock {MetaMath}: Bootstrap your own mathematical questions for large language models.
\newblock {\em arXiv:2309.12284}, 2023.

\bibitem{lindstrom2022clevrmath}
Adam~Dahlgren Lindstr{\"o}m and Savitha~Sam Abraham.
\newblock Clevr-math: A dataset for compositional language, visual and mathematical reasoning.
\newblock {\em arXiv:2208.05358}, 2022.

\bibitem{li2023superclevr}
Zhuowan Li, Xingrui Wang, Elias Stengel-Eskin, Adam Kortylewski, Wufei Ma, Benjamin Van~Durme, and Alan~L Yuille.
\newblock Super-clevr: A virtual benchmark to diagnose domain robustness in visual reasoning.
\newblock In {\em CVPR}, 2023.

\bibitem{lu2021geometry3k}
Pan Lu, Ran Gong, Shibiao Jiang, Liang Qiu, Siyuan Huang, Xiaodan Liang, and Song-Chun Zhu.
\newblock Inter-gps: Interpretable geometry problem solving with formal language and symbolic reasoning.
\newblock {\em arXiv:2105.04165}, 2021.

\bibitem{zhang2024mavis}
Renrui Zhang, Xinyu Wei, Dongzhi Jiang, Yichi Zhang, Ziyu Guo, Chengzhuo Tong, Jiaming Liu, Aojun Zhou, Bin Wei, Shanghang Zhang, et~al.
\newblock Mavis: Mathematical visual instruction tuning.
\newblock {\em arXiv:2407.08739}, 2024.

\bibitem{lu2021intergps}
Pan Lu, Ran Gong, Shibiao Jiang, Liang Qiu, Siyuan Huang, Xiaodan Liang, and Song-Chun Zhu.
\newblock Inter-gps: Interpretable geometry problem solving with formal language and symbolic reasoning.
\newblock {\em arXiv:2105.04165}, 2021.

\bibitem{zhang2019raven}
Chi Zhang, Feng Gao, Baoxiong Jia, Yixin Zhu, and Song-Chun Zhu.
\newblock Raven: A dataset for relational and analogical visual reasoning.
\newblock In {\em CVPR}, 2019.

\bibitem{seo2015geos}
Minjoon Seo, Hannaneh Hajishirzi, Ali Farhadi, Oren Etzioni, and Clint Malcolm.
\newblock Solving geometry problems: Combining text and diagram interpretation.
\newblock In {\em EMNLP}, 2015.

\bibitem{chen2022unigeo}
Jiaqi Chen, Tong Li, Jinghui Qin, Pan Lu, Liang Lin, Chongyu Chen, and Xiaodan Liang.
\newblock Unigeo: Unifying geometry logical reasoning via reformulating mathematical expression.
\newblock {\em arXiv:2212.02746}, 2022.

\bibitem{kembhavi2016ai2d}
Aniruddha Kembhavi, Mike Salvato, Eric Kolve, Minjoon Seo, Hannaneh Hajishirzi, and Ali Farhadi.
\newblock A diagram is worth a dozen images.
\newblock In {\em ECCV}, 2016.

\bibitem{lu2022scienceqa}
Pan Lu, Swaroop Mishra, Tanglin Xia, Liang Qiu, Kai-Wei Chang, Song-Chun Zhu, Oyvind Tafjord, Peter Clark, and Ashwin Kalyan.
\newblock Learn to explain: Multimodal reasoning via thought chains for science question answering.
\newblock {\em NeurIPS}, 2022.

\bibitem{kembhavi2017tqa}
Aniruddha Kembhavi, Minjoon Seo, Dustin Schwenk, Jonghyun Choi, Ali Farhadi, and Hannaneh Hajishirzi.
\newblock Are you smarter than a sixth grader? textbook question answering for multimodal machine comprehension.
\newblock In {\em CVPR}, 2017.

\bibitem{he2020pathvqa}
Xuehai He, Yichen Zhang, Luntian Mou, Eric Xing, and Pengtao Xie.
\newblock Pathvqa: 30000+ questions for medical visual question answering.
\newblock {\em arXiv:2003.10286}, 2020.

\bibitem{auer2023sciqa}
S{\"o}ren Auer, Dante~AC Barone, Cassiano Bartz, Eduardo~G Cortes, Mohamad~Yaser Jaradeh, Oliver Karras, Manolis Koubarakis, Dmitry Mouromtsev, Dmitrii Pliukhin, Daniil Radyush, et~al.
\newblock The sciqa scientific question answering benchmark for scholarly knowledge.
\newblock {\em Scientific Reports}, 13(1):7240, 2023.

\bibitem{lau2018vqarad}
Jason~J Lau, Soumya Gayen, Asma Ben~Abacha, and Dina Demner-Fushman.
\newblock A dataset of clinically generated visual questions and answers about radiology images.
\newblock {\em Scientific data}, 5(1):1--10, 2018.

\bibitem{tiger_lab_visualwebinstruct}
TIGER-Lab.
\newblock Visualwebinstruct dataset.
\newblock \url{https://huggingface.co/datasets/TIGER-Lab/VisualWebInstruct}, 2024.

\bibitem{masry2022chartqa}
Ahmed Masry, Xuan~Long Do, Jia~Qing Tan, Shafiq Joty, and Enamul Hoque.
\newblock {ChartQA}: A benchmark for question answering about charts with visual and logical reasoning.
\newblock In {\em ACL}, 2022.

\bibitem{liu2023mmcinst}
Fuxiao Liu, Xiaoyang Wang, Wenlin Yao, Jianshu Chen, Kaiqiang Song, Sangwoo Cho, Yaser Yacoob, and Dong Yu.
\newblock {MMC}: Advancing multimodal chart understanding with large-scale instruction tuning.
\newblock {\em arXiv:2311.10774}, 2023.

\bibitem{kafle2018dvqa}
Kushal Kafle, Brian Price, Scott Cohen, and Christopher Kanan.
\newblock Dvqa: Understanding data visualizations via question answering.
\newblock In {\em CVPR}, 2018.

\bibitem{methani2020plotqa}
Nitesh Methani, Pritha Ganguly, Mitesh~M Khapra, and Pratyush Kumar.
\newblock {PlotQA}: Reasoning over scientific plots.
\newblock In {\em WACV}, 2020.

\bibitem{liu2023lrv-instruction}
Fuxiao Liu, Kevin Lin, Linjie Li, Jianfeng Wang, Yaser Yacoob, and Lijuan Wang.
\newblock Aligning large multi-modal model with robust instruction tuning.
\newblock {\em arXiv:2306.14565}, 2023.

\bibitem{lu2022tablemwp}
Pan Lu, Liang Qiu, Kai-Wei Chang, Ying~Nian Wu, Song-Chun Zhu, Tanmay Rajpurohit, Peter Clark, and Ashwin Kalyan.
\newblock Dynamic prompt learning via policy gradient for semi-structured mathematical reasoning.
\newblock {\em arXiv:2209.14610}, 2022.

\bibitem{masry2023unichart}
Ahmed Masry, Parsa Kavehzadeh, Xuan~Long Do, Enamul Hoque, and Shafiq Joty.
\newblock {UniChart}: A universal vision-language pretrained model for chart comprehension and reasoning.
\newblock {\em arXiv:2305.14761}, 2023.

\bibitem{tang2023vistext}
Benny~J Tang, Angie Boggust, and Arvind Satyanarayan.
\newblock Vistext: A benchmark for semantically rich chart captioning.
\newblock {\em arXiv:2307.05356}, 2023.

\bibitem{zhu2022tatdqa}
Fengbin Zhu, Wenqiang Lei, Fuli Feng, Chao Wang, Haozhou Zhang, and Tat-Seng Chua.
\newblock Towards complex document understanding by discrete reasoning.
\newblock In {\em ACMMM}, 2022.

\bibitem{VQAonDB}
VQAonDB.
\newblock Vqaondb dataset.
\newblock \url{https://ilocr.iiit.ac.in/vqabd/}.

\bibitem{kahou2017figureqa}
Samira~Ebrahimi Kahou, Vincent Michalski, Adam Atkinson, {\'A}kos K{\'a}d{\'a}r, Adam Trischler, and Yoshua Bengio.
\newblock Figureqa: An annotated figure dataset for visual reasoning.
\newblock {\em arXiv:1710.07300}, 2017.

\bibitem{kantharaj2022chart2text}
Shankar Kantharaj, Rixie Tiffany~Ko Leong, Xiang Lin, Ahmed Masry, Megh Thakkar, Enamul Hoque, and Shafiq Joty.
\newblock Chart-to-text: A large-scale benchmark for chart summarization.
\newblock {\em arXiv:2203.06486}, 2022.

\bibitem{zhao2023robut}
Yilun Zhao, Chen Zhao, Linyong Nan, Zhenting Qi, Wenlin Zhang, Xiangru Tang, Boyu Mi, and Dragomir Radev.
\newblock Robut: A systematic study of table qa robustness against human-annotated adversarial perturbations.
\newblock {\em arXiv:2306.14321}, 2023.

\bibitem{zhao2022multihiertt}
Yilun Zhao, Yunxiang Li, Chenying Li, and Rui Zhang.
\newblock Multihiertt: Numerical reasoning over multi hierarchical tabular and textual data.
\newblock {\em arXiv:2206.01347}, 2022.

\bibitem{kim2022synthdog}
Geewook Kim, Teakgyu Hong, Moonbin Yim, JeongYeon Nam, Jinyoung Park, Jinyeong Yim, Wonseok Hwang, Sangdoo Yun, Dongyoon Han, and Seunghyun Park.
\newblock {OCR-Free} document understanding transformer.
\newblock In {\em ECCV}, 2022.

\bibitem{he2018icpr2018_MTWI}
Mengchao He, Yuliang Liu, Zhibo Yang, Sheng Zhang, Canjie Luo, Feiyu Gao, Qi~Zheng, Yongpan Wang, Xin Zhang, and Lianwen Jin.
\newblock Icpr2018 contest on robust reading for multi-type web images.
\newblock In {\em ICPR}, 2018.

\bibitem{sun2019lsvt}
Yipeng Sun, Zihan Ni, Chee-Kheng Chng, Yuliang Liu, Canjie Luo, Chun~Chet Ng, Junyu Han, Errui Ding, Jingtuo Liu, Dimosthenis Karatzas, et~al.
\newblock Icdar 2019 competition on large-scale street view text with partial labeling-rrc-lsvt.
\newblock In {\em ICDAR}, 2019.

\bibitem{huang2019icdar_sroie}
Zheng Huang, Kai Chen, Jianhua He, Xiang Bai, Dimosthenis Karatzas, Shijian Lu, and CV~Jawahar.
\newblock Icdar 2019 robust reading challenge on scanned receipts ocr and information extraction.
\newblock In {\em International conference on document analysis recognition}, 2019.

\bibitem{jaume2019funsd}
Guillaume Jaume, Hazim~Kemal Ekenel, and Jean-Philippe Thiran.
\newblock Funsd: A dataset for form understanding in noisy scanned documents.
\newblock In {\em ICDAR Workshops}, 2019.

\bibitem{oleehyo_latex_formulas}
OleehyO.
\newblock Latex formulas dataset.
\newblock \url{https://huggingface.co/datasets/OleehyO/latex-formulas}, 2024.

\bibitem{marti2002iam}
U-V Marti and Horst Bunke.
\newblock The iam-database: an english sentence database for offline handwriting recognition.
\newblock {\em International journal on document analysis and recognition}, 5:39--46, 2002.

\bibitem{aida}
aidapearson.
\newblock Aida calculus math handwriting recognition dataset.
\newblock \url{https://www.kaggle.com/datasets/aidapearson/ocr-data}, 2023.

\bibitem{chng2019art}
Chee~Kheng Chng, Yuliang Liu, Yipeng Sun, Chun~Chet Ng, Canjie Luo, Zihan Ni, ChuanMing Fang, Shuaitao Zhang, Junyu Han, Errui Ding, et~al.
\newblock Icdar2019 robust reading challenge on arbitrary-shaped text-rrc-art.
\newblock In {\em ICDAR}, 2019.

\bibitem{yuan2019ctw}
Tai-Ling Yuan, Zhe Zhu, Kun Xu, Cheng-Jun Li, Tai-Jiang Mu, and Shi-Min Hu.
\newblock A large chinese text dataset in the wild.
\newblock {\em Journal of Computer Science and Technology}, 34:509--521, 2019.

\bibitem{zhang2019rects}
Rui Zhang, Yongsheng Zhou, Qianyi Jiang, Qi~Song, Nan Li, Kai Zhou, Lei Wang, Dong Wang, Minghui Liao, Mingkun Yang, et~al.
\newblock Icdar 2019 robust reading challenge on reading chinese text on signboard.
\newblock In {\em ICDAR}, 2019.

\bibitem{veit2016cocotext}
Andreas Veit, Tomas Matera, Lukas Neumann, Jiri Matas, and Serge Belongie.
\newblock {COCO-Text}: Dataset and benchmark for text detection and recognition in natural images.
\newblock {\em arXiv:1601.07140}, 2016.

\bibitem{yu2023icdar_svrd}
Wenwen Yu, Chengquan Zhang, Haoyu Cao, Wei Hua, Bohan Li, Huang Chen, Mingyu Liu, Mingrui Chen, Jianfeng Kuang, Mengjun Cheng, et~al.
\newblock Icdar 2023 competition on structured text extraction from visually-rich document images.
\newblock In {\em ICDAR}, 2023.

\bibitem{long2023icdar_hiertext}
Shangbang Long, Siyang Qin, Dmitry Panteleev, Alessandro Bissacco, Yasuhisa Fujii, and Michalis Raptis.
\newblock {ICDAR} 2023 competition on hierarchical text detection and recognition.
\newblock In {\em ICDAR}, 2023.

\bibitem{tom2023icdar_roadtext}
George Tom, Minesh Mathew, Sergi Garcia-Bordils, Dimosthenis Karatzas, and CV~Jawahar.
\newblock {ICDAR} 2023 competition on roadtext video text detection, tracking and recognition.
\newblock In {\em ICDAR}, 2023.

\bibitem{li2024icdar_maptext}
Zekun Li, Yijun Lin, Yao-Yi Chiang, Jerod Weinman, Solenn Tual, Joseph Chazalon, Julien Perret, Bertrand Dum{\'e}nieu, and Nathalie Abadie.
\newblock {ICDAR} 2024 competition on historical map text detection, recognition, and linking.
\newblock In {\em ICDAR}, 2024.

\bibitem{captcha}
parasam.
\newblock Captcha dataset.
\newblock \url{https://www.kaggle.com/datasets/parsasam/captcha-dataset}, 2024.

\bibitem{wang2020estvqa}
Xinyu Wang, Yuliang Liu, Chunhua Shen, Chun~Chet Ng, Canjie Luo, Lianwen Jin, Chee~Seng Chan, Anton van~den Hengel, and Liangwei Wang.
\newblock On the general value of evidence, and bilingual scene-text visual question answering.
\newblock In {\em CVPR}, 2020.

\bibitem{tal}
TAL.
\newblock Tal open dataset.
\newblock \url{https://ai.100tal.com/dataset}, 2023.

\bibitem{krishnan2023textstylebrush_Imgur5K}
Praveen Krishnan, Rama Kovvuri, Guan Pang, Boris Vassilev, and Tal Hassner.
\newblock Textstylebrush: transfer of text aesthetics from a single example.
\newblock {\em IEEE Trans. PAMI}, 45(7):9122--9134, 2023.

\bibitem{diem2014icfhr_RAND_CAR}
Markus Diem, Stefan Fiel, Florian Kleber, Robert Sablatnig, Jose~M Saavedra, David Contreras, Juan~Manuel Barrios, and Luiz~S Oliveira.
\newblock Icfhr 2014 competition on handwritten digit string recognition in challenging datasets (hdsrc 2014).
\newblock In {\em International Conference on Frontiers in Handwriting Recognition}, 2014.

\bibitem{mychen76_invoices_receipts_ocr_v1}
mychen76.
\newblock Invoices and receipts ocr v1 dataset.
\newblock \url{https://huggingface.co/datasets/mychen76/invoices-and-receipts_ocr_v1}, 2024.

\bibitem{mouchere2016icfhr2016_chrome_writing}
Harold Mouch{\`e}re, Christian Viard-Gaudin, Richard Zanibbi, and Utpal Garain.
\newblock Icfhr2016 crohme: Competition on recognition of online handwritten mathematical expressions.
\newblock In {\em International Conference on Frontiers in Handwriting Recognition}, 2016.

\bibitem{mishra2012scene_iiit5k}
Anand Mishra, Karteek Alahari, and CV~Jawahar.
\newblock Scene text recognition using higher order language priors.
\newblock In {\em BMVC}, 2012.

\bibitem{ramamoorthy2022memotion}
Sathyanarayanan Ramamoorthy, Nethra Gunti, Shreyash Mishra, S~Suryavardan, Aishwarya Reganti, Parth Patwa, Amitava DaS, Tanmoy Chakraborty, Amit Sheth, Asif Ekbal, et~al.
\newblock Memotion 2: Dataset on sentiment and emotion analysis of memes.
\newblock In {\em De-Factify: Workshop on Multimodal Fact Checking and Hate Speech Detection, CEUR}, 2022.

\bibitem{Azu}
Azu.
\newblock Handwritten-mathematical-expression-convert-latex.
\newblock \url{https://huggingface.co/datasets/Azu/Handwritten-Mathematical-Expression-Convert-LaTeX}, 2023.

\bibitem{xie2022toward_wordart}
Xudong Xie, Ling Fu, Zhifei Zhang, Zhaowen Wang, and Xiang Bai.
\newblock Toward understanding wordart: Corner-guided transformer for scene text recognition.
\newblock In {\em ECCV}, 2022.

\bibitem{wendlerc_renderedtext}
wendlerc.
\newblock Renderedtext dataset.
\newblock \url{https://huggingface.co/datasets/wendlerc/RenderedText}, 2024.

\bibitem{ift_handwriting_forms}
ift.
\newblock Handwriting forms dataset.
\newblock \url{https://huggingface.co/datasets/ift/handwriting_forms}, 2024.

\bibitem{clark2017docqa}
Christopher Clark and Matt Gardner.
\newblock Simple and effective multi-paragraph reading comprehension.
\newblock In {\em ACL}, 2018.

\bibitem{mathew2022infographicvqa}
Minesh Mathew, Viraj Bagal, Rub{\`e}n Tito, Dimosthenis Karatzas, Ernest Valveny, and CV~Jawahar.
\newblock {InfographicVQA}.
\newblock In {\em WACV}, 2022.

\bibitem{singh2019textvqa}
Amanpreet Singh, Vivek Natarajan, Meet Shah, Yu~Jiang, Xinlei Chen, Dhruv Batra, Devi Parikh, and Marcus Rohrbach.
\newblock Towards vqa models that can read.
\newblock In {\em CVPR}, 2019.

\bibitem{li2024multimodal_arxivQA}
Lei Li, Yuqi Wang, Runxin Xu, Peiyi Wang, Xiachong Feng, Lingpeng Kong, and Qi~Liu.
\newblock Multimodal arxiv: A dataset for improving scientific comprehension of large vision-language models.
\newblock {\em arXiv:2403.00231}, 2024.

\bibitem{hsiao2022screenqa}
Yu-Chung Hsiao, Fedir Zubach, Gilles Baechler, Victor Carbune, Jason Lin, Maria Wang, Srinivas Sunkara, Yun Zhu, and Jindong Chen.
\newblock Screenqa: Large-scale question-answer pairs over mobile app screenshots.
\newblock {\em arXiv:2209.08199}, 2022.

\bibitem{mplug_docreason25k}
Anwen Hu, Haiyang Xu, Jiabo Ye, Ming Yan, Liang Zhang, Bo~Zhang, Chen Li, Ji~Zhang, Qin Jin, Fei Huang, et~al.
\newblock mplug-docowl 1.5: Unified structure learning for ocr-free document understanding.
\newblock {\em arXiv:2403.12895}, 2024.

\bibitem{ye2023ureader}
Jiabo Ye, Anwen Hu, Haiyang Xu, Qinghao Ye, Ming Yan, Guohai Xu, Chenliang Li, Junfeng Tian, Qi~Qian, Ji~Zhang, et~al.
\newblock Ureader: Universal ocr-free visually-situated language understanding with multimodal large language model.
\newblock {\em arXiv:2310.05126}, 2023.

\bibitem{Sujet-Finance-QA-Vision-100k}
Hamed~Rahimi Sujet~AI, Allaa~Boutaleb.
\newblock Sujet-finance-qa-vision-100k: A large-scale dataset for financial document vqa.
\newblock https://huggingface.co/datasets/sujet-ai/Sujet-Finance-QA-Vision-100k, 2024.

\bibitem{laurenccon2024building_docmatrix}
Hugo Lauren{\c{c}}on, Andr{\'e}s Marafioti, Victor Sanh, and L{\'e}o Tronchon.
\newblock Building and better understanding vision-language models: insights and future directions.
\newblock {\em arXiv:2408.12637}, 2024.

\bibitem{schwenk2022aokvqa}
Dustin Schwenk, Apoorv Khandelwal, Christopher Clark, Kenneth Marino, and Roozbeh Mottaghi.
\newblock {A-OKVQA}: A benchmark for visual question answering using world knowledge.
\newblock In {\em ECCV}, 2022.

\bibitem{kamizuru00_diagram_image_to_text}
Kamizuru00.
\newblock Diagram image to text dataset.
\newblock \url{https://huggingface.co/datasets/Kamizuru00/diagram_image_to_text}, 2024.

\bibitem{chang2022mapqa}
Shuaichen Chang, David Palzer, Jialin Li, Eric Fosler-Lussier, and Ningchuan Xiao.
\newblock Mapqa: A dataset for question answering on choropleth maps.
\newblock {\em arXiv:2211.08545}, 2022.

\bibitem{mishra2019ocrvqa}
Anand Mishra, Shashank Shekhar, Ajeet~Kumar Singh, and Anirban Chakraborty.
\newblock {OCR-VQA}: Visual question answering by reading text in images.
\newblock In {\em ICDAR}, 2019.

\bibitem{biten2019stvqa}
Ali~Furkan Biten, Ruben Tito, Andres Mafla, Lluis Gomez, Mar{\c{c}}al Rusinol, Ernest Valveny, CV~Jawahar, and Dimosthenis Karatzas.
\newblock Scene text visual question answering.
\newblock In {\em ICCV}, 2019.

\bibitem{tanaka2023slidevqa}
Ryota Tanaka, Kyosuke Nishida, Kosuke Nishida, Taku Hasegawa, Itsumi Saito, and Kuniko Saito.
\newblock Slidevqa: A dataset for document visual question answering on multiple images.
\newblock In {\em AAAI}, 2023.

\bibitem{ding2023PDFvqa}
Yihao Ding, Siwen Luo, Hyunsuk Chung, and Soyeon~Caren Han.
\newblock Vqa: A new dataset for real-world vqa on pdf documents.
\newblock In {\em Joint European Conference on Machine Learning and Knowledge Discovery in Databases}, 2023.

\bibitem{mahamoud2024chic_vqa_cd}
Ibrahim~Souleiman Mahamoud, Micka{\"e}l Coustaty, Aur{\'e}lie Joseph, Vincent~Poulain d’Andecy, and Jean-Marc Ogier.
\newblock Chic: Corporate document for visual question answering.
\newblock In {\em ICDAR}, 2024.

\bibitem{shreyanshu09_block_diagram}
shreyanshu09.
\newblock Block diagram dataset.
\newblock \url{https://huggingface.co/datasets/shreyanshu09/Block_Diagram}, 2024.

\bibitem{tang2024mtvqa}
Jingqun Tang, Qi~Liu, Yongjie Ye, Jinghui Lu, Shu Wei, Chunhui Lin, Wanqing Li, Mohamad Fitri Faiz~Bin Mahmood, Hao Feng, Zhen Zhao, et~al.
\newblock Mtvqa: Benchmarking multilingual text-centric visual question answering.
\newblock {\em arXiv:2405.11985}, 2024.

\bibitem{faysse2024colpali}
Manuel Faysse, Hugues Sibille, Tony Wu, Gautier Viaud, C{\'e}line Hudelot, and Pierre Colombo.
\newblock Colpali: Efficient document retrieval with vision language models.
\newblock {\em arXiv:2407.01449}, 2024.

\bibitem{mathew2021asking_benthamqa}
Minesh Mathew, Lluis Gomez, Dimosthenis Karatzas, and CV~Jawahar.
\newblock Asking questions on handwritten document collections.
\newblock {\em International Journal on Document Analysis and Recognition}, 24(3):235--249, 2021.

\bibitem{acharya2019tallyqa}
Manoj Acharya, Kushal Kafle, and Christopher Kanan.
\newblock {TallyQA}: Answering complex counting questions.
\newblock In {\em AAAI}, 2019.

\bibitem{tu2023many}
Haoqin Tu, Chenhang Cui, Zijun Wang, Yiyang Zhou, Bingchen Zhao, Junlin Han, Wangchunshu Zhou, Huaxiu Yao, and Cihang Xie.
\newblock How many unicorns are in this image? a safety evaluation benchmark for vision llms.
\newblock {\em arXiv:2311.16101}, 2023.

\bibitem{yu2016refcoco}
Licheng Yu, Patrick Poirson, Shan Yang, Alexander~C Berg, and Tamara~L Berg.
\newblock Modeling context in referring expressions.
\newblock In {\em ECCV}, 2016.

\bibitem{mao2016refcocog}
Junhua Mao, Jonathan Huang, Alexander Toshev, Oana Camburu, Alan~L Yuille, and Kevin Murphy.
\newblock Generation and comprehension of unambiguous object descriptions.
\newblock In {\em CVPR}, 2016.

\bibitem{zheng2024agentstudio_groundui}
Longtao Zheng, Zhiyuan Huang, Zhenghai Xue, Xinrun Wang, Bo~An, and Shuicheng Yan.
\newblock {AgentStudio}: A toolkit for building general virtual agents.
\newblock {\em arXiv:2403.17918}, 2024.

\bibitem{wang2023lvisinstruct4v}
Junke Wang, Lingchen Meng, Zejia Weng, Bo~He, Zuxuan Wu, and Yu-Gang Jiang.
\newblock To see is to believe: Prompting gpt-4v for better visual instruction tuning.
\newblock {\em arXiv:2311.07574}, 2023.

\bibitem{chen2024allava}
Guiming~Hardy Chen, Shunian Chen, Ruifei Zhang, Junying Chen, Xiangbo Wu, Zhiyi Zhang, Zhihong Chen, Jianquan Li, Xiang Wan, and Benyou Wang.
\newblock Allava: Harnessing gpt4v-synthesized data for a lite vision-language model.
\newblock {\em arXiv:2402.11684}, 2024.

\bibitem{laion_gpt4v_dataset}
LAION.
\newblock Gpt-4v dataset.
\newblock \url{https://huggingface.co/datasets/laion/gpt4v-dataset}, 2023.

\bibitem{zhang2023llavar}
Yanzhe Zhang, Ruiyi Zhang, Jiuxiang Gu, Yufan Zhou, Nedim Lipka, Diyi Yang, and Tong Sun.
\newblock {LLaVAR}: Enhanced visual instruction tuning for text-rich image understanding.
\newblock {\em arXiv:2306.17107}, 2023.

\bibitem{gurari2018vizwiz}
Danna Gurari, Qing Li, Abigale~J Stangl, Anhong Guo, Chi Lin, Kristen Grauman, Jiebo Luo, and Jeffrey~P Bigham.
\newblock {VizWiz Grand Challenge}: Answering visual questions from blind people.
\newblock In {\em CVPR}, 2018.

\bibitem{cha2024visually}
Sungguk Cha, Jusung Lee, Younghyun Lee, and Cheoljong Yang.
\newblock Visually dehallucinative instruction generation.
\newblock In {\em ICASSP}, 2024.

\bibitem{pont2020connecting_lnqa}
Jordi Pont-Tuset, Jasper Uijlings, Soravit Changpinyo, Radu Soricut, and Vittorio Ferrari.
\newblock Connecting vision and language with localized narratives.
\newblock In {\em ECCV}, 2020.

\bibitem{fastjob_visual_emotional_analysis}
FastJobs.
\newblock Visual emotional analysis dataset.
\newblock \url{https://huggingface.co/datasets/FastJobs/Visual_Emotional_Analysis}, 2024.

\bibitem{yang2019spatialsense}
Kaiyu Yang, Olga Russakovsky, and Jia Deng.
\newblock Spatialsense: An adversarially crowdsourced benchmark for spatial relation recognition.
\newblock In {\em ICCV}, 2019.

\bibitem{keremberke_indoor_scene_classification}
keremberke.
\newblock Indoor scene classification dataset.
\newblock \url{https://huggingface.co/datasets/keremberke/indoor-scene-classification}, 2024.

\bibitem{zhou2017places365}
Bolei Zhou, Agata Lapedriza, Aditya Khosla, Aude Oliva, and Antonio Torralba.
\newblock Places: A 10 million image database for scene recognition.
\newblock {\em IEEE Trans. PAMI}, 40(6):1452--1464, 2017.

\bibitem{liu2024mminstruct}
Yangzhou Liu, Yue Cao, Zhangwei Gao, Weiyun Wang, Zhe Chen, Wenhai Wang, Hao Tian, Lewei Lu, Xizhou Zhu, Tong Lu, et~al.
\newblock Mminstruct: A high-quality multi-modal instruction tuning dataset with extensive diversity.
\newblock {\em arXiv:2407.15838}, 2024.

\bibitem{sima2023drivelm}
Chonghao Sima, Katrin Renz, Kashyap Chitta, Li~Chen, Hanxue Zhang, Chengen Xie, Jens Bei{\ss}wenger, Ping Luo, Andreas Geiger, and Hongyang Li.
\newblock Drivelm: Driving with graph visual question answering.
\newblock {\em arXiv:2312.14150}, 2023.

\bibitem{nandy2024yesbut}
Abhilash Nandy, Yash Agarwal, Ashish Patwa, Millon~Madhur Das, Aman Bansal, Ankit Raj, Pawan Goyal, and Niloy Ganguly.
\newblock Yesbut: A high-quality annotated multimodal dataset for evaluating satire comprehension capability of vision-language models.
\newblock {\em arXiv:2409.13592}, 2024.

\bibitem{lu2024wildvision}
Yujie Lu, Dongfu Jiang, Wenhu Chen, William~Yang Wang, Yejin Choi, and Bill~Yuchen Lin.
\newblock Wildvision: Evaluating vision-language models in the wild with human preferences.
\newblock {\em arXiv:2406.11069}, 2024.

\bibitem{xiong2024llava_critic}
Tianyi Xiong, Xiyao Wang, Dong Guo, Qinghao Ye, Haoqi Fan, Quanquan Gu, Heng Huang, and Chunyuan Li.
\newblock Llava-critic: Learning to evaluate multimodal models.
\newblock {\em arXiv:2410.02712}, 2024.

\bibitem{yu2024rlaif_v}
Tianyu Yu, Haoye Zhang, Yuan Yao, Yunkai Dang, Da~Chen, Xiaoman Lu, Ganqu Cui, Taiwen He, Zhiyuan Liu, Tat-Seng Chua, et~al.
\newblock Rlaif-v: Aligning mllms through open-source ai feedback for super gpt-4v trustworthiness.
\newblock {\em arXiv:2405.17220}, 2024.

\bibitem{goyal2017vqav2}
Yash Goyal, Tejas Khot, Douglas Summers-Stay, Dhruv Batra, and Devi Parikh.
\newblock Making the v in vqa matter: Elevating the role of image understanding in visual question answering.
\newblock In {\em CVPR}, 2017.

\bibitem{wu2024mmra}
Siwei Wu, Kang Zhu, Yu~Bai, Yiming Liang, Yizhi Li, Haoning Wu, Jiaheng Liu, Ruibo Liu, Xingwei Qu, Xuxin Cheng, et~al.
\newblock Mmra: A benchmark for multi-granularity multi-image relational association.
\newblock {\em arXiv:2407.17379}, 2024.

\bibitem{hosu2020koniq}
Vlad Hosu, Hanhe Lin, Tamas Sziranyi, and Dietmar Saupe.
\newblock {KonIQ-10k}: An ecologically valid database for deep learning of blind image quality assessment.
\newblock {\em IEEE Trans. Image Processing}, 29:4041--4056, 2020.

\bibitem{liu2024mmdu}
Ziyu Liu, Tao Chu, Yuhang Zang, Xilin Wei, Xiaoyi Dong, Pan Zhang, Zijian Liang, Yuanjun Xiong, Yu~Qiao, Dahua Lin, et~al.
\newblock Mmdu: A multi-turn multi-image dialog understanding benchmark and instruction-tuning dataset for lvlms.
\newblock {\em arXiv:2406.11833}, 2024.

\bibitem{jhamtani2018learning_spotthediff}
Harsh Jhamtani and Taylor Berg-Kirkpatrick.
\newblock Learning to describe differences between pairs of similar images.
\newblock {\em arXiv:1808.10584}, 2018.

\bibitem{kiela2020hateful_memes}
Douwe Kiela, Hamed Firooz, Aravind Mohan, Vedanuj Goswami, Amanpreet Singh, Pratik Ringshia, and Davide Testuggine.
\newblock The hateful memes challenge: Detecting hate speech in multimodal memes.
\newblock {\em NeurIPS}, 2020.

\bibitem{ren2015exploring_cocoqa}
Mengye Ren, Ryan Kiros, and Richard Zemel.
\newblock Exploring models and data for image question answering.
\newblock {\em NeurIPS}, 2015.

\bibitem{suhr2017corpus_nlvr2}
Alane Suhr, Mike Lewis, James Yeh, and Yoav Artzi.
\newblock A corpus of natural language for visual reasoning.
\newblock In {\em ACL}, 2017.

\bibitem{laurenccon2024matters_mimic_cgd}
Hugo Lauren{\c{c}}on, L{\'e}o Tronchon, Matthieu Cord, and Victor Sanh.
\newblock What matters when building vision-language models?
\newblock {\em arXiv:2405.02246}, 2024.

\bibitem{belouadi2023automatikz_datikz}
Jonas Belouadi, Anne Lauscher, and Steffen Eger.
\newblock Automatikz: Text-guided synthesis of scientific vector graphics with tikz.
\newblock {\em arXiv:2310.00367}, 2023.

\bibitem{emo_visual_data_chinese_meme}
LLM-Red-Team Contributors.
\newblock {emo-visual-data}: Emotion and visual data analysis project.
\newblock \url{https://github.com/LLM-Red-Team/emo-visual-data}, 2024.

\bibitem{lu2021iconqa}
Pan Lu, Liang Qiu, Jiaqi Chen, Tony Xia, Yizhou Zhao, Wei Zhang, Zhou Yu, Xiaodan Liang, and Song-Chun Zhu.
\newblock {IconQA}: A new benchmark for abstract diagram understanding and visual language reasoning.
\newblock {\em arXiv:2110.13214}, 2021.

\bibitem{laurenccon2024unlocking_websight}
Hugo Lauren{\c{c}}on, L{\'e}o Tronchon, and Victor Sanh.
\newblock Unlocking the conversion of web screenshots into html code with the websight dataset.
\newblock {\em arXiv:2403.09029}, 2024.

\bibitem{lian2023openorca}
W~Lian, B~Goodson, E~Pentland, et~al.
\newblock {OpenOrca}: An open dataset of gpt augmented flan reasoning traces, 2023.

\bibitem{mitra2024orca}
Arindam Mitra, Hamed Khanpour, Corby Rosset, and Ahmed Awadallah.
\newblock {Orca-Math}: Unlocking the potential of slms in grade school math.
\newblock {\em arXiv:2402.14830}, 2024.

\bibitem{zheng2024opencodeinterpreter}
Tianyu Zheng, Ge~Zhang, Tianhao Shen, Xueling Liu, Bill~Yuchen Lin, Jie Fu, Wenhu Chen, and Xiang Yue.
\newblock {OpenCodeInterpreter}: Integrating code generation with execution and refinement.
\newblock {\em arXiv:2402.14658}, 2024.

\bibitem{yue2023mammoth_mathinstruct}
Xiang Yue, Xingwei Qu, Ge~Zhang, Yao Fu, Wenhao Huang, Huan Sun, Yu~Su, and Wenhu Chen.
\newblock Mammoth: Building math generalist models through hybrid instruction tuning.
\newblock {\em arXiv:2309.05653}, 2023.

\bibitem{xu2023wizardlm}
Can Xu, Qingfeng Sun, Kai Zheng, Xiubo Geng, Pu~Zhao, Jiazhan Feng, Chongyang Tao, and Daxin Jiang.
\newblock Wizardlm: Empowering large language models to follow complex instructions.
\newblock {\em arXiv:2304.12244}, 2023.

\bibitem{chen2023theoremqa}
Wenhu Chen, Ming Yin, Max Ku, Pan Lu, Yixin Wan, Xueguang Ma, Jianyu Xu, Xinyi Wang, and Tony Xia.
\newblock Theoremqa: A theorem-driven question answering dataset.
\newblock In {\em EMNLP}, 2023.

\bibitem{OpenHermes2_5}
Teknium.
\newblock Openhermes 2.5: An open dataset of synthetic data for generalist llm assistants.
\newblock \url{https://huggingface.co/datasets/teknium/OpenHermes-2.5}, 2023.

\bibitem{numina_math_datasets}
Jia LI, Edward Beeching, Lewis Tunstall, Ben Lipkin, Roman Soletskyi, Shengyi~Costa Huang, Kashif Rasul, Longhui Yu, Albert Jiang, Ziju Shen, Zihan Qin, Bin Dong, Li~Zhou, Yann Fleureau, Guillaume Lample, and Stanislas Polu.
\newblock Numinamath.
\newblock \url{https://huggingface.co/datasets/AI-MO/NuminaMath-CoT}, 2024.

\bibitem{flytech_python_codes_25k}
FLOCK4H.
\newblock Python codes 25k dataset.
\newblock \url{https://huggingface.co/datasets/flytech/python-codes-25k}, 2024.

\bibitem{baai_infinity_instruct}
BAAI.
\newblock Infinity-instruct dataset.
\newblock \url{https://huggingface.co/datasets/BAAI/Infinity-Instruct}, 2024.

\bibitem{iamtarun_python_code_instructions_18k_alpaca}
Tarun Bisht.
\newblock Python code instructions 18k alpaca dataset.
\newblock \url{https://huggingface.co/datasets/iamtarun/python_code_instructions_18k_alpaca}, 2024.

\bibitem{looksjuicy_ruozhiba}
LooksJuicy.
\newblock Ruozhiba dataset.
\newblock \url{https://huggingface.co/datasets/LooksJuicy/ruozhiba}, 2024.

\bibitem{zhang2024infinitymath}
Bo-Wen Zhang, Yan Yan, Lin Li, and Guang Liu.
\newblock Infinitymath: A scalable instruction tuning dataset in programmatic mathematical reasoning.
\newblock In {\em ACM International Conference on Information and Knowledge Management}, 2024.

\bibitem{lai2024stepDPO}
Xin Lai, Zhuotao Tian, Yukang Chen, Senqiao Yang, Xiangru Peng, and Jiaya Jia.
\newblock {Step-DPO}: Step-wise preference optimization for long-chain reasoning of llms.
\newblock {\em arXiv:2406.18629}, 2024.

\bibitem{zhang2024tablellm}
Xiaokang Zhang, Jing Zhang, Zeyao Ma, Yang Li, Bohan Zhang, Guanlin Li, Zijun Yao, Kangli Xu, Jinchang Zhou, Daniel Zhang-Li, et~al.
\newblock {TableLLM}: Enabling tabular data manipulation by llms in real office usage scenarios.
\newblock {\em arXiv:2403.19318}, 2024.

\bibitem{yuan2024advancing_ultrainteract}
Lifan Yuan, Ganqu Cui, Hanbin Wang, Ning Ding, Xingyao Wang, Jia Deng, Boji Shan, Huimin Chen, Ruobing Xie, Yankai Lin, et~al.
\newblock Advancing llm reasoning generalists with preference trees.
\newblock {\em arXiv:2404.02078}, 2024.

\bibitem{sharma2018cc3m}
Piyush Sharma, Nan Ding, Sebastian Goodman, and Radu Soricut.
\newblock Conceptual captions: A cleaned, hypernymed, image alt-text dataset for automatic image captioning.
\newblock In {\em ACL}, 2018.

\bibitem{sidorov2020textcaps}
Oleksii Sidorov, Ronghang Hu, Marcus Rohrbach, and Amanpreet Singh.
\newblock Textcaps: a dataset for image captioning with reading comprehension.
\newblock In {\em ECCV}, 2020.

\bibitem{chen2023sharegpt4v}
Lin Chen, Jisong Li, Xiaoyi Dong, Pan Zhang, Conghui He, Jiaqi Wang, Feng Zhao, and Dahua Lin.
\newblock Sharegpt4v: Improving large multi-modal models with better captions.
\newblock {\em arXiv:2311.12793}, 2023.

\bibitem{li2024densefusion}
Xiaotong Li, Fan Zhang, Haiwen Diao, Yueze Wang, Xinlong Wang, and Ling-Yu Duan.
\newblock Densefusion-1m: Merging vision experts for comprehensive multimodal perception.
\newblock {\em arXiv:2407.08303}, 2024.

\bibitem{shao2019objects365}
Shuai Shao, Zeming Li, Tianyuan Zhang, Chao Peng, Gang Yu, Xiangyu Zhang, Jing Li, and Jian Sun.
\newblock Objects365: A large-scale, high-quality dataset for object detection.
\newblock In {\em ICCV}, 2019.

\bibitem{toshniwal2024openmathinstruct}
Shubham Toshniwal, Ivan Moshkov, Sean Narenthiran, Daria Gitman, Fei Jia, and Igor Gitman.
\newblock Openmathinstruct-1: A 1.8 million math instruction tuning dataset.
\newblock {\em arXiv:2402.10176}, 2024.

\bibitem{pizzi2022sscd}
Ed~Pizzi, Sreya~Dutta Roy, Sugosh~Nagavara Ravindra, Priya Goyal, and Matthijs Douze.
\newblock A self-supervised descriptor for image copy detection.
\newblock In {\em CVPR}, 2022.

\bibitem{all_mpnet_base_v2}
Sentence-Transformers.
\newblock All-minilm-l6-v2.
\newblock \url{https://huggingface.co/sentence-transformers/all-mpnet-base-v2}, 2023.

\bibitem{liu2023ocrbench}
Yuliang Liu, Zhang Li, Hongliang Li, Wenwen Yu, Mingxin Huang, Dezhi Peng, Mingyu Liu, Mingrui Chen, Chunyuan Li, Lianwen Jin, et~al.
\newblock On the hidden mystery of ocr in large multimodal models.
\newblock {\em arXiv:2305.07895}, 2023.

\bibitem{zheng2024llamafactory}
Yaowei Zheng, Richong Zhang, Junhao Zhang, Yanhan Ye, Zheyan Luo, Zhangchi Feng, and Yongqiang Ma.
\newblock {LlamaFactory}: Unified efficient fine-tuning of 100+ language models.
\newblock {\em arXiv:2403.13372}, 2024.

\bibitem{krell2021efficient}
Mario~Michael Krell, Matej Kosec, Sergio~P Perez, and Andrew Fitzgibbon.
\newblock Efficient sequence packing without cross-contamination: Accelerating large language models without impacting performance.
\newblock {\em arXiv:2107.02027}, 2021.

\bibitem{cherti2023reproducible}
Mehdi Cherti, Romain Beaumont, Ross Wightman, Mitchell Wortsman, Gabriel Ilharco, Cade Gordon, Christoph Schuhmann, Ludwig Schmidt, and Jenia Jitsev.
\newblock Reproducible scaling laws for contrastive language-image learning.
\newblock In {\em CVPR}, 2023.

\bibitem{openai2023gpt4o}
{OpenAI}.
\newblock Hello {GPT-4o}.
\newblock \url{https://openai.com/index/hello-gpt-4o/}, 2023.
\newblock Accessed: 2024-11-12.

\bibitem{reid2024gemini1_5}
Machel Reid, Nikolay Savinov, Denis Teplyashin, Dmitry Lepikhin, Timothy Lillicrap, Jean-baptiste Alayrac, Radu Soricut, Angeliki Lazaridou, Orhan Firat, Julian Schrittwieser, et~al.
\newblock Gemini 1.5: Unlocking multimodal understanding across millions of tokens of context.
\newblock {\em arXiv:2403.05530}, 2024.

\bibitem{hu2024minicpm}
Shengding Hu, Yuge Tu, Xu~Han, Chaoqun He, Ganqu Cui, Xiang Long, Zhi Zheng, Yewei Fang, Yuxiang Huang, Weilin Zhao, et~al.
\newblock Minicpm: Unveiling the potential of small language models with scalable training strategies.
\newblock {\em arXiv:2404.06395}, 2024.

\bibitem{OpenAI_ChatGPT}
OpenAI.
\newblock Chatgpt: Optimizing language models for dialogue.
\newblock \url{https://openai.com/blog/chatgpt}, 2023.
\newblock Accessed: 2023.

\bibitem{ouyang2022training}
Long Ouyang, Jeffrey Wu, Xu~Jiang, Diogo Almeida, Carroll Wainwright, Pamela Mishkin, Chong Zhang, Sandhini Agarwal, Katarina Slama, Alex Ray, et~al.
\newblock Training language models to follow instructions with human feedback.
\newblock {\em NeurIPS}, 2022.

\bibitem{li2023blip2}
Junnan Li, Dongxu Li, Silvio Savarese, and Steven Hoi.
\newblock Blip-2: Bootstrapping language-image pre-training with frozen image encoders and large language models.
\newblock In {\em ICML}, 2023.

\bibitem{zhu2023minigpt4}
Deyao Zhu, Jun Chen, Xiaoqian Shen, Xiang Li, and Mohamed Elhoseiny.
\newblock Minigpt-4: Enhancing vision-language understanding with advanced large language models.
\newblock In {\em ICLR}, 2024.

\bibitem{wang2023cogvlm}
Weihan Wang, Qingsong Lv, Wenmeng Yu, Wenyi Hong, Ji~Qi, Yan Wang, Junhui Ji, Zhuoyi Yang, Lei Zhao, Xixuan Song, et~al.
\newblock {CogVLM}: Visual expert for pretrained language models.
\newblock {\em arXiv:2311.03079}, 2023.

\bibitem{li2023otter}
Bo~Li, Yuanhan Zhang, Liangyu Chen, Jinghao Wang, Jingkang Yang, and Ziwei Liu.
\newblock Otter: A multi-modal model with in-context instruction tuning.
\newblock {\em arXiv:2305.03726}, 2023.

\bibitem{liu2024llavanext}
Haotian Liu, Chunyuan Li, Yuheng Li, Bo~Li, Yuanhan Zhang, Sheng Shen, and Yong~Jae Lee.
\newblock Llava-next: Improved reasoning, ocr, and world knowledge, January 2024.

\bibitem{yao2024minicpm}
Yuan Yao, Tianyu Yu, Ao~Zhang, Chongyi Wang, Junbo Cui, Hongji Zhu, Tianchi Cai, Haoyu Li, Weilin Zhao, Zhihui He, et~al.
\newblock {MiniCPM-V}: A gpt-4v level mllm on your phone.
\newblock {\em arXiv:2408.01800}, 2024.

\bibitem{beyer2024paligemma}
Lucas Beyer, Andreas Steiner, Andr{\'e}~Susano Pinto, Alexander Kolesnikov, Xiao Wang, Daniel Salz, Maxim Neumann, Ibrahim Alabdulmohsin, Michael Tschannen, Emanuele Bugliarello, et~al.
\newblock Paligemma: A versatile 3b vlm for transfer.
\newblock {\em arXiv:2407.07726}, 2024.

\bibitem{dai2024nvlm}
Wenliang Dai, Nayeon Lee, Boxin Wang, Zhuoling Yang, Zihan Liu, Jon Barker, Tuomas Rintamaki, Mohammad Shoeybi, Bryan Catanzaro, and Wei Ping.
\newblock {NVLM}: Open frontier-class multimodal llms.
\newblock {\em arXiv:2409.11402}, 2024.

\bibitem{lin2023vila}
Ji~Lin, Hongxu Yin, Wei Ping, Yao Lu, Pavlo Molchanov, Andrew Tao, Huizi Mao, Jan Kautz, Mohammad Shoeybi, and Song Han.
\newblock {VILA}: On pre-training for visual language models, 2023.

\bibitem{fang2024vila}
Yunhao Fang, Ligeng Zhu, Yao Lu, Yan Wang, Pavlo Molchanov, Jan Kautz, Jang~Hyun Cho, Marco Pavone, Song Han, and Hongxu Yin.
\newblock {VILA2}: Vila augmented vila.
\newblock {\em arXiv:2407.17453}, 2024.

\bibitem{liu2024nvila}
Zhijian Liu, Ligeng Zhu, Baifeng Shi, Zhuoyang Zhang, Yuming Lou, Shang Yang, Haocheng Xi, Shiyi Cao, Yuxian Gu, Dacheng Li, et~al.
\newblock {NVILA}: Efficient frontier visual language models.
\newblock {\em arXiv:2412.04468}, 2024.

\bibitem{idefics2023}
{IDEFICS}.
\newblock Introducing idefics: An open reproduction of state-of-the-art visual language model.
\newblock \url{https://huggingface.co/blog/idefics}, 2023.

\bibitem{radford2021clip}
Alec Radford, Jong~Wook Kim, Chris Hallacy, Aditya Ramesh, Gabriel Goh, Sandhini Agarwal, Girish Sastry, Amanda Askell, Pamela Mishkin, Jack Clark, et~al.
\newblock Learning transferable visual models from natural language supervision.
\newblock In {\em ICML}, 2021.

\bibitem{openclip}
Gabriel Ilharco, Mitchell Wortsman, Ross Wightman, Cade Gordon, Nicholas Carlini, Rohan Taori, Achal Dave, Vaishaal Shankar, Hongseok Namkoong, John Miller, Hannaneh Hajishirzi, Ali Farhadi, and Ludwig Schmidt.
\newblock Openclip.
\newblock Zenodo. Version 0.1. \url{https://doi.org/10.5281/zenodo.5143773}, 2021.
\newblock DOI: 10.5281/zenodo.5143773.

\bibitem{sun2023evaclip}
Quan Sun, Yuxin Fang, Ledell Wu, Xinlong Wang, and Yue Cao.
\newblock {EVA-Clip}: Improved training techniques for clip at scale.
\newblock {\em arXiv:2303.15389}, 2023.

\bibitem{chen2023internvl}
Zhe Chen, Jiannan Wu, Wenhai Wang, Weijie Su, Guo Chen, Sen Xing, Zhong Muyan, Qinglong Zhang, Xizhou Zhu, Lewei Lu, et~al.
\newblock {InternVL}: Scaling up vision foundation models and aligning for generic visual-linguistic tasks.
\newblock {\em arXiv:2312.14238}, 2023.

\bibitem{xiao2024florence}
Bin Xiao, Haiping Wu, Weijian Xu, Xiyang Dai, Houdong Hu, Yumao Lu, Michael Zeng, Ce~Liu, and Lu~Yuan.
\newblock Florence-2: Advancing a unified representation for a variety of vision tasks.
\newblock In {\em CVPR}, 2024.

\bibitem{ranzinger2024radio}
Mike Ranzinger, Greg Heinrich, Jan Kautz, and Pavlo Molchanov.
\newblock {AM-RADIO}: Agglomerative vision foundation model reduce all domains into one.
\newblock In {\em CVPR}, 2024.

\bibitem{heinrich2024radio}
Greg Heinrich, Mike Ranzinger, Yao Lu, Jan Kautz, Andrew Tao, Bryan Catanzaro, Pavlo Molchanov, et~al.
\newblock {RADIO Amplified}: Improved baselines for agglomerative vision foundation models.
\newblock {\em arXiv:2412.07679}, 2024.

\bibitem{lin2023sphinx}
Ziyi Lin, Chris Liu, Renrui Zhang, Peng Gao, Longtian Qiu, Han Xiao, Han Qiu, Chen Lin, Wenqi Shao, Keqin Chen, et~al.
\newblock Sphinx: The joint mixing of weights, tasks, and visual embeddings for multi-modal large language models.
\newblock {\em arXiv:2311.07575}, 2023.

\bibitem{liu2024prismer}
Shikun Liu, Linxi Fan, Edward Johns, Zhiding Yu, Chaowei Xiao, and Anima Anandkumar.
\newblock Prismer: A vision-language model with an ensemble of experts.
\newblock {\em TMLR}, 2024.

\bibitem{karamcheti2024prismatic}
Siddharth Karamcheti, Suraj Nair, Ashwin Balakrishna, Percy Liang, Thomas Kollar, and Dorsa Sadigh.
\newblock {Prismatic VLMs}: Investigating the design space of visually-conditioned language models.
\newblock {\em arXiv:2402.07865}, 2024.

\bibitem{li2024miniGemini}
Yanwei Li, Yuechen Zhang, Chengyao Wang, Zhisheng Zhong, Yixin Chen, Ruihang Chu, Shaoteng Liu, and Jiaya Jia.
\newblock Mini-gemini: Mining the potential of multi-modality vision language models.
\newblock {\em arXiv:2403.18814}, 2024.

\bibitem{luo2024llava_hr}
Gen Luo, Yiyi Zhou, Yuxin Zhang, Xiawu Zheng, Xiaoshuai Sun, and Rongrong Ji.
\newblock Feast your eyes: Mixture-of-resolution adaptation for multimodal large language models.
\newblock {\em arXiv:2403.03003}, 2024.

\bibitem{chen2023palix}
Xi~Chen, Josip Djolonga, Piotr Padlewski, Basil Mustafa, Soravit Changpinyo, Jialin Wu, Carlos~Riquelme Ruiz, Sebastian Goodman, Xiao Wang, Yi~Tay, et~al.
\newblock {PaLI-X}: On scaling up a multilingual vision and language model.
\newblock {\em arXiv:2305.18565}, 2023.

\bibitem{chen2023pali3}
Xi~Chen, Xiao Wang, Lucas Beyer, Alexander Kolesnikov, Jialin Wu, Paul Voigtlaender, Basil Mustafa, Sebastian Goodman, Ibrahim Alabdulmohsin, Piotr Padlewski, et~al.
\newblock {PaLI-3} vision language models: Smaller, faster, stronger.
\newblock {\em arXiv:2310.09199}, 2023.

\bibitem{li2023monkey}
Zhang Li, Biao Yang, Qiang Liu, Zhiyin Ma, Shuo Zhang, Jingxu Yang, Yabo Sun, Yuliang Liu, and Xiang Bai.
\newblock Monkey: Image resolution and text label are important things for large multi-modal models.
\newblock {\em arXiv:2311.06607}, 2023.

\bibitem{xu2024llava_uhd}
Ruyi Xu, Yuan Yao, Zonghao Guo, Junbo Cui, Zanlin Ni, Chunjiang Ge, Tat-Seng Chua, Zhiyuan Liu, Maosong Sun, and Gao Huang.
\newblock Llava-uhd: an lmm perceiving any aspect ratio and high-resolution images.
\newblock {\em arXiv:2403.11703}, 2024.

\bibitem{dong2024xc24khd}
Xiaoyi Dong, Pan Zhang, Yuhang Zang, Yuhang Cao, Bin Wang, Linke Ouyang, Songyang Zhang, Haodong Duan, Wenwei Zhang, Yining Li, et~al.
\newblock {InternLM-XComposer2-4KHD}: A pioneering large vision-language model handling resolutions from 336 pixels to {4K} {HD}.
\newblock {\em arXiv:2404.06512}, 2024.

\bibitem{liu2023improved}
Haotian Liu, Chunyuan Li, Yuheng Li, and Yong~Jae Lee.
\newblock Improved baselines with visual instruction tuning.
\newblock {\em arXiv:2310.03744}, 2023.

\bibitem{instructblip}
Wenliang Dai, Junnan Li, Dongxu Li, Anthony Meng~Huat Tiong, Junqi Zhao, Weisheng Wang, Boyang Li, Pascale~N Fung, and Steven Hoi.
\newblock Instructblip: Towards general-purpose vision-language models with instruction tuning.
\newblock {\em NeurIPS}, 36, 2024.

\bibitem{li2023mimicit}
Bo~Li, Yuanhan Zhang, Liangyu Chen, Jinghao Wang, Fanyi Pu, Jingkang Yang, Chunyuan Li, and Ziwei Liu.
\newblock {MIMIC-IT}: Multi-modal in-context instruction tuning.
\newblock 2023.

\bibitem{li2024omnicorpus}
Qingyun Li, Zhe Chen, Weiyun Wang, Wenhai Wang, Shenglong Ye, Zhenjiang Jin, et~al.
\newblock {OmniCorpus}: A unified multimodal corpus of 10 billion-level images interleaved with text.
\newblock {\em arXiv:2406.08418}, 2024.

\bibitem{awadalla2024mint1t}
Anas Awadalla, Le~Xue, Oscar Lo, Manli Shu, Hannah Lee, Etash~Kumar Guha, Matt Jordan, Sheng Shen, Mohamed Awadalla, Silvio Savarese, Caiming Xiong, Ran Xu, Yejin Choi, and Ludwig Schmidt.
\newblock {MINT-1T}: Scaling open-source multimodal data by 10x: A multimodal dataset with one trillion tokens.
\newblock 2024.

\bibitem{deitke2024molmo}
Matt Deitke, Christopher Clark, Sangho Lee, Rohun Tripathi, Yue Yang, Jae~Sung Park, Mohammadreza Salehi, Niklas Muennighoff, Kyle Lo, Luca Soldaini, et~al.
\newblock Molmo and pixmo: Open weights and open data for state-of-the-art multimodal models.
\newblock {\em arXiv:2409.17146}, 2024.

\bibitem{bai2024survey}
Tianyi Bai, Hao Liang, Binwang Wan, Yanran Xu, Xi~Li, Shiyu Li, Ling Yang, Bozhou Li, Yifan Wang, Bin Cui, et~al.
\newblock A survey of multimodal large language model from a data-centric perspective.
\newblock {\em arXiv:2405.16640}, 2024.

\bibitem{yin2023survey}
Shukang Yin, Chaoyou Fu, Sirui Zhao, Ke~Li, Xing Sun, Tong Xu, and Enhong Chen.
\newblock A survey on multimodal large language models.
\newblock {\em arXiv:2306.13549}, 2023.

\bibitem{wu2023multimodal}
Jiayang Wu, Wensheng Gan, Zefeng Chen, Shicheng Wan, and S~Yu Philip.
\newblock Multimodal large language models: A survey.
\newblock In {\em 2023 IEEE International Conference on Big Data}, 2023.

\end{thebibliography}
}
\clearpage
\appendix
\setcounter{page}{1}
\setcounter{figure}{0}
\setcounter{table}{0}
\setcounter{section}{0}
\renewcommand{\thesection}{\Alph{section}}
\renewcommand{\thetable}{\Alph{table}}
\renewcommand{\thefigure}{\Alph{figure}}

\section{Computing}
We show our training resource for Eagle2-9B in Tab.~\ref{tab:app_gpu}.
In actual development, we rarely iterate the Stage-1 model. Usually, we iterate Stage-1.5 once after iterating Stage-2 $> 10$ times.
\begin{table}[tbh]
    \centering
    \resizebox{\linewidth}{!}{%
    \begin{tabular}{@{}ll|c|c|c@{}}
    \toprule
    & & \textbf{Stage-1} & \textbf{Stage-1.5} & \textbf{Stage-2} \\ 
    \midrule 
    \multirow{2}{*}{Qwen-2-9B}
    & \textbf{GPUs} & H100$\times$128 & H100$\times$256 & H100$\times$256\\
    & Hours & 2.5 & 28 & 6\\
    \bottomrule
    \end{tabular}}
    \caption{ Training Information of Eagle2-9B.}
    \label{tab:app_gpu}
    \end{table}
\begin{table}[h!]
\centering
\renewcommand{\arraystretch}{1.2}
\resizebox{0.94\linewidth}{!}{%
\begin{tabular}{l|c|c|c}
Dataset & Category & \#Stage-1.5 & \#Stage-2\\
\hline
LLaVa-150K-EN & General & 158K & 57K\\
LLaVa-150K-CN & General & 158K & 50K\\
LVIS-Instruct4V & General & 223K & 12K\\
ALLaVa-laion & General & 505K & 20K\\
ALLaVa-vflan & General & 202K & 26K\\
Laion-GPT4V & General & 11K & 11K\\
LLAVAR & General & 20K & 20K\\
SketchyVQA & General & 4K & 4K\\
IDK & General & 11K & 11K\\
AlfworldGPT & General & 45K & 9K\\
LNQA & General & 303K & 23K\\
Face-Emotion & General & 1K & 1K\\
SpatialSense & General & 10K & 10K\\
Indoor-QA & General & 3K & 3K\\
Place365 & General & 19K & 19K\\
MMInsturct-QA & General & 167K & 23K\\
DriveLM & General & 4K & 4K\\
YesBut & General & 1K & 1K\\
WildVision & General & 6K & 6K\\
LLaVa-Critic-113K & General & 113K & 56K\\
RLAIF-V & General & 83K & 14K\\
VQAv2 & General & 83K & 18K\\
MMRA & General & 1K & 1K\\
KONIQ & General & 30K & 30K\\
MMDU & General & 45K & 23K\\
Spot-The-Diff & General & 9K & 9K\\
Hatefull-Memes & General & 9K & 9K\\
COCO-QA & General & 46K & 23K\\
NLVR2 & General & 50K & 25K\\
Mimic-CGD & General & 71K & 7K\\
Datikz & General & 44K & 8K\\
Chinese-Meme & General & 5K & 5K\\
IconQA & General & 27K & 27K\\
Websight & General & 10K & 10K\\
\hline
\end{tabular}%
}

\caption{General VQA Data.}
\label{app_general_vqa}
\end{table}

\section{Dataset}
We show the detailed used number of samples of every data source in Tab.~\ref{app_general_vqa}, Tab.~\ref{app_nvive_ocr}, Tab.~\ref{app_counting}, Tab.~\ref{app_science}, Tab.~\ref{app_math}, Tab.~\ref{app_caption}, Tab.~\ref{app_chart_table}, Tab.~\ref{app_text_only}, and Tab.~\ref{app_ocrqa}.

In addition to these existing data, we will also provide the augmented data information later.

\begin{table}[tbh]
\centering
\renewcommand{\arraystretch}{1.1}
\resizebox{\linewidth}{!}{%
\begin{tabular}{l|c|c|c}
Dataset&Category&  \#Stage-1.5& \#Stage-2\\
\hline
SynthDog & Naive OCR & 100K & 400\\
MTWI & Naive OCR & 10K & 10K\\
LVST & Naive OCR & 30K & 30K\\
SROIE & Naive OCR & 34K & 1K\\
FUNSD & Naive OCR & 199 & 199\\
Latex-Formula & Naive OCR & 110K & 6K\\
IAM & Naive OCR & 58K & 16K\\
Handwriting-Latex & Naive OCR & 100K & 3K\\
ArT & Naive OCR & 55K & 14K\\
CTW & Naive OCR & 26K & 26K\\
ReCTs & Naive OCR & 20K & 20K\\
COCO-Text & Naive OCR & 16K & 16K\\
SVRD & Naive OCR & 2K & 2K\\
Hiertext & Naive OCR & 10K & 10K\\
RoadText & Naive OCR & 200 & 200\\
MapText & Naive OCR & 240 & 240\\
CAPTCHA & Naive OCR & 10K & 10K\\
Est-VQA & Naive OCR & 17K & 17K\\
HME-100K & Naive OCR & 75K & 37K\\
TAL-OCR-ENG & Naive OCR & 10K & 10K\\
TAL-HW-Math & Naive OCR & 22K & 22K\\
IMGUR5K & Naive OCR & 6K & 6K\\
ORAND-CAR & Naive OCR & 5K & 5K\\
Invoices-and-Receipts & Naive OCR & 2K & 2K\\
Chrome-Writting & Naive OCR & 9K & 9K\\
IIIT5K & Naive OCR & 2K & 2K\\
K12-Printing & Naive OCR & 257K & 51K\\
Memotion & Naive OCR & 6K & 6K\\
Arxix2Markdown & Naive OCR & 502K & 50K\\
HW-Mathematicsl-Exp. & Naive OCR & 12K & 12K\\
WordArt & Naive OCR & 5K & 5K\\
Rendered Text & Naive OCR & 10K & 10K\\
Handwriting-Forms & Naive OCR & 1K & 1K\\ 
\hline
\end{tabular}%
}

\caption{Naive OCR Data.}
\label{app_nvive_ocr}
\end{table}

\begin{table}[h!]
\centering
\renewcommand{\arraystretch}{1.2} 
\resizebox{0.45\textwidth}{!}{%
\begin{tabular}{l|c|c|c}
Dataset&Category&  \#Stage-1.5& \#Stage-2\\
\hline
TallyQA & Counting & 133K & 12K\\
OODVQA & Counting & 3K & 3K \\
RefCOCO/+/g & Grounding & 105K & 25K\\
GroundUI & Grounding  & 17K & 8K  \\
Object365 & Grounding & 1184K & 0\\
\hline
\end{tabular}%
}
\caption{Counting \& Grounding Data.}
\label{app_counting}
\end{table}

\begin{table}[h!]
\centering
\renewcommand{\arraystretch}{1.2} 
\resizebox{0.45\textwidth}{!}{%

\begin{tabular}{l|c|c|c}
Dataset&Category&  \#Stage-1.5& \#Stage-2\\
\hline
AI2D & Science & 12K$\times$4 & 12K$\times$4\\
ScienceQA & Science & 13K$\times$4 &  13K$\times$2\\
TQA & Science & 7K & 7K \\
PathVQA & Science  & 33K & 1K \\
SciQA & Science & 296K & 7K\\
VQA-RAD & Science & 313 & 313\\
VisualWebInsturct & Science & 263K & 263K\\
TextBooks-QA & Science & 47K & 47K\\
\hline
\end{tabular}%
}

\caption{Science Data. $\times n$ notes repeat the data by $n$ times.}
\label{app_science}
\end{table}

\begin{table}[h!]
\centering
\renewcommand{\arraystretch}{1.1} 
\resizebox{0.45\textwidth}{!}{%

\begin{tabular}{l|c|c|c}
Dataset&Category&  \#Stage-1.5& \#Stage-2\\
\hline
GeoQA+ & Math & 177K & 13K\\
MathQA & Math & 40K & 40K\\
CLEVR & Math & 70K & 3K\\
CLEVR-Math & Math & 70K & 3K\\
MAVIS-math-rule-geo & Math & 100K & 100K \\
MAVIS-math-mategen & Math & 86K & 86K\\
InterGPS & Math & 1280 & 1280 \\
Raven & Math & 43K & 31K \\
GEOS & Math & 498 & 498\\
UniGeo & Math & 12K & 12K\\
\hline
\end{tabular}%
}

\caption{Math Data.}
\label{app_math}
\end{table}

\begin{table}[h!]
\centering
\renewcommand{\arraystretch}{1.1} 
\resizebox{0.45\textwidth}{!}{%

\begin{tabular}{l|c|c|c}
Dataset&Category&  \#Stage-1.5& \#Stage-2\\
\hline
ShareGPT4o & Captioning & 57K & 12K \\
KVQA & Knowledge & 24K & 24K \\
Movie-Posters & Knowledge & 15K & 15K \\
Google-Landmark & Knowledge & 26K & 26K\\
WikitArt & Knowledge & 12K & 12K\\
Weather-QA & Knowledge & 1100 & 1100\\
Coco-colors & Captioning & 44K & 22K\\
music-sheet & Knowledge & 9K & 9K\\
SPARK & Captioning & 6K & 6K \\
SAM-caption & Captioning & 78K & 39K \\
Tmbd-Celeb-10K & Knowledge & 8K & 8K\\
CC3M & Captioning & 2237K & 0 \\
Textcaps & Captioning & 110K & 0\\
ShareGPT-4V & Captioning & 767K & 0\\
DenseFusion & Captioning & 1171K & 0\\
\hline
\end{tabular}%
}
\caption{Caption \& Knowledge Data.}
\label{app_caption}
\end{table}

\begin{table}[h!]
\centering
\renewcommand{\arraystretch}{1.0} 
\resizebox{0.45\textwidth}{!}{%
\begin{tabular}{l|c|c|c}
Dataset&Category&  \#Stage-1.5& \#Stage-2\\
\hline
ChartQA & Chart & 60K & 60K\\
MMC-Inst & Chart & 363K & 11K\\
DVQA & Chart & 197K & 8K\\
PlotQA & Chart & 157K & 7K \\
LRV-Instruction & Chart & 7K & 7K\\
TamMWP & Table & 23K & 23K\\
UniChart & Chart & 956K & 33K\\
Vistext & Table & 10K & 10K\\
TAT-DQA & Table & 2K & 2K\\
VQAonDB & Table & 34K & 40K\\
FigureQA & Chart & 100K & 29K \\
Chart2Text & Chart & 27K & 27K\\
Robut & Table & 111K & 23K\\
MultiHiertt & Table & 7K & 7K\\
\hline
\end{tabular}%
}

\caption{Chart \& Table Data. We heavily use some low-quality data such as MMC-Inst, PlotQA in Stage-1.5. But in our final stage, we just sample a very small part from these sources.}
\label{app_chart_table}
\end{table}

\begin{table}[h!]
\centering
\renewcommand{\arraystretch}{1.1} 
\resizebox{0.48\textwidth}{!}{%

\begin{tabular}{l|c|c|c}
Dataset&Category&  \#Stage-1.5& \#Stage-2\\
\hline
DocVQA & OCR QA & 39K$\times$3 & 39K\\
InfoVQA & OCR QA & 24K$\times$4 & 24K$\times$4\\
TextVQA & OCR QA & 35K$\times$4 & 35K$\times$2 \\
ArxivQA & OCR QA & 54K & 3K\\
ScreenQA & OCR QA & 33K & 1K\\
DocReason & OCR QA & 9K & 9K\\
Ureader & OCR QA & 75K & 37K \\
FinanceQA & OCR QA & 10K & 10K\\
DocMatrix & OCR QA & 250K & 7K\\
A-OKVQA & OCR QA & 8K & 8K\\
Diagram-Image-To-Text & OCR QA & 300 & 300\\
MapQA & OCR QA & 37K & 37K\\
OCRVQA & OCR QA & 166K & 83K \\
ST-VQA & OCR QA & 17K & 17K\\
SlideVQA & OCR QA & 6K & 6K\\
PDF-VQA & OCR QA & 9K & 9K\\
SQuAD-VQA & OCR QA & 87K & 46K\\
VQA-CD & OCR QA & 330 & 330 \\
Block-Diagram & OCR QA & 48K & 1K\\
MTVQA & OCR QA & 7K$\times$4 & 7K$\times$4\\
ColPali & OCR QA & 46K & 23K \\
BenthanQA & OCR QA & 19K & 19K\\
\hline
\end{tabular}%
}

\caption{OCR QA Data. ``$\times 4$" means we repeat every sample 4 times.}
\label{app_ocrqa}
\end{table}

\begin{table}[htbp]
\centering
\renewcommand{\arraystretch}{1.1} 
\resizebox{0.45\textwidth}{!}{%

\begin{tabular}{l|c|c|c}
Dataset&Category&  \#Stage-1.5& \#Stage-2\\
\hline
Orca & Text-only & 492K & 49K \\
Orca-math & Text-only  & 199K & 99K \\
MathInsturct & Text-only & 279K & 130K \\
OpenMathInsturct & Text-only & 1580K & 0\\
WizardLM & Text-only  & 70K & 42K\\
TheoremQA & Text-only & 796 & 796 \\
OpenHermes2.5 & Text-only & 99K & 50K\\
NuminaMath-CoT & Text-only & 349K & 140K \\
Python-Code-25k & Text-only & 25K & 25K \\
Infinity-Instruct & Text-only & 303K & 121K\\
Python-18k-Alpaca & Text-only & 18K & 18K \\
Ruozhiba & Text-only & 1734 & 1734 \\
Infinity-Math & Text-only & 74K & 74K\\
StepDPO & Text-only & 11K & 11K\\
TableLLM & Text-only & 73K & 36K\\
UltraInteract-sft & Text-only & 279K & 84K \\
\hline
\end{tabular}%
}

\caption{Text-only Data. The quality of text-only data still matters for multi-modal LLMs. We collect a diverse collection of open-source text-only datasets. We also convert some preference datasets into SFT format.}
\label{app_text_only}
\end{table}

\subsection{Dataset Collection}

\begin{figure*}[t!]
    \centering
    \includegraphics[width=0.8\textwidth]{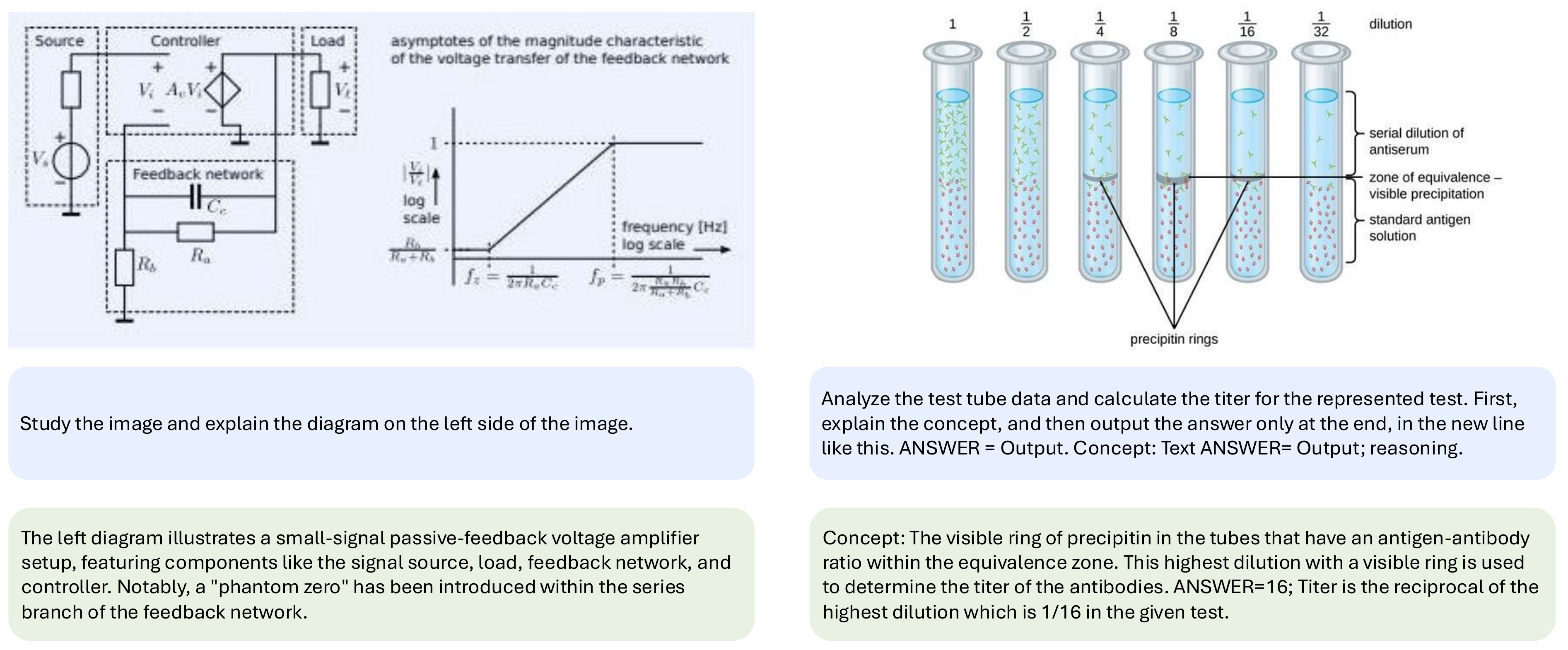}
    \caption{Samples of Internal TextbookQA Dataset. }
    \label{fig:app_textbookqa}
\end{figure*}

\noindent\textbf{Internal Data}
To augment the existing OCR data, we used some internal PDF OCR annotated data, notated as Arxiv2Markdown in this work, which converts each page of papers from into the corresponding Markdown format. For this dataset, we use 500k in stage-1.5 and 50k in stage-2. We also use a dataset created from the textbooks as shown in Fig.~\ref{fig:app_textbookqa}, 47K samples are used in both stage-1.5 and stage-2.

\begin{figure}[t!]
	\centering
	\includegraphics[width=0.45\textwidth]{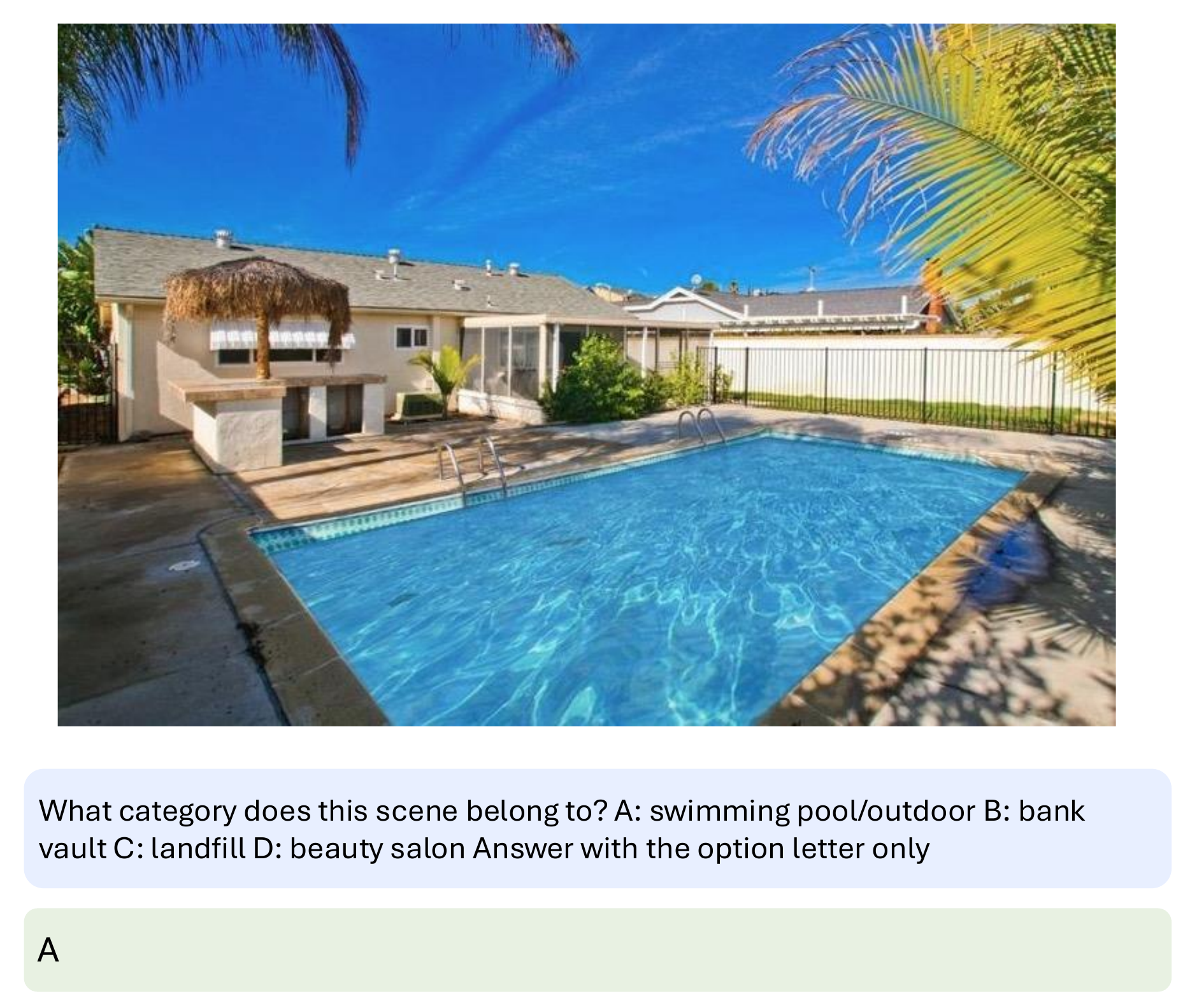}
	\caption{Sample of SpatialScene Dataset. }
	\label{fig:app_spatialscene}
\end{figure}

\noindent\textbf{Non-QA data conversion}
Some of the data source was originally in a non-QA format. If it is classification data, we convert it into multiple-choice questions, as shown in Fig.~\ref{fig:app_spatialscene}. For certain datasets with only images, we use automated annotation tools to generate image descriptions, transforming them into captioning data.

\noindent\textbf{Similarity Score}
our designed similarity score can quickly help us assess the overlap between new data and the existing data pool. Here, we provide an example: if the current data pool is Cambrian-7M, and we aim to introduce new datasets such as UReader, COCO-Colors, and Textbook-QA as shown in Tab.~\ref{appendix_data_rel}. 
Since UReader is a data collection contains DocQA, InfoQA and ChartQA which is already included in Cambrian-7M, we can observe that it has a high similarity score 0.45. The dataset COCO-colors uses COCO images but has new instruction about image colors, so that it has a middle-level score 0.10. Textbooks-QA is our internal data, so it has a relative low score 0.02. In our practice, sources with a score below 0.3 are considered different from the existing data pool. 
Data with a score above 0.3 may also be retained or removed based on specific considerations. Given the relatively high quality of the data within Ureader, we chose to retain it.

\begin{table}[tbh]
\centering
\renewcommand{\arraystretch}{1.1}
\setlength{\tabcolsep}{14pt}
\resizebox{0.45\textwidth}{!}{%
\begin{tabular}{l|c|c}
Dataset&Similarity Score& Max value\\
\hline
Ureader & 0.45 & 0.95\\
Coco-Colors & 0.10 & 0.3\\
Textbooks-QA & 0.02 & 0.1\\
\hline
\end{tabular}%
}

\caption{Similarity Score of new data source to Cambrian. We can found that using similarity score can roughly reflect the overlap between new introduced dataset with the existing data pool.}
\label{appendix_data_rel}
\end{table}

\subsection{Dataset Filtering}

Our data filtering strategy mainly relies on manual inspection to extract the key features of erroneous data and then filter them through rules.

In addition to the common error data mentioned in the paper, there are other types of errors, but their proportion is relatively small.
For example. we notice a particular type of annotated data where the responses are similar to “I cannot answer this.” For questions involving safety or ethical issues, such responses are appropriate and even necessary. However, some data, such as in pure-text form where the question is “Can you help me describe this image” and the answer is “Sorry, I cannot”, are evidently unsuitable for continued use in VLM training. Therefore, we designed a set of keyword-based filtering rules to exclude these samples.

\subsection{Subset Selection}

As shown in Tab.~\ref{app_general_vqa},
Tab.~\ref{app_nvive_ocr}, 
Tab.~\ref{app_counting},
Tab.~\ref{app_science}, 
Tab.~\ref{app_math}, 
Tab.~\ref{app_caption}, 
Tab.~\ref{app_chart_table}, 
Tab.~\ref{app_text_only} and Tab.~\ref{app_ocrqa}, we have several general rules from subset selection.
\begin{itemize}
    \item For datasets with fewer than 20,000 samples, we do not perform subset selection.
    \item If we perform subset selection, we remove at least half of the data. For datasets originally exceeding 100,000 samples, in most cases, we limit the subset to no more than 50,000 samples.
\end{itemize}

\subsection{Data Augmentation}
\noindent\textbf{CoT Augmentation.} 
We use existing SOTA VLMs to help us re-write some Science, Math or Chart data to  generate detailed chain-of-thought answer.
We list datasets used to augment here in Tab.~\ref{appendix_cot_data}. We show our prompt used for CoT augmentation in List~\ref{lst:cot_prompt}.
\begin{table}[t]
\centering
\resizebox{\linewidth}{!}{%

\begin{tabular}{l|c|c|c}
Dataset & Category & Original Answer Type & \#Samples\\
\hline
TQA-CoT & Science & Only Option Letter & 5K \\
ChartQA-CoT & Science & Only Final Answer & 24K \\
DVQA-CoT & Chart & Only Final Answer & 25K \\
Clever-CoT & Math & Only Final Answer & 13K \\
Clever-Math-CoT & Math & Only Final Answer & 59K \\
SketchyVQA-CoT & General & Yes/No Answer & 8K \\
Tab-MWP-CoT & Table & Brief Explanation & 20K \\
RAVEN-CoT & Math & Option Letter & 9K \\
MAVIS-math-metagen-CoT & Math & Unformatted Long Answer & 86K \\
UniGeo-CoT & Math & Only Final Answer & 12K \\
\hline 
\end{tabular}%
}

\caption{Dataset for CoT data augmentation.}
\label{appendix_cot_data}
\end{table}

\begin{table}[t!]
\centering
\resizebox{0.45\textwidth}{!}{%

\begin{tabular}{c|c|c}
training w/ CoT Data& Evaluating w/  CoT Prompt & MathVista-Mini\\
\hline
\xmarker &   \cmarker & 61.0\\
\xmarker &   \cmarker & 60.5\\
\cmarker &   \cmarker & 63.2\\
\cmarker &   \cmarker & 63.5\\

\end{tabular}%
}

\caption{With CoT training data, adding "Solve this problem step-by-step" prompt can help to improve the performance.}
\label{appendix_cot_prompt}
\end{table}
\begin{table*}[t]
\scriptsize
\centering
\setlength\tabcolsep{2.5pt}
\renewcommand{\arraystretch}{1.1}
\resizebox{\textwidth}{!}{
\begin{tabular}{l|c|ccccccccccccc|c}
\hline
\multirow{2}{*}{Model} & packing method & DocVQA & ChartQA & InfoVQA & TextVQA & OCRBench & MMstar & RWQA & AI2D & MMMU & MMB$_{1.1}$ & MMVet & HallB & MathVista & Average \\
                       &                                           & Test   & Test    & Test    & Val     & Test    & Test   & Test & Test & Val  & EN-Val       & Test  & Test  & Test-Mini & Score   \\
\hline
Eagle2-9B        & Greedy & 92.6  & 84.7  & 76.5  & 83.8 & 855 & 62.7 & 67.8 & 84.0 & 54.7 & 81.7 & 63.0 & 47.9 & 61.6 & 72.8  \\
Eagle2-9B        & Blanced & 92.6  & 86.4  & 77.2  & 83.0 & 868 & 62.6 & 69.3 & 83.9 & 56.1 & 81.9 & 62.2 & 49.3 & 63.8 & 73.5  \\
\hline
\end{tabular}}
\caption{The overall performance of using  naive greedy packing is inferior to that of using balanced-aware packing.}
\label{tab:app_packing_abl}
\end{table*}


In fact, using this automated augmentation method can easily generate incorrect answers or solution steps, even when the correct answer is provided in the prompt. To address this, we employed another LLM to compare the generated answers with the original answers, filtering out some erroneous sample, the prompt is shown in List~\ref{lst:cot_prompt_judger}.

To verify the effectiveness of the constructed CoT data, we evaluate it on MathVista. For models that do not train on the generated CoT data, when adding the CoT prompt "Think it step-by-step" to the question, they do give answers in a CoT format, but unfortunately do not improve accuracy or even significantly reduce it. However, after incorporating our constructed CoT data, using the CoT prompt bring performance boost, as shown in Tab.~\ref{appendix_cot_prompt}.

\begin{figure}[t!]
    \centering
    \includegraphics[width=0.47\textwidth]{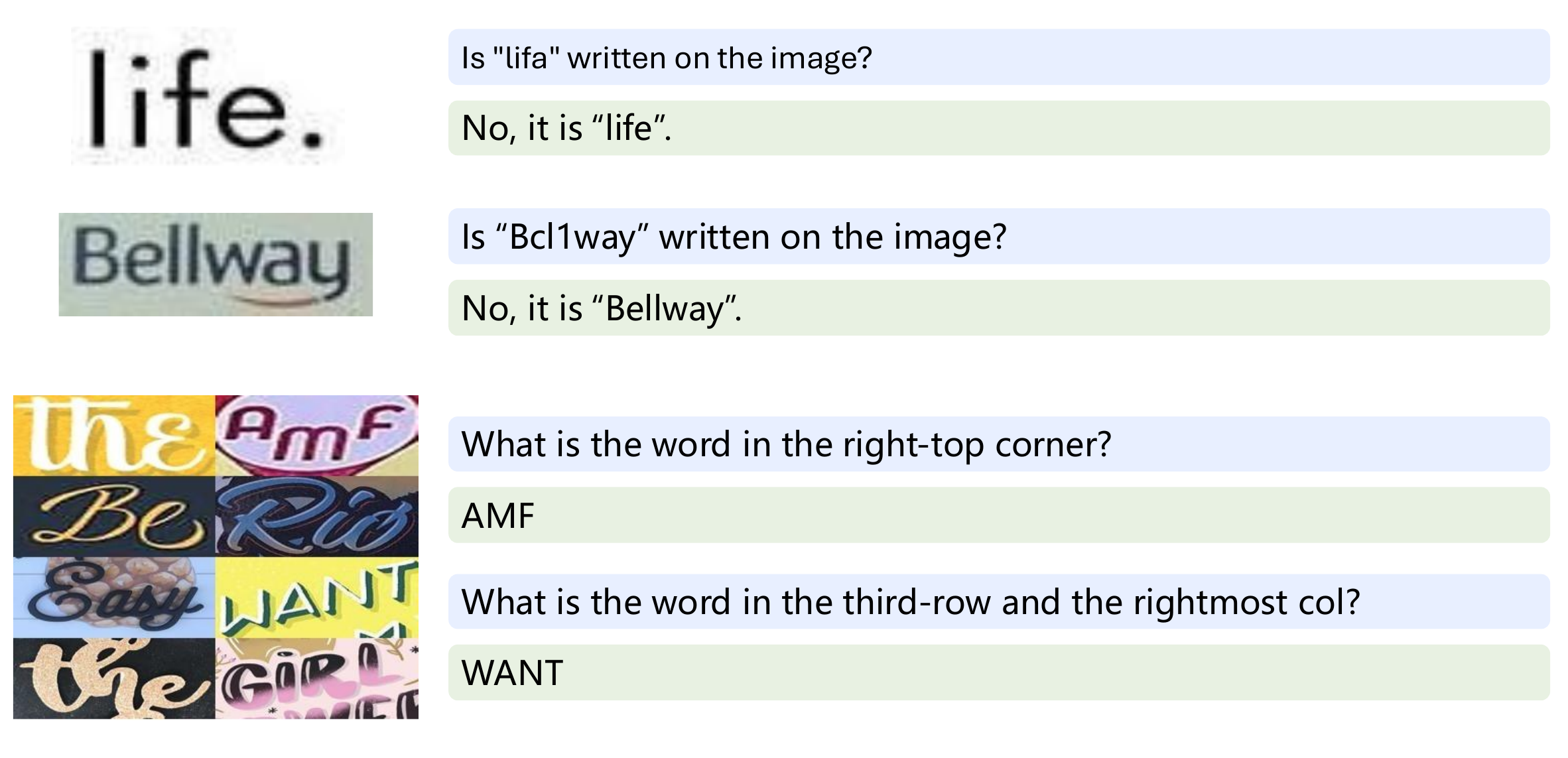}
    \caption{Rule-based data augmentation for OCR data. }
    \label{fig:app_ocr_data_aug}
\end{figure}

\noindent\textbf{Rule-Based QA Generation}
In addition to the previous In addition to the augmentation method for table data, we also designed several other rules to expand the existing dataset as shown in Fig.~\ref{fig:app_ocr_data_aug}. These tasks are designed to reduce model hallucinations and enhance the model's spatial awareness capabilities. Specifically, we use IIIT5K and WordArt dataset for OCR data augmentation.

\noindent\textbf{Expanding Short Answers.}
We extending the short response in dataset VQAv2, GQA and VSR to detailed response via the prompt List~\ref{lst:cot_prompt_expanding}.

\section{Packing}
The knapsacks generated by the naive greedy packing strategy exhibit an unnecessary length distribution bias. To address this, we designed a balanced-aware knapsack method aimed at producing knapsacks with a more uniform length distribution. We show the ablation results with different packing methods in Tab.~\ref{tab:app_packing_abl}. Additional details on practical implementation are provided in List~\ref{lst:greedy_knapsack_actually_using}.

\begin{lstlisting}[
	float=t,
	language=Python,
	floatplacement=t,
	xleftmargin=2mm,
	xrightmargin=2mm,
	framexleftmargin=2mm,
	framexrightmargin=2mm,
	showlines=true,
	belowskip=-1\baselineskip,
	basicstyle=\ttfamily\scriptsize,
	breakatwhitespace=false,
	breaklines=true,
	captionpos=b,
	keepspaces=true,
	showspaces=false,
	showstringspaces=false,
	showtabs=false,
	label={lst:cot_prompt},
	caption={Prompt we used for CoT Augementation.}]
	
	f"""Rewrite the following answer using a **Chain of Thought (CoT)** approach. The final answers should adhere to the following structure and constraints:
	
	1. **Problem Restatement**: Start by restating the problem clearly to set the context.
	
	2. **Step-by-Step Process**:
	- **Explicit Steps**: Break the solution into **discrete steps**, showing all calculations.
	- **Justifications**: Include a brief explanation for each step (e.g., referencing mathematical rules such as the distributive property, derivative rules, or solving equations).
	
	3. **Mathematical Principles**: Where relevant, mention the specific mathematical principles or theorems being applied (e.g., chain rule, Pythagoras' theorem, etc.).
	
	4. **Final Answer**: End with the final solution, clearly boxed or highlighted.
	
	5. **Consistent Structure**: Ensure every solution follows this format:
	- **Restatement of the problem**
	- **Steps and calculations with justifications**
	- **Final answer**
	
	The output should be detailed but concise, explaining each step logically while avoiding excessive repetition. Clarity and logical flow are crucial.
	
	Here is a question and answer pair of this image:
	Question: {question}
	Answer: {answer}
	"""
	
\end{lstlisting}

\begin{lstlisting}[
	float=t,
	language=Python,
	floatplacement=t,
	xleftmargin=2mm,
	xrightmargin=2mm,
	framexleftmargin=2mm,
	framexrightmargin=2mm,
	showlines=true,
	belowskip=-1\baselineskip,
	basicstyle=\ttfamily\scriptsize,
	breakatwhitespace=false,
	breaklines=true,
	captionpos=b,
	keepspaces=true,
	showspaces=false,
	showstringspaces=false,
	showtabs=false,
	label={lst:cot_prompt_judger},
	caption={Prompt we used for judging the correctness of generation chain-of-thought response.}]
	
	f"""Please evaluate if the correctness of my answer based on the provided question and the correct answer.
	
	Question: {question}
	Correct Answer: {ori_answer}
	My Answer: {new_answer}
	
	Please only return "True" if my answer is correct, or "False" if it is incorrect.
	My answer is:"""
	
\end{lstlisting}

\begin{lstlisting}[
	float=t,
	language=Python,
	floatplacement=h,
	xleftmargin=2mm,
	xrightmargin=2mm,
	framexleftmargin=2mm,
	framexrightmargin=2mm,
	showlines=true,
	belowskip=-1\baselineskip,
	basicstyle=\ttfamily\scriptsize,
	breakatwhitespace=false,
	breaklines=true,
	captionpos=b,
	keepspaces=true,
	showspaces=false,
	showstringspaces=false,
	showtabs=false,
	label={lst:cot_prompt_expanding},
	caption={Prompt we used for expanding short answers.}]
	
	f"""Given the question {question}. The original answer is {answer}.
	Please reply with a more specific answer based on the existing answer, as detailed as possible."""
	
\end{lstlisting}

\begin{lstlisting}[
	float=t,
	language=Python,
	floatplacement=h,
	xleftmargin=2mm,
	xrightmargin=2mm,
	framexleftmargin=2mm,
	framexrightmargin=2mm,
	showlines=true,
	belowskip=-1\baselineskip,
	basicstyle=\ttfamily\scriptsize,
	breakatwhitespace=false,
	breaklines=true,
	captionpos=b,
	keepspaces=true,
	showspaces=false,
	showstringspaces=false,
	showtabs=false,
	label={lst:greedy_knapsack_actually_using},
	caption={In practical implementation, we added an extra redundancy delta to min\_knapsacks to avoid creating new knapsacks within the loop. Without this delta, knapsacks with imbalanced distributions could be introduced. The delta value is set based on the data length distribution. The size of samples in our settings is 4k.}]
	
	# Our proposed greedy knapsack method
	def balanced_greedy_knapsack(samples, L, delta=20):
	# Step 1: Sort the samples
	samples.sort(reverse=True)
	total_length = sum(samples)
	min_knapsacks = (total_length + L - 1) // L + delta
	# Step 2: Initialize knapsacks
	knapsacks=[[] for _ in range(min_knapsacks)]
	knapsack_lengths = [0] * min_knapsacks
	# Step 3: Distribute samples across knapsacks
	ks_index = 0
	sample_index = 0
	while sample_index < len(samples):
	length = samples[sample_index]
	if knapsack_lengths[ks_index]+length<=L:
	knapsacks[ks_index].append(length)
	knapsack_lengths[ks_index] += length
	sample_index += 1
	else:
	knapsacks.append([])
	knapsack_lengths.append(0)
	ks_index = argmin(knapsack_lengths)
	
	return knapsacks
	
\end{lstlisting}

\end{document}